\documentclass{article} %
\usepackage[preprint]{goodfire/goodfire}
\IfFileExists{goodfire/goodfire.sty}{}{} %

\usepackage{amsmath,amsfonts,bm}

\def\eqref#1{equation~\ref{#1}}

\def\1{\bm{1}}

\DeclareMathAlphabet{\mathsfit}{\encodingdefault}{\sfdefault}{m}{sl}
\SetMathAlphabet{\mathsfit}{bold}{\encodingdefault}{\sfdefault}{bx}{n}

\usepackage{hyperref}
\usepackage{url}
\usepackage{graphicx}
\usepackage{booktabs}
\usepackage{inconsolata} %
\usepackage{tcolorbox}\tcbuselibrary{raster,skins}
\usepackage[T1]{fontenc}

\usepackage{xcolor}
\definecolor{burgundy}{HTML}{7B2D26}

\title{Neural Geometry as a Learning Signal}

\definecolor{primary}{HTML}{7B2D26} %
\definecolor{secondary}{HTML}{B8973A} %

\providecommand{\aut}[1]{\textbf{#1}}
\providecommand{\af}[1]{{\small #1}}

\providecommand{\afn}[1]{\textcolor{primary}{$^{#1}$}}

\newcommand{\goodfiremark}{\raisebox{0.5pt}{\hspace{0.5mm}\includegraphics[height=6pt]{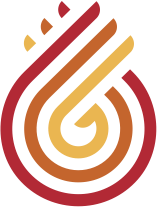}}}

\newcommand{\goodfireaff}{%
  \includegraphics[height=14pt]{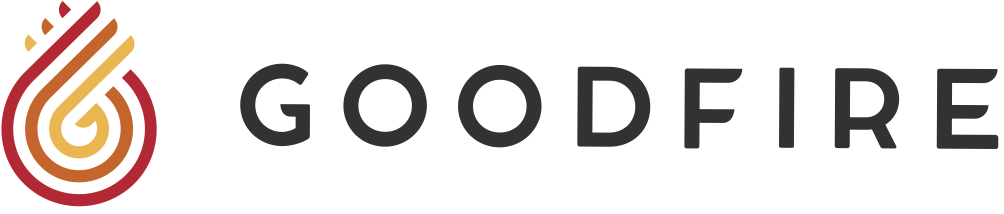}
}

\newcommand{\authorentry}[2]{\aut{#1}\afn{#2}}
\newcommand{\afflabel}[2]{\afn{#1}\af{#2}}
\newcommand{\authorsep}{\quad}
\newcommand{\repolink}[1]{%
  {\small
    \href{#1}{\raisebox{-2.8pt}{\includegraphics[height=10pt]{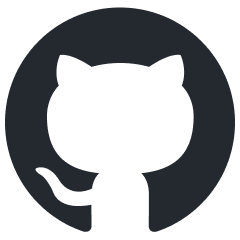}}%
    \texttt{\textcolor{secondary}{#1}}}
  }%
}

\newtoggle{goodfireonly}
\togglefalse{goodfireonly}  %
\newcommand{\extline}[1]{\iftoggle{goodfireonly}{}{#1}}
\newcommand{\extaff}[1]{\iftoggle{goodfireonly}{}{#1}}

\newcommand{\paperauthors}{%
\authorentry{Eric Todd}{\goodfiremark\extaff{,a}} 
\authorentry{Tal Haklay}{\goodfiremark} 
\authorentry{Nikhil Prakash}{\goodfiremark\extaff{,a}} 
\authorentry{Matthew Finlayson}{\goodfiremark\extaff{,b}}
\authorentry{Michael Hanna}{\goodfiremark\extaff{,c}}\\
\authorentry{Fenil R. Doshi}{\goodfiremark\extaff{,d}} 
\authorentry{Daniel Wurgaft}{\goodfiremark\extaff{,e}} 
\authorentry{Thomas McGrath}{\goodfiremark}
\authorentry{Owen Lewis}{\goodfiremark} 
\authorentry{David Bau}{\extaff{a}}\\
\authorentry{Jack Merullo}{\goodfiremark}
\authorentry{Ekdeep Singh Lubana}{\goodfiremark}
\authorentry{Matthew Kowal}{\goodfiremark}
\authorentry{Thomas Fel}{\goodfiremark}
\authorentry{Atticus Geiger}{\goodfiremark}\\
\goodfireaff \\
\extline{%
  \afflabel{a}{Northeastern University} \authorsep
  \afflabel{b}{University of Southern California} \authorsep
  \afflabel{c}{University of Amsterdam} \authorsep\\
  \afflabel{d}{Harvard University} \authorsep
  \afflabel{e}{Stanford University}
}
\vspace{2mm}\\
\repolink{https://github.com/goodfire-ai/geometric-supervision}
\vspace{-4mm}
}

\author{\paperauthors}

\begin{document}

\maketitle

\begin{abstract}
We investigate whether the geometric structure of neural representations can provide a sufficient learning signal for improving model behavior without any supervision on model outputs.
First, we fine-tune Llama-3.1-8B to answer questions such as “What month is four months after June?” by supervising internal representations of the input month (June = 6) and offset (4), which are encoded with Fourier harmonics. Second, we fine-tune the model to predict line breaks in text wrapped to a fixed width by supervising an internal representation of the number of characters since the preceding newline, which is represented by a low-dimensional helical structure.
In both tasks, we supervise a low-dimensional subspace of the residual stream using an idealized target structure, and find that internal geometric supervision achieves similar performance to standard fine-tuning (despite no supervision on outputs). 
Moreover, we find that simply patching in an ideal geometry is not sufficient to recover fine-tuning performance, and thus demonstrate geometric supervision provides beneficial effects beyond the representations it targets. 
Together, these findings show how mechanistic understanding of neural geometry can be leveraged to shape the behavior of neural networks.
\end{abstract}

\section{Introduction}
\label{sec:introduction}

In \citeyear{rumelhart1986learning}, \citeauthor*{rumelhart1986learning} showed
that backpropagating errors from model outputs could shape the representations inside a neural network. Nearly every model trained since then has relied on that same training process: where \textit{behavior} is specified and \textit{representations} follow.
Forty years later, interpretability research has begun to map out the neural geometry of large language models (LLMs), describing how representations are encoded and used to perform specific tasks \citep{park2025hierarchical, zhou2024fourier, engels2025language, kantamneni2025trigonometry, gurnee2025when, karkada2026symmetry, wurgaft2026manifold,bhalla2026do,sarfati2026shape, feucht2026arithmetic}.
However, many accounts of representational geometry describe an implicit idealization that the model's representations do not fully satisfy.
In this paper, we ask whether this descriptive paradigm of interpretability can be reversed: can training these representations toward their ideal geometry improve the model's behavior without directly supervising it?

We study this question by focusing on two tasks with mechanistic accounts of geometric structure in LLM representations with an established algorithmic role.
Specifically, we study Llama-3.1-8B~\citep{grattafiori2024llama3} on queries such as ``What month is four months after June?'' and on predicting line breaks in text wrapped to a fixed width, building on accounts of base-10 addition \citep{levy2025language, kantamneni2025trigonometry, feucht2026arithmetic} and character counting \citep{gurnee2025when}, respectively.
In each case, we train low-rank adapters (LoRAs; \citealt{hu2022lora}) with a loss that aligns observed representations with an idealized geometric target, based on the descriptive account. 
Across both tasks, we find that internal geometric supervision achieves similar performance to standard fine-tuning despite no supervision on outputs. 

We further investigate why geometric supervision is successful and uncover a surprising finding: while both standard fine-tuning and geometric supervision sharpen the existing task-relevant geometry, when we patch in either sharpened representations back into the original model, we do not recover the performance gain. This demonstrates that geometric supervision produces beneficial effects beyond the representations it targets. 
Together, these results both provide a functional validation of prior mechanistic explanations \citep{feucht2026arithmetic, kantamneni2025trigonometry, zhou2024fourier, gurnee2025when} and show that behavioral errors on these tasks can be corrected by specifying what the model's representations should look like, rather than what its outputs should be.

\section{Methodology}
\label{sec:notation}
\textbf{Notation.} Let $f_\theta$ be a transformer language model (LM) with parameters $\theta$, where $\theta_0$ denotes frozen base model parameters. Let $X$ be a set of prompts corresponding to a given task. Each prompt $x \in X$ is associated with a set of task variables $\mathcal{C}$, where each variable $c\in\mathcal{C}$ is a function $c:X \rightarrow\mathcal{V}_c$ that maps a prompt $x$ to a value $v\in\mathcal{V}_c$.
For a given variable $c$, we denote the set of prompts that share the value $v \in \mathcal{V}_c$ as $X_{v} = \{x\in X : c(x) = v\}$. 
Let $\mathbf{h}_x(\theta) \in \mathbb{R}^{d}$ represent a hidden state activation extracted from the model's computation $f_\theta(x)$ under prompt $x$.

\begin{figure}
    \centering
    \includegraphics[width=\linewidth]{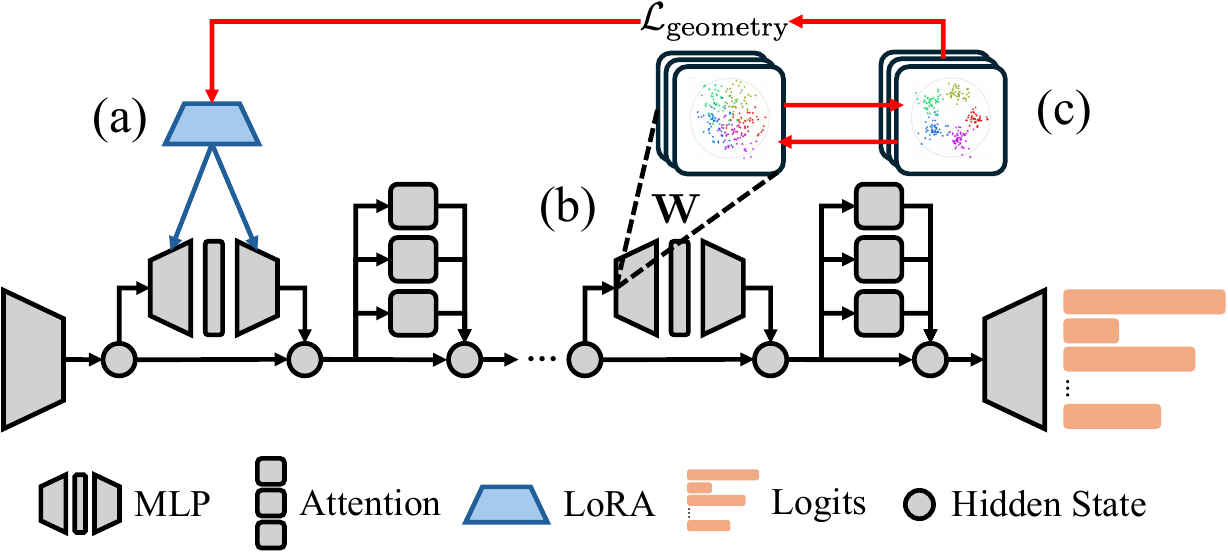}
    \caption{\textbf{An overview of geometric supervision.} \textbf{(a)} A LoRA is placed at an early layer with the goal of aligning empirical geometric structure to an idealized target geometry. 
    \textbf{(b)} Supervision on the task-relevant geometry is applied at an intermediate layer in the $k$-dimensional coordinates of $\mathbf{W}$. \textbf{(c)} The geometry LoRA is trained by minimizing a loss (Equation~\ref{eq:full_loss}) that aligns the observed structure \smash{$\mathbf{\hat G}$} to the target structure $\mathbf{G}$, without any output-level task loss. 
    }
    \label{fig:geometric_supervision_overview}
\end{figure}

For each task we investigate, prior work has identified task-relevant activation geometry that can be read out of the hidden state by a map $\phi:\mathbb{R}^d\to\mathbb{R}^k$, which we call a \textit{readout}.
In both tasks studied here, $\phi$ is affine, $\phi(\mathbf{h}) = \mathbf{A}\mathbf{h} + \mathbf{b}$, and we measure geometry in the readout coordinates.
The activation that a given prompt $x$ takes on under the readout can be written as $\phi(\mathbf{h}_x(\theta))$, and for prompts that share a value $v$, we write the average readout as:
\begin{equation}
    \label{eq:empirical_centroid}
    \boldsymbol{\hat\mu}_{v}(\theta) = \frac{1}{|X_{v}|}\sum_{x\in X_{v}}\phi(\mathbf{h}_x(\theta)).
\end{equation}
\textbf{Our Approach.} We choose tasks that have previously identified task-relevant geometric structure with the goal of using that structure to fix model behavior. Rather than train on the model's output, our approach consists of fine-tuning a model with supervision that aligns empirical geometric structure with an idealized target geometry (see Figure~\ref{fig:geometric_supervision_overview}). Each target structure is task-dependent (e.g., points equally spaced on a circle or helix) and will be described in more detail for each task in Sections~\ref{sec:months_geometric_alignment} and \ref{sec:linebreak_geometric_alignment}.

Specifically, given a model $f_{\theta}$ with empirical geometric structure \smash{$\mathbf{\hat G} = \{\boldsymbol{\hat\mu}_{v}\}_{v\in \mathcal{V}_c} \subset \mathbb{R}^k$}, that has been observed under $\phi$, we supervise the model's representations to match an ``idealized'' target geometric structure $\mathbf{G} = \{\boldsymbol{\mu}_v\}_{v\in\mathcal{V}_c} \subset \mathbb{R}^k$. 
The training objective we use for geometric alignment between \smash{$\mathbf{\hat G}$} and $\mathbf{G}$ consists of two loss terms: 
\begin{equation}
    \label{eq:full_loss}
    \mathcal{L}_{\text{geometry}} = \mathcal{L}_{\text{mean}} + \lambda \mathcal{L}_{\text{spread}}
\end{equation}
where $\lambda$ is a hyperparameter balancing the two terms.
The first term, $\mathcal{L}_{\text{mean}}$, pulls each empirical centroid $\boldsymbol{\hat\mu}_v$ toward its target position $\boldsymbol{\mu}_v$, as seen in Equation~\ref{eq:loss_mean}:
\begin{equation}
\label{eq:loss_mean}
    \mathcal{L}_{\text{mean}} = \frac{1}{|\mathcal{V}_c|}\sum_{v\in\mathcal{V}_c}\frac{1}{r_0^{2}}\bigl\lVert\boldsymbol{\hat\mu}_{v}(\theta) - \boldsymbol{\mu}_v\bigr\rVert_2^{2}
\end{equation}
where $r_0^{2} = \frac{1}{|\mathcal{V}_c|}\sum_{v\in\mathcal{V}_c}||\boldsymbol{\hat\mu}_{v}(\theta_0) - \boldsymbol{\bar\mu}_c(\theta_0)||_2^{2}$ normalizes the loss by the variance of the frozen base model's centroids around their mean $\boldsymbol{\bar\mu}_c(\theta_0) = \frac{1}{|\mathcal{V}_c|}\sum_{v\in\mathcal{V}_c}\boldsymbol{\hat\mu}_{v}(\theta_0)$. 

The second loss term, $\mathcal{L}_{\text{spread}}$, targets a ratio $\tau$ between the fine-tuned and base-model's within-value spread (see Appendix~\ref{appdx:geom_ablations} or Appendix~\ref{appdx:helix_hparam_sweeps} for an ablation study), shown in Equation~\ref{eq:loss_spread}: 
\begin{equation}
\label{eq:loss_spread}
    \mathcal{L}_{\text{spread}} = \left(\frac{\sigma_c(\theta)}{\sigma_c(\theta_0)} - \tau\right)^{2}
\end{equation}
where $\sigma_c(\theta)$ is the root-mean-square (RMS) distance of projected activations to their value centroids:
\begin{equation}
    \label{eq:rms_spread}
    \sigma_c(\theta) = \left( \frac{1}{|X|} \sum_{v \in \mathcal{V}_c} \sum_{x \in X_v} \|\phi(\mathbf{h}_x(\theta)) - \boldsymbol{\hat\mu}_v(\theta)\|_2^2 \right)^{1/2}.
\end{equation}
A ratio of $\tau=1$ would encourage the fine-tuned model's projected activation spread to match the spread of the base model exactly, whereas $\tau > 1$ would increase the spread in the fine-tuned model.

\begin{figure}
    \centering
    \includegraphics[width=\linewidth]{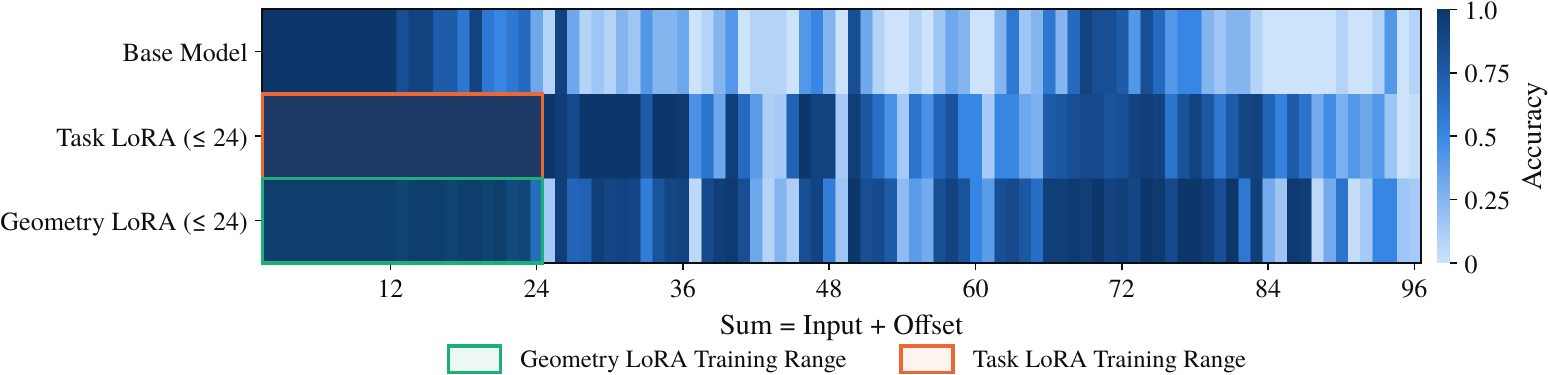}
    \caption{\textbf{Performance of Llama-3.1-8B and fine-tuned variants on the months task.} Inputs are grouped by $\texttt{sum}$, e.g., ``What month is four months after June?'' is grouped with ``What month is nine months after January?'' for having the same \texttt{sum} of $10$.
    The base model (top row) achieves $100\%$ accuracy when the $\texttt{sum}\leq12$, but performance degrades for larger \texttt{sums}, achieving a total accuracy of only $36.3\%$.
    Training a LoRA with end-to-end supervision (``Task LoRA'', middle row) on sums up to 24 improves performance to $71.9\%$, with the strongest gains primarily located within the training range ($X_{\texttt{sum}\leq24}$, orange shaded region) and the first held-out cycle ($\texttt{sum}\in[25,36]$). 
    Training a LoRA with \textit{only} geometric supervision (\S\ref{sec:months_geometric_alignment}) over the same training range (``Geometry LoRA'', bottom row) achieves similar improvements across the board, with a total accuracy of $73.3\%$.}
    \label{fig:months_behavioral_full}
\end{figure}

\section{Supervising Fourier Features for Month Arithmetic}
\label{sec:months}
The first task we study is the \textbf{months task} from \citet{feucht2026arithmetic}.
An individual prompt for the months task takes the form $x=$``\textit{Q: What month is \{offset\} months after \{input\}?}\textbackslash\textit{nA:}'', and the goal of the model is to predict the correct month.
The months task has three variables: $\mathcal{C} = \{\texttt{input}, \texttt{offset}, \texttt{sum}\}$, and the set of prompts $X$ for this task is constructed via the values of two variables---the \texttt{input} and the \texttt{offset}---which determine the third \texttt{sum} variable (i.e., $\texttt{sum} = \texttt{input} + \texttt{offset}$). 
The \texttt{input} takes on values in $\mathcal{V}_{\text{input}} =\{\text{January, February, ... December}\}$ (where January is 1-indexed), and the offset takes on values in $\mathcal{V}_{\text{offset}} = \{1, 2, ..., 95\}$ such that $X$ includes all prompts where \texttt{sum} $\leq 96$ (i.e., $\mathcal{V}_{\text{sum}} = \{2, 3, ..., 96\}$), giving a total of $|X| = 1074$ prompts.
We note that \citeauthor{feucht2026arithmetic} only evaluate the model on prompts where \texttt{offset} $\leq 24 = 2p$ (so that \texttt{sum} $\leq36$), where $p=12$ is the cycle length of the \texttt{input} concept, but we include a larger range of \texttt{offset} values to test for broader generalization (i.e., \texttt{sum}~$\leq 96 = 8p$).

\textbf{Evaluation of end-to-end fine-tuning.} We evaluate Llama-3.1-8B~\citep{grattafiori2024llama3} on the months dataset ($X_{\texttt{sum}\leq96}$) and find that it achieves $100\%$ accuracy when the value of \texttt{sum} is small $(\leq 12)$, but quickly degrades as the value of \texttt{sum} increases, achieving an overall accuracy of $36.3\%$ (Figure~\ref{fig:months_behavioral_full}, top row). 
We test whether fine-tuning the model can correct these errors by training a low-rank adapter~\citep[LoRA;][]{hu2022lora} using a standard cross-entropy loss between the model's prediction and the correct month.
We use $X_{\texttt{sum}\leq24}$ as the training set, and evaluate on all prompts in $X_{\texttt{sum}\leq 96}$.
With the LoRA, the model achieves a total accuracy of $71.9\%$ (Figure~\ref{fig:months_behavioral_full}, middle row), fixing almost all errors within the LoRA's training distribution (Figure~\ref{fig:months_behavioral_full}, shaded orange region), and also improving accuracy on the first held-out cycle $(\texttt{sum}{\in}[25,36])$.
Here, we report the results of the best task LoRA, but include results and training details for other configurations in Appendix~\ref{appdx:task_lora_training}. 

\subsection{End-to-end LoRAs Strengthen Existing Representations}
\label{sec:month_sharpening}
To understand what has changed in the model under LoRA fine-tuning, we analyze the model's task-relevant geometry.
Specifically, \cite{feucht2026arithmetic} has identified a set of Fourier features with base-10 periods (e.g., 2, 5, 10, etc.) at layer $18$ that Llama-3.1-8B uses to compute the \texttt{sum} variable for both cyclic arithmetic and standard addition tasks.
We train a set of Fourier probes for each period $T \in \{2, 5,10,20,50, 100\}$ for both sine and cosine at each model layer $l$.
The probes are trained on a simple addition task (``\textit{a+b=}'') to match either $\sin(\frac{2\pi c(x)}{T})$ or $\cos(\frac{2\pi c(x)}{T})$, where $c \in \{a, b, a+b\}$ is the variable being probed for (see Appendix~\ref{appdx:fourier_probes} for details).
For a prompt $x$, we can project the final token hidden state $\mathbf{h}_x(\theta)$ onto the Fourier plane for variable $c$ and period $T$ at layer $l$, denoted as \smash{$\phi_c^{(l,T)}$}, by applying the corresponding trained probes:
\begin{equation}
\label{eq:affine_probe}
    \smash{\phi_c^{(l,T)}(\mathbf{h}_x(\theta)) = ({\mathbf{w}^{(l,T)}_{\text{cos}}} \cdot \mathbf{h}_x(\theta) + b^{(l,T)}_{\text{cos}}, \mathbf{w}^{(l,T)}_{\text{sin}} \cdot \mathbf{h}_x(\theta) + b^{(l,T)}_{\text{sin}}) \in \mathbb{R}^{2}}
\end{equation}
where \smash{$\mathbf{w}_{\text{cos}}^{(l, T)}, \mathbf{w}_{\text{sin}}^{(l, T)} \in \mathbb{R}^{d}$} and \smash{$b_{\text{cos}}^{(l, T)}, b_{\text{sin}}^{(l, T)} \in \mathbb{R}$} are the probe weights. 

\begin{figure}
    \centering
    \includegraphics[width=\linewidth]{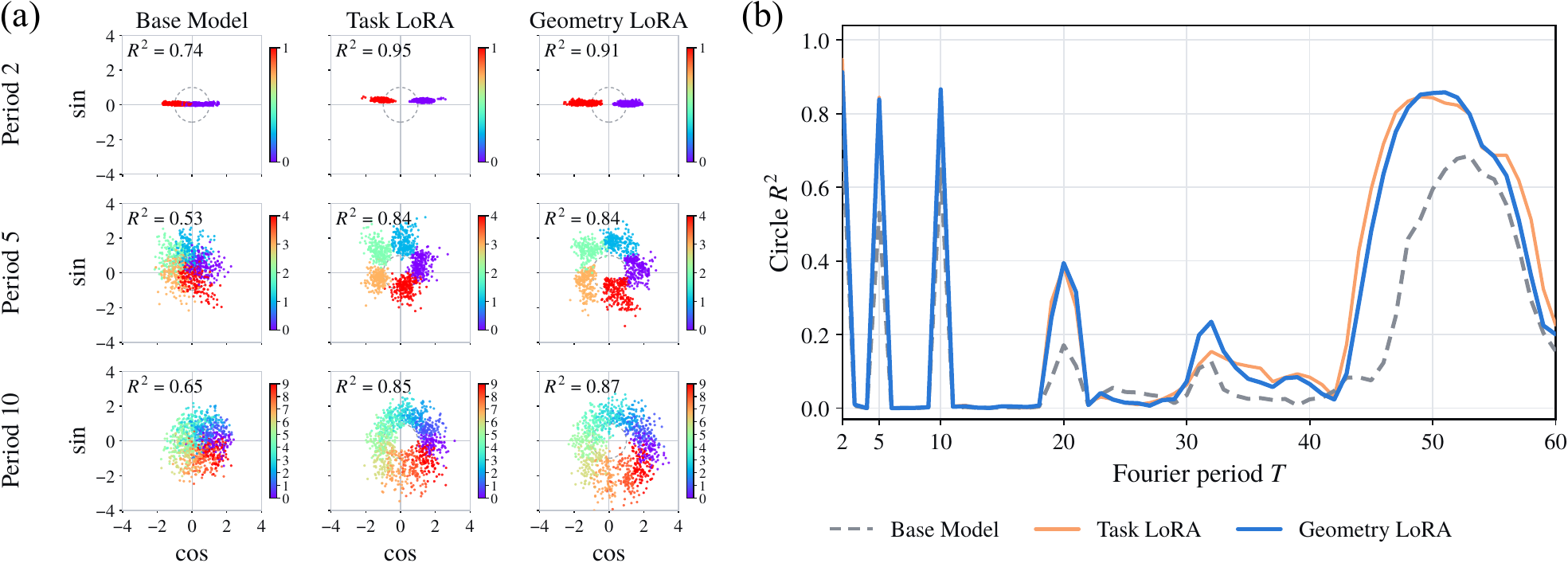}
    \caption{\textbf{Fine-tuning sharpens base-10 Fourier representations.} \textbf{(a)} Projection of the months data from the base model, task LoRA, and geometry LoRA onto planes containing Fourier features for the \texttt{sum} variable with periods $T=\{2, 5, 10\}$ at layer $18$.
    Each point corresponds to a prompt $x \in X_{\texttt{sum}\leq96}$, and is colored by $\texttt{sum}(x) \bmod T$.
    Both end-to-end training and geometric supervision sharpen existing base-10 Fourier representations, resulting in tighter per-modulo clusters, and higher alignment with idealized Fourier structure. \textbf{(b)} Alignment of representations with ideal Fourier planes shown for periods $T\in\{2,...,60\}$. Base-10 periods sharpen under the task LoRA (orange) and geometry LoRA (blue) compared to the base model (gray), while other periods do not.}
    \label{fig:months-L18-sum-projection}
\end{figure}

We inspect the geometry by projecting the months data onto each Fourier plane and find that the circular structure of the base-10 periods is visibly sharper in the LoRA fine-tuned models compared to the base model (Figure~\ref{fig:months-L18-sum-projection}a; see also Appendix~\ref{appdx:fourier_probes}). 
We measure the $R^2$ value of the \texttt{sum} Fourier planes and find that for the LoRA fine-tuned models, the $R^2$ value of the base-10 periods ($T\in\{2, 5, 10, 20, 50\}$) increases on average by $0.23$ at layer $18$, while the average change of other periods is ${\sim}0.04$ (Figure~\ref{fig:months-L18-sum-projection}b).
An increased alignment of base-10 periods with the ideal Fourier geometry and a minimal effect on other periods suggests that LoRA fine-tuning might achieve its performance gain by strengthening the existing representational structure.

\subsection{Geometric Supervision on Internal Representations}
\label{sec:months_geometric_alignment}
Motivated by the sharpening of existing representational structure seen in Section~\ref{sec:month_sharpening}, we investigate whether targeting geometric structure directly can provide a sufficient learning signal to correct the model's behavior without a loss on the model output. 
Training a LoRA with only a geometric loss requires choosing: (1) where to supervise the geometry (e.g., an affine readout of the residual stream), and (2) what the geometry should look like (i.e., an idealized target geometry $\mathbf{G}$).

We experimented with many configurations of geometric supervision (see Appendix~\ref{appdx:geometry_lora_training}) and achieved the best results at the final-token residual stream at layer $l^*=18$, (the layer at which \citet{feucht2026arithmetic} localize the \texttt{sum} computation based on the \texttt{input} and \texttt{offset} representations).
We use a set of eight Fourier planes, one for each pair $(c,T)$ with $c\in\mathcal{C}'=\{\texttt{input}, \texttt{offset}\}$ and $T\in \mathcal{T}=\{2, 5, 10, 20\}$. 
Together, they define coordinates in a $k$-dimensional subspace $\mathbf{W}$:
\begin{equation}
    \mathbf{W}\mathbf{h}_x(\theta) \;=\; \Big[\, \phi_c^{(l^*,T)}(\mathbf{h}_x(\theta)) \,\Big]_{(c,T)\,\in\,\mathcal{C}'\times\mathcal{T}}\;\in\; \mathbb{R}^{k}, \; k \approx 2|\mathcal{C}'||\mathcal{T}|.
\end{equation}
We train a LoRA with geometric supervision by minimizing the loss in Equation~\ref{eq:full_loss}. Each loss term is computed for each plane separately and then averaged over all eight planes.\footnote{For period $T=2$, the sine probe is zero everywhere, so we only supervise the cosine probe, making the effective dimension $k=14$.}

\textbf{Computing the empirical structure and target structure for Fourier features.}
Consider a plane $\phi_c^{(l^*, T)}$ containing the Fourier features for variable $c$ and period $T$.
The empirical centroids $\boldsymbol{\hat\mu}_v(\theta)$ are computed using Equation~\ref{eq:empirical_centroid}, for each value $v$ that $c$ can take over the training distribution (i.e., our training data realizes only a subset of values $\mathcal{V}'_c \subseteq \mathcal{V}_c$). 
The empirical structure is the set of these centroids, \smash{$\mathbf{\hat G}_{(c,T)}(\theta) = \{\boldsymbol{\hat\mu}_v(\theta)\}_{v\in\mathcal{V}'_c} \subset \mathbb{R}^2$}.
For the same $(c,T)$ Fourier plane, the idealized target structure places each value $v$ on a ring: the target position of $v$ on the period-$T$ plane is a point $\mathbf{u}_v$ on the unit circle at phase $2\pi v/T$:
\begin{equation}
\label{eq:ideal_ring}
    \mathbf{u}_v = \bigl(\cos\left(\tfrac{2\pi v}{T}\right),\; \sin\left(\tfrac{2\pi v}{T}\right)\bigr)\in\mathbb{R}^2.
\end{equation}
This ensures that values with the same residue $v \bmod T$ share the same target, e.g., $3\bmod 5$ and $8\bmod5$. A Fourier plane is only defined up to rotation and reflection, so we align the ideal pattern $\{\mathbf{u}_v\}_{v\in\mathcal{V}'_c}$ to the base model's per-value centroids $\{\boldsymbol{\hat\mu}_v(\theta_0)\}_{v \in \mathcal{V}'_c}$ via orthogonal Procrustes alignment \citep{schonemann1966generalized} where the target centroid $\boldsymbol{\mu}_v$ and structure $\mathbf{G}_{(c,T)}$ are computed as:
\begin{equation}
\label{eq:ring_target}
    \boldsymbol{\mu}_v = \boldsymbol{\bar\mu}_c(\theta_0) + \alpha_r\, r_0\, \mathbf{Q}_0\, \mathbf{u}_v,
    \qquad
    \mathbf{G}_{(c,T)} = \{\boldsymbol{\mu}_v\}_{v \in \mathcal{V}'_c} \subset \mathbb{R}^2 
\end{equation}
where $\mathbf{Q}_0\in O(2)$ is a rotation-reflection matrix, and $\alpha_r$ scales the target radius relative to the base model's centroid spread $r_0$. 
We fit $\mathbf{Q}_0$ once before training, and keep the target centroids $\mathbf{G}_{(c,T)}$ fixed throughout.

\textbf{Geometric supervision achieves parity with end-to-end supervision.} As with the task LoRA, we use $X_{\texttt{sum}\leq24}$ as the training set for our geometry LoRA and evaluate on all prompts in $X_{\texttt{sum}\leq 96}$.
The geometry LoRA improves over the base model across the board, achieving a total accuracy of $73.3\%$ (Figure~\ref{fig:months_behavioral_full}, bottom row).
It scores $37.0\%$ higher than the base model and is on-par with the best task LoRA ($71.9\%$) trained on the same data, despite never training with a loss on model outputs. 

\subsection{Sharpened Fourier features alone are not sufficient to improve behavior}
\label{sec:fourier_patching_not_sufficient}
The geometry LoRA's training objective directly supervises the model's representations to produce a similar sharpening effect to the one observed under the task LoRA (Figure~\ref{fig:months-L18-sum-projection}b). However, even though both task and geometry fine-tuning successfully \textit{sharpen} these representations, we find that simply patching these activations back into the base model to ``clean up'' its geometry is not sufficient to recover the gains made by either fine-tuning objective.

Specifically, we test this by replacing the base model's activations with those of a fine-tuned model only within the supervised Fourier planes, i.e., the coordinates under $\mathbf{W}$, for each prompt. For an individual Fourier plane, we write this intervention as: \smash{$\phi_c^{(l,T)}(\mathbf{h}_x(\theta_0)) \leftarrow \phi_c^{(l,T)}(\mathbf{h}_x(\theta))$}. 
When we perform this intervention on the supervised Fourier planes, it has almost no effect on the base model's accuracy (going from $36.3\%$ to $37.3\%$). Patching a random subspace of the same rank has the same null effect.
This result holds for patching from either a task LoRA or a geometry LoRA.
By comparison, patching the $16$ directions along which the fine-tuned and base model activations differ the most, can recover most of the improvement (see Appendix~\ref{appdx:subspace_difference} for details).
This demonstrates that there are beneficial off-target effects of fine-tuning that are not directly captured by the representations we supervise. Patching the geometry of planes from Fourier probes is not sufficient to fix the model's behavior, but a loss that trains the model to fix the geometry is successful.

\section{Supervising Helical Structure for In-Context Linebreaking}
\label{sec:linebreak}
The second task we study is the \textbf{linebreaking task} from \citet{gurnee2025when} where LMs must learn to produce lines of text that wrap according to a maximum width of characters. 
We construct prompts for this task by taking documents from FineWeb~\citep{penedo2024fineweb}, stripping their newlines, and reinserting a newline (\textbackslash\texttt{n}) at the nearest word boundary that will produce a line of at most $m$ characters for $m \in \{1,...,300\}$.\footnote{For small values of $m$ (e.g., $\leq5$), this produces text where a newline is inserted after almost every word.} For a wrapped document, we treat each wrap point (including its preceding lines of text) as a separate prompt $x$.
Given a passage of text where all lines are wrapped to a fixed width $m$, the model must predict a newline at the appropriate width (see Appendix~\ref{appdx:linebreak_dataset} for examples).
A wrap point appearing later in a document is preceded by many earlier lines of the same width, acting as in-context examples of when to linebreak.
This task has a single variable: $\mathcal{C} = \{\texttt{character-count}\}$, where $\texttt{character-count}(x_t) \leq m$ is the number of characters since the previous newline at the token position $t$ of prompt $x$.

\begin{figure}
    \centering
    \includegraphics[width=\linewidth]{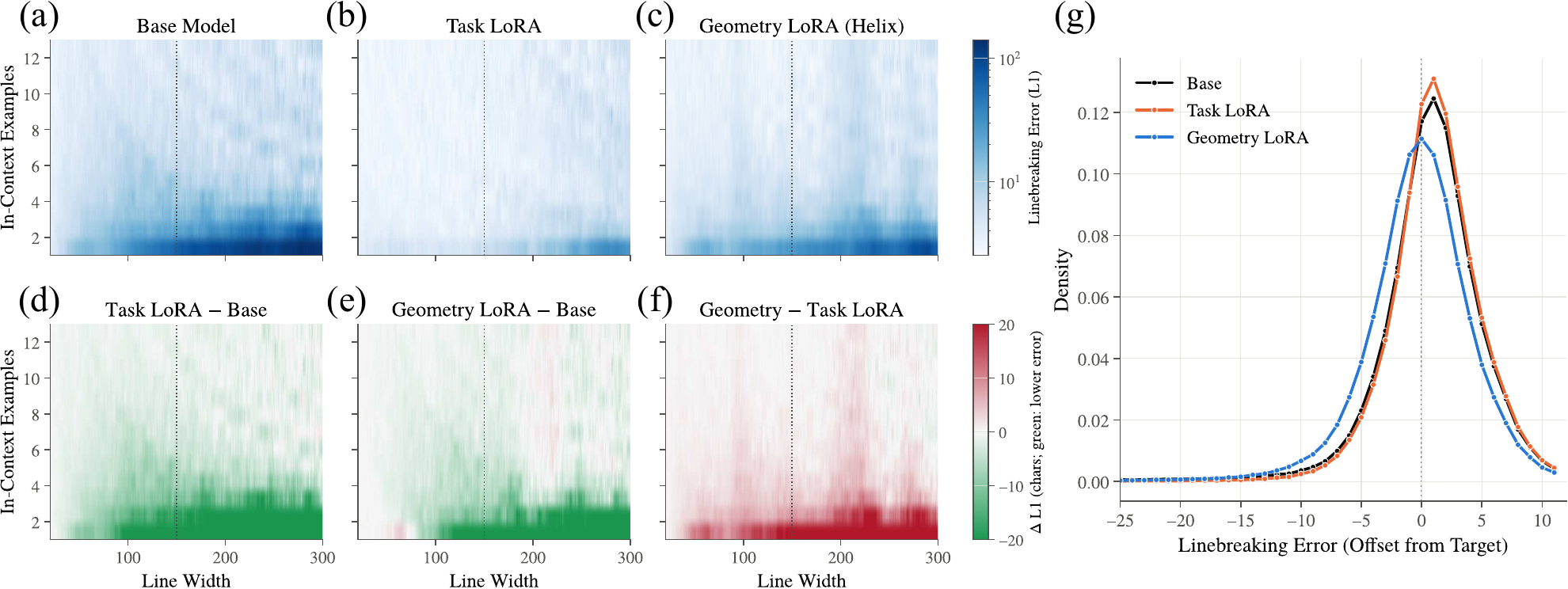}
    \caption{\textbf{Evaluating on the linebreaking task.} \textbf{Top:} Linebreaking error in characters as a function of line width (x-axis) and number of in-context examples (y-axis) for \textbf{(a)} the base Llama-3.1-8B model (mean error of $13.5$ characters), \textbf{(b)} the task LoRA (mean error of $4.6$ characters) and \textbf{(c)} geometry LoRA (mean error of $7.3$ characters). \textbf{Bottom:} Difference in linebreaking error between two models, where green indicates the first model has lower error, while red indicates the second model has lower error.
    \textbf{(d)} The task LoRA and \textbf{(e)} the geometry LoRA improve over the base model at nearly every cell ($97\%$ and $95\%$, respectively), with the strongest improvements in short contexts of long widths. 
    \textbf{(f)} Compared to each other, the task LoRA has lower linebreaking error than the geometry LoRA in most cells, with larger gaps in shorter contexts ($1{-}3$ in-context examples), while the difference in linebreaking error is similar (within ${\sim}0.4$ characters) with five or more in-context examples.
    \textbf{(g)} Distribution of linebreaking error (in characters) of the base model and each fine-tuned variant. Most linebreaking errors are concentrated near the optimal target width.}
    \label{fig:linebreak_behavioral}
\end{figure}

\textbf{Linebreaking error.} To correctly predict a newline at width $m$, the model must keep track of the \texttt{character-count} and predict either words that will lead to a character count less than $m$ or a newline character \textbackslash\texttt{n}. We use $300$ documents wrapped to widths $m \leq 300$ to construct our evaluation set: $X_{m\leq300}$.
We measure performance in terms of \textit{linebreaking error}---which is the number of characters by which a predicted linebreak misses the target width $m$, accounting for the length of the next word. 
A predicted linebreak at token $t$ is considered valid if the sum of the $\texttt{character-count}(x_{<t})$ and the length of the next predicted word $(\operatorname{len}(x_{t+1}))$ is greater than or equal to the target width $m$ (i.e., $\texttt{character-count}(x_{<t}) + \operatorname{len}(x_{t+1}) \geq m$).
For each prompt, we take the model's \emph{first} predicted newline and record its offset from the target width: $\texttt{character-count}(x_{<t}) + \operatorname{len}(x_{t+1}) - m$, pooled over all depths and widths and plot the distributions in Figure~\ref{fig:linebreak_behavioral}g.
Offsets below $0$ are errors where the model linebreaks too early, while offsets ${\geq}0$ are valid (the next word would overflow $m$).

\textbf{Evaluation of task fine-tuning.} We evaluate Llama-3.1-8B on our evaluation dataset $X_{m\leq300}$, and find that 
the base model attains a linebreaking error of $13.5$ characters on average (Figure~\ref{fig:linebreak_behavioral}a). Error is bigger with fewer in-context examples and at larger widths. 
As before, we test whether fine-tuning can improve the base model by training a task LoRA with a cross-entropy loss on the newline prediction task using a training set with widths $m \leq 150$ (i.e., $X_{m\leq150}$; see Appendix~\ref{appdx:linebreak_task_loras} for more details). The task LoRA reduces the base model's linebreaking error to an average of $4.6$ characters (Figure~\ref{fig:linebreak_behavioral}b), and improves error in $97\%$ of cells (Figure~\ref{fig:linebreak_behavioral}d). The largest gains are in short contexts ($1{-}3$ in-context examples), where the linebreaking error drops from $32.9$ to $6.8$ characters.
We find that most linebreak errors for each model are still concentrated near the optimal target width (Figure~\ref{fig:linebreak_behavioral}g; see also Appendix~\ref{appdx:error_distributions}).
The task LoRA (orange) makes its first linebreak prediction too early $29.9\%$ of the time, and the base model (black) is similar, ($33.9\%$ too early).

\subsection{Characterizing the Internal Representation of Character Count}
\label{sec:count_representation}
Next, we ask whether Llama-3.1-8B represents the $\texttt{character-count}$ variable with a helix-like manifold similar to the one reported by \citet{gurnee2025when} in their study of the closed-weight model Claude 3.5 Haiku.
Following \citet{gurnee2025when}, we characterize the structure by: (1) training linear probes $\mathbf{w}^{l}_{\texttt{count}}$ on each layer's residual stream to predict $\texttt{character-count}(x_t)$, 
(2) applying PCA to the per-\texttt{character-count} mean activations to visualize the geometry, and (3) running interventions to test whether the \texttt{character-count} is causally implicated in model behavior.

\begin{figure}
    \centering
    \includegraphics[width=\linewidth]{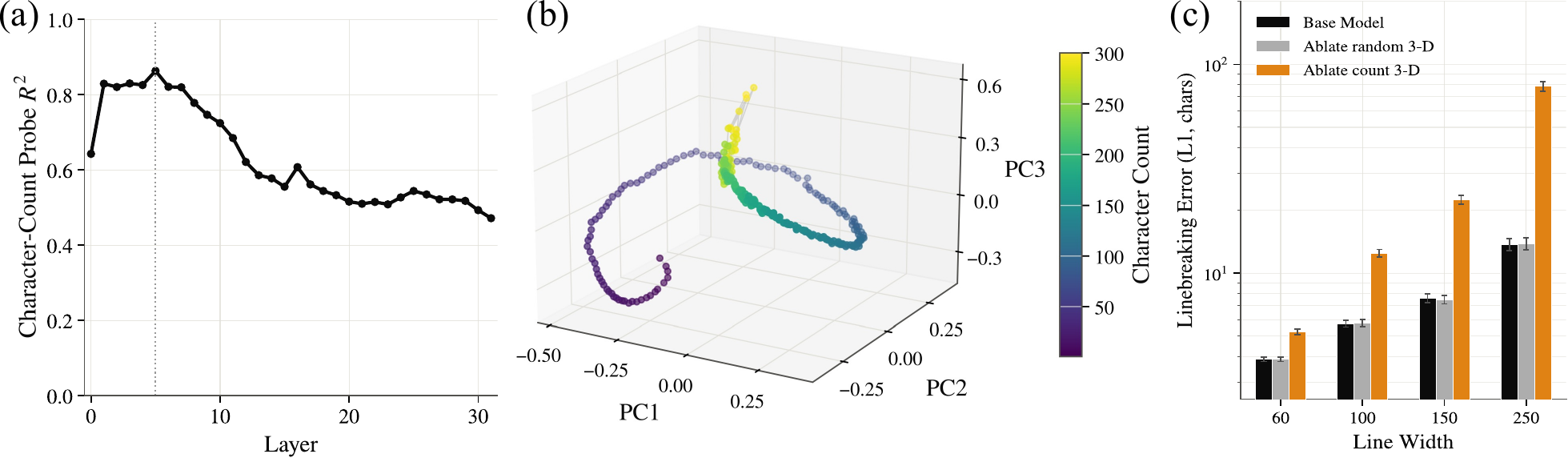}
    \caption{\textbf{Characterizing the internal representation of \texttt{character-count} in Llama-3.1-8B.} \textbf{(a)} Fitting a linear regression probe at each layer to predict \texttt{character-count}. 
    The highest scoring probes occur at early layers, peaking at layer $l=5$ with $R^2=0.86$.
    \textbf{(b)} Average \texttt{character-count} activations projected via PCA follow a helix-like pattern. 
    \textbf{(c)} Ablating the top-$3$ PCA directions that contain the \texttt{character-count} manifold at layer $5$ sharply increases linebreaking error at each width tested.
    Ablating a random $3$-dimensional subspace has no effect on the error.
    }
    \label{fig:linebreak_count_representation}
\end{figure}

We fit linear-regression probes at each layer of Llama-3.1-8B to predict the \texttt{character-count} variable. 
Probes are fit on tokens from FineWeb documents wrapped to widths $m{=}150$ over counts $20$--$150$.
We find that the $\texttt{character-count}$ is linearly decodable, as linear probes achieve high $R^2 = 0.86$ at early layers (peaking at $l{=}5$) while scores drop off in later layers (see Figure~\ref{fig:linebreak_count_representation}a).

We compute the average layer $5$ hidden state activation for each value of $\texttt{character-count}\in[1,150]$ and apply PCA. We find the top $3$ PCs explain $96.8\%$ of the mean's variance. To visualize their structure, we project the average hidden state activations onto the top 3 PCs.
We find that the means lie on a helix-like structure in the first $3$ PCs (Figure~\ref{fig:linebreak_count_representation}b): two components rotate with the \texttt{character-count} variable, while a third axis increases almost monotonically as \texttt{character-count} increases.
To investigate that the \texttt{character-count} representation is relevant for linebreaking, we ablate the top-$3$ PCA components containing the \texttt{character-count} manifold at layer $5$ and test its effect on newline prediction.
For comparison, we also ablate activations within a random $3$-dimensional subspace.
We find that ablating a random $3$-dimensional subspace has little to no effect on the model's performance while ablating the top PCA components increases linebreaking error by a large margin (Figure~\ref{fig:linebreak_count_representation}c; see also Appendix~\ref{appdx:linebreak_manifold}). 

Together, these results suggest that a helix-like structure may serve as a reasonable target geometry for improving the \texttt{character-count} representation, and the model's ability to do linebreaking. We also test other alternative supervision objectives, including using PCA dimensions as the readout in Appendix~\ref{appdx:alternate_linebreak_supervision}. 

\begin{figure}
    \centering
    \includegraphics[width=\linewidth]{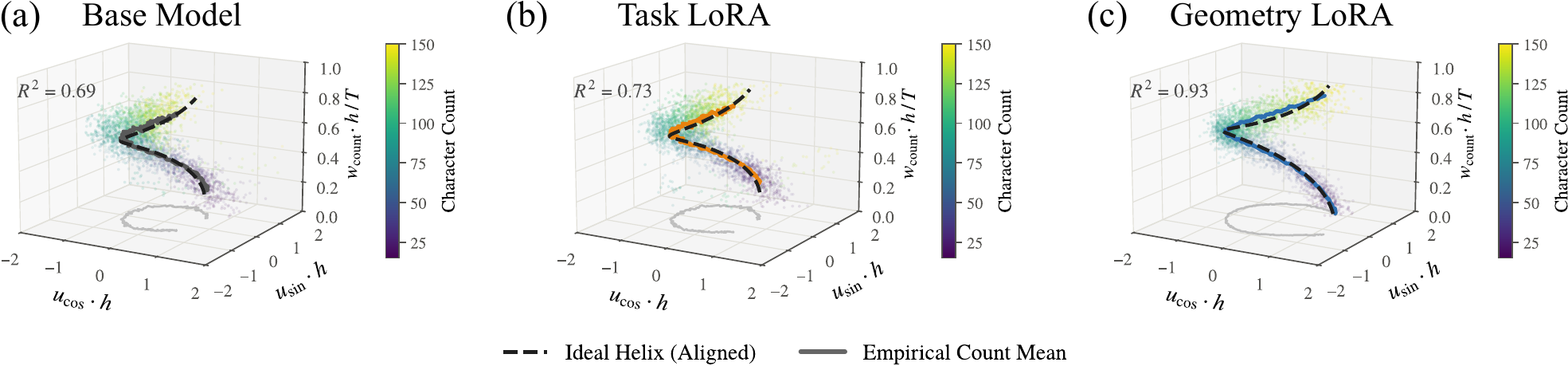}
    \caption{\textbf{The layer-$5$ \texttt{character-count} manifold} under the \textbf{(a)} base model, \textbf{(b)} task LoRA, and \textbf{(c)} geometry LoRA. We measure the fraction of variance an idealized helix captures over all tokens ($R^2$). Plotted points are individual token hidden states projected onto a fixed helix (\S~\ref{sec:linebreak_geometric_alignment}) and colored by $\texttt{character-count} \in [20,150]$. The solid curve in each plot is the empirical per-\texttt{character-count} mean and the dashed curve is the idealized helix. The base model is already well-aligned with an ideal helix ($R^2{=}0.69$), and task fine-tuning sharpens it slightly ($R^2{=}0.73$). Direct geometric supervision aligns the manifold the most ($R^2{=}0.93$).}
    \label{fig:helix_fit}
\end{figure}

\textbf{Effect of task fine-tuning on the count manifold.}
As in the months task (\S\ref{sec:month_sharpening}), we ask whether the task LoRA fixes newline errors by strengthening the geometry the model already has. We fit an idealized helix to the layer-$5$ hidden states of all tokens, aligning each token to the ideal position for its \texttt{character-count} by orthogonal Procrustes---and report the fraction of variance it captures ($R^2$; Appendix~\ref{appdx:linebreak_manifold}). We find that the base model already follows the helix well ($R^2{=}0.69$; Figure~\ref{fig:helix_fit}a), and task fine-tuning has a small positive improvement ($R^2{=}0.73$; Figure~\ref{fig:helix_fit}b).

\subsection{Geometric Alignment on Helical Structure}
\label{sec:linebreak_geometric_alignment}
Similar to Section~\ref{sec:months_geometric_alignment}, we train a geometry LoRA that targets the helical structure directly, without any output-level task loss. 
We apply supervision at layer $l^*{=}5$, where the \texttt{character-count} variable is strongly represented by our linear probe (Figure~\ref{fig:linebreak_count_representation}a).
In this case, the readout $\phi_\text{helix}$ is a $k{=}3$ dimensional \textit{helix}: two dimensions are Fourier probes $\mathbf{w}_{\cos}, \mathbf{w}_{\sin}$ trained with the \texttt{character-count} variable and period $T$, and the third dimension is the \texttt{character-count} probe, \smash{$\mathbf{w}_{\texttt{count}}$}: 
\begin{equation}
    \smash{\phi_\text{helix}(\mathbf{h}_x(\theta)) = \bigl(\mathbf{w}_{\cos}\!\cdot \mathbf{h}_x(\theta) + b_{\cos},\;
    \mathbf{w}_{\sin}\!\cdot \mathbf{h}_x(\theta) + b_{\sin},\;
    \mathbf{w}_{\texttt{count}}\!\cdot \mathbf{h}_x(\theta) + b_{\texttt{count}}\bigr)}\in\mathbb{R}^{3}.
\end{equation}
The empirical structure \smash{$\mathbf{\hat G}_\text{helix}(\theta) = \{\boldsymbol{\hat\mu}_v(\theta)\}_{v \in \mathcal{V}'_{\texttt{count}}}$} is the set of per-\texttt{character-count} centroids (Equation~\ref{eq:empirical_centroid}), and the idealized target structure $\mathbf{G}_\text{helix} = \{\boldsymbol{\mu}_v\}_{v \in \mathcal{V}'_{\texttt{count}}}$ places each \texttt{character-count} value $v$ on the ideal helix point $\mathbf{u}_v = (\cos(2\pi v/T),\ \sin(2\pi v/T),\ v/T)$. We align an ideal helix to match the orientation of the base model's per-\texttt{character-count} centroids $\{\boldsymbol{\hat\mu}_v(\theta_0)\}$ via orthogonal Procrustes once before training (see Equation~\ref{eq:ring_target}) and keep $\mathbf{G}_\text{helix}$ fixed throughout. 
We train the geometry LoRA by minimizing the geometric loss of Equation~\ref{eq:full_loss} over the same width range as the task LoRA ($m\leq150$; see Appendix~\ref{appdx:helix_geometry_lora} for more details).

\textbf{Evaluating geometric supervision for the linebreaking task.} As seen in Figure~\ref{fig:linebreak_behavioral}c, we find that the geometry LoRA is able to improve the base model's linebreaking error from $13.5$ to an average of $7.3$ characters, improving over the base model in $95\%$ of (width $\times$ in-context) cells (Figure~\ref{fig:linebreak_behavioral}e). 
The geometry LoRA does have higher error overall compared to the task LoRA, but converges to similar performance with more in-context examples---a difference in linebreaking error of only $0.4$ characters for contexts with five or more examples (Figure~\ref{fig:linebreak_behavioral}f).
We also see that the incorrect linebreak predictions the geometry LoRA makes are concentrated near the optimal target width (Figure~\ref{fig:linebreak_behavioral}g).
In this setting, the geometry LoRA reshapes the \texttt{character-count} manifold more than task supervision (as expected), aligning to the ideal helix at $R^2{=}0.93$ vs.\ $0.73$ for the task LoRA and $0.69$ for the base model (Figure~\ref{fig:helix_fit}c). 

\textbf{Patching supervised helix representations is not sufficient.} Similar to \S\ref{sec:fourier_patching_not_sufficient}, we test whether patching a ``cleaned up'' helix-like structure from the fine-tuned model can account for its decrease in linebreaking error.
We find that patching activations within the supervised helix from either fine-tuned model (task LoRA or geometry LoRA) into the base model has little effect on the base model's linebreaking error, dropping from $13.5$ to $12.9$ in both cases.
However, patching in activations within a low-dimensional subspace or from later layers can recover much of the improvement achieved by fine-tuning with the geometry LoRA (see Appendix~\ref{appdx:helix_patching} for details).

\section{Related Work}
\label{sec:related_work}

\textbf{Neural geometry of arithmetic and counting.} Our work builds on a series of mechanistic accounts describing circular Fourier features or helical representations that enable numeric and positional computation in both small neural networks~\citep{nanda2023progress, zhong2023pizza, morwani2024feature, furuta2024towards, marchetti2026sequential} and pretrained LLMs~\citep{zhou2024fourier, engels2025language, levy2025language, kantamneni2025trigonometry, zhou2025fone, hu2026understanding, fu2026convergent, kadlcik2025pretrained, zhu2025language, feucht2026arithmetic, gurnee2025when}.
Our work is also related to \citet{sinii2026chasing} who identify helical counting manifolds in additional open-weight models, and \citet{dunefsky2026shared} who study a countdown task that shares similar helical circuitry to the linebreaking task.
Recent work shows that this geometry arises from symmetries and correlations in the training data~\citep{karkada2026symmetry, prieto2026from, korchinski2025emergence, modell2025origins, nishi2025representation}, and that moving activations along them changes behavior in predictable ways~\citep{wurgaft2026manifold, sarfati2026shape}.
While these prior works provide descriptions of neural geometric structure, we investigate whether their idealization can be used as a sufficient learning signal for improving model behavior without output supervision.

\textbf{Supervision using model internals.} 
Our work builds on several lines of work inspired by a long-standing goal of interpretability---that mechanistic understanding of model computations can inform how models are trained or updated~\citep{bau2020rewriting, meng2022locating, bergen2026anatomy}.
Closest to our approach are methods that define a loss directly on internal computation: \citet{cho2025mechanisticfinetuning} supervise attention scores toward idealized induction-head behavior and \citet{akyurek2024incontext} augment existing LLMs with n-gram heads with similar idealized attention patterns. \citet{bai2025why} show that an auxiliary loss supervising representations to contain intermediate sums lets transformers learn multi-digit multiplication, 
and \citet{zou2023representation} train LoRAs with a loss computed on steered hidden states. 
In our work, the supervision target is a fixed geometry taken from a prior descriptive account of the base model, and we train without any output-level loss.

\textbf{Imposing structure on neural networks.} Our work is also related to methods that enforce symmetries through architecture design, such as equivariant networks~\citep{bronstein2021geometricdeeplearninggrids, cohen2018spherical, wang2021incorporating, brehmer2023geometricalgebratransformer} or through physical or equivariance constraints~\citep{raissi2019physics, finzi2021residual}.  
In our work, we regularize an intermediate representation toward a target geometry constraint constructed from a mechanistic account of a task variable.
Several other methods impose internal structure by supervising the outputs of intervened models to align internal variables with a causal model~\citep{geiger2021iit, wu2021distill, wu2023causalproxy, zur2024updating}, including inducing character-level structure in subword models~\citep{huang2023inducing}.
Similarly, concept bottleneck models route predictions through supervised intermediate concepts~\citep{koh2020concept}.
Another set of methods localize \textit{where} to update selected representations or circuit components with a task loss~\citep{meng2022locating, wu2024reft, yin2024lofitlocalizedfinetuningllm, prakash2025cca} or intervene on concept directions during training~\citep{casademunt2026steering, chen2025personavectorsmonitoringcontrolling}, though \citet{hase2023localization} argue that localization does not always coincide with the best edit location.

\textbf{Causal evaluation.} The mechanistic accounts we build on rely on frameworks such as causal mediation and causal abstraction~\citep{pearl2001directindirecteffects, vig2020investigating, finlayson2021causal, stolfo2023mechanistic, todd2024function, geiger2021causal, geiger2026causal, mueller2026quest}, which evaluate proposed mechanisms through targeted interventions on internal states, including learned subspaces or probes~\citep{geiger2023das, wu2023interpretability, prakash2026lookback, todd2026incontext}.
Our finding that geometric supervision has beneficial effects beyond the targeted representations (\S\ref{sec:fourier_patching_not_sufficient}) is reminiscent of prior work cautioning that probe accuracy alone does not establish that a model uses the decoded information~\citep{belinkov2022probing, ravichander2021probing, elazar2020amnesicprobing}, or that activation patching can produce illusory effects~\citep{makelov2023illusions, wu2024replyillusions} or fall outside the model's natural distribution of activations~\citep{grant2026addressing}.

\section{Discussion}
\label{sec:discussion}
We have studied two cases where prior work identified causally-relevant geometric structure that an LLM uses to compute its answer: Fourier features for month arithmetic~\citep{feucht2026arithmetic}, and helical representations for predicting linebreaks in structured text~\citep{gurnee2025when}. 
While previous mechanistic accounts of these variables describe an idealized circular or helix-like structure, the model's actual representations are noisier.
Prior work suggests that representational geometry is shaped by the statistics of the training data~\citep{karkada2026symmetry, korchinski2025emergence}, so we might expect the geometry to be most distorted, and errors most frequent, for rarely observed values.
This is consistent with where we observe most errors (large month offsets and longer line widths). 
We asked whether training these representations toward their ideal geometry can improve the model's behavior without directly supervising it.
We found that a loss defined only on the geometry of an intermediate representation, without output-level supervision, can achieve similar performance as standard LoRA fine-tuning on these tasks. These results provide a functional validation of the mechanistic accounts we build on.
We also find that simply patching in an ideal geometry does not match fine-tuning, suggesting geometric supervision has beneficial effects beyond the representations it targets.
Our findings show that behavioral errors on these tasks can be corrected by specifying a model's representational structure, rather than its outputs. This suggests that detailed mechanistic accounts of a particular behavior might be useful not only to gain understanding, but also to inform targeted training for more surgical control.

A major goal of interpretability research is that understanding a neural network should allow us to predict when it will fail to generalize, and to surgically edit its representations or weights to fix those errors.
While this is a hard problem in general, the tasks we study are easy to verify and have plentiful training data, which allowed us to compare geometric supervision directly against output-level supervision.
More broadly, we expect neural geometry to be helpful in settings where output supervision is scarce, expensive, or hard to specify, such as for scientific foundation models trained on biological data, where model outputs are not natural language, but the domain has rich structure.
Extending geometric supervision to more settings will require methods for automatically discovering causally-relevant target geometries, choosing idealizations, and mechanistically characterizing the task variables involved.

\section*{Disclosure of AI Use}
We used generative AI tools to help write and organize experiment code for all experiments, and to obtain feedback on experiments.
AI was used to help write plotting code used to generate some figures for the paper, and to write code for building the synthetic linebreaking dataset described in (\S~\ref{sec:linebreak}).
AI was also used to help draft portions of the appendix, check definitions and equations, and help polish writing in the main paper. All AI suggestions went through several additional editing rounds. We have reviewed all AI contributions to this work and take responsibility for its final content, including its text, methods, and results.

\bibliography{references}

\begin{thebibliography}{81}
\providecommand{\natexlab}[1]{#1}
\providecommand{\url}[1]{\texttt{#1}}
\expandafter\ifx\csname urlstyle\endcsname\relax
  \providecommand{\doi}[1]{doi: #1}\else
  \providecommand{\doi}{doi: \begingroup \urlstyle{rm}\Url}\fi

\bibitem[Aky{\"u}rek et~al.(2024)Aky{\"u}rek, Wang, Kim, and
  Andreas]{akyurek2024incontext}
Ekin Aky{\"u}rek, Bailin Wang, Yoon Kim, and Jacob Andreas.
\newblock In-context language learning: Architectures and algorithms.
\newblock In \emph{Forty-first International Conference on Machine Learning},
  2024.
\newblock URL \url{https://openreview.net/forum?id=3Z9CRr5srL}.

\bibitem[Bai et~al.(2025)Bai, Pres, Deng, Tan, Shieber, Viégas, Wattenberg,
  and Lee]{bai2025why}
Xiaoyan Bai, Itamar Pres, Yuntian Deng, Chenhao Tan, Stuart Shieber, Fernanda
  Viégas, Martin Wattenberg, and Andrew Lee.
\newblock Why can't transformers learn multiplication? reverse-engineering
  reveals long-range dependency pitfalls, 2025.
\newblock URL \url{https://arxiv.org/abs/2510.00184}.

\bibitem[Bau et~al.(2020)Bau, Liu, Wang, Zhu, and Torralba]{bau2020rewriting}
David Bau, Steven Liu, Tongzhou Wang, Jun-Yan Zhu, and Antonio Torralba.
\newblock Rewriting a deep generative model.
\newblock In Andrea Vedaldi, Horst Bischof, Thomas Brox, and Jan-Michael Frahm
  (eds.), \emph{Computer Vision -- ECCV 2020}, pp.\  351--369, Cham, 2020.
  Springer International Publishing.
\newblock ISBN 978-3-030-58452-8.
\newblock URL \url{https://doi.org/10.1007/978-3-030-58452-8\_21}.

\bibitem[Belinkov(2022)]{belinkov2022probing}
Yonatan Belinkov.
\newblock Probing classifiers: Promises, shortcomings, and advances.
\newblock \emph{Computational Linguistics}, 48\penalty0 (1):\penalty0 207--219,
  March 2022.
\newblock \doi{10.1162/coli_a_00422}.
\newblock URL \url{https://aclanthology.org/2022.cl-1.7/}.

\bibitem[Bergen et~al.(2026)Bergen, Bhalla, Baskaran, Loeffler, Sarfati, Gala,
  Panwar, Aranguri, Fel, Geiger, Kowal, Boppana, Balsam, Lewis, Merullo,
  McGrath, and Lubana]{bergen2026anatomy}
Leon Bergen, Usha Bhalla, Sidharth Baskaran, Max Loeffler, Raphael Sarfati,
  Dhruvil Gala, Ryan Panwar, Santiago Aranguri, Thomas Fel, Atticus Geiger,
  Matthew Kowal, Siddharth Boppana, Daniel Balsam, Owen Lewis, Jack Merullo,
  Thomas McGrath, and Ekdeep~Singh Lubana.
\newblock Anatomy of post-training: Using interpretability to characterize data
  and shape the learning signal, 2026.
\newblock URL \url{https://arxiv.org/abs/2606.12360}.

\bibitem[Bhalla et~al.(2026)Bhalla, Fel, Rager, Feucht, Haklay, Wurgaft,
  Boppana, Kowal, Merullo, Geiger, and Lubana]{bhalla2026do}
Usha Bhalla, Thomas Fel, Can Rager, Sheridan Feucht, Tal Haklay, Daniel
  Wurgaft, Siddharth Boppana, Matthew Kowal, Jack Merullo, Atticus Geiger, and
  Ekdeep~Singh Lubana.
\newblock Do sparse autoencoders capture concept manifolds?
\newblock In \emph{Third Conference on Language Modeling}, 2026.
\newblock URL \url{https://openreview.net/forum?id=euoAoOflaN}.

\bibitem[Brehmer et~al.(2023)Brehmer, de~Haan, Behrends, and
  Cohen]{brehmer2023geometricalgebratransformer}
Johann Brehmer, Pim de~Haan, S\"{o}nke Behrends, and Taco Cohen.
\newblock Geometric algebra transformer.
\newblock In A.~Oh, T.~Naumann, A.~Globerson, K.~Saenko, M.~Hardt, and
  S.~Levine (eds.), \emph{Advances in Neural Information Processing Systems},
  volume~36, pp.\  35472--35496. Curran Associates, Inc., 2023.
\newblock \doi{10.52202/075280-1541}.
\newblock URL
  \url{https://proceedings.neurips.cc/paper_files/paper/2023/file/6f6dd92b03ff9be7468a6104611c9187-Paper-Conference.pdf}.

\bibitem[Bronstein et~al.(2021)Bronstein, Bruna, Cohen, and
  Veličković]{bronstein2021geometricdeeplearninggrids}
Michael~M. Bronstein, Joan Bruna, Taco Cohen, and Petar Veličković.
\newblock Geometric deep learning: Grids, groups, graphs, geodesics, and
  gauges, 2021.
\newblock URL \url{https://arxiv.org/abs/2104.13478}.

\bibitem[Casademunt et~al.(2026)Casademunt, Juang, Karvonen, Marks,
  Rajamanoharan, and Nanda]{casademunt2026steering}
Helena Casademunt, Caden Juang, Adam Karvonen, Samuel Marks, Senthooran
  Rajamanoharan, and Neel Nanda.
\newblock Steering out-of-distribution generalization with concept ablation
  fine-tuning.
\newblock In \emph{Forty-third International Conference on Machine Learning},
  2026.
\newblock URL \url{https://openreview.net/forum?id=yuJehN4oMW}.

\bibitem[Chen et~al.(2025)Chen, Arditi, Sleight, Evans, and
  Lindsey]{chen2025personavectorsmonitoringcontrolling}
Runjin Chen, Andy Arditi, Henry Sleight, Owain Evans, and Jack Lindsey.
\newblock Persona vectors: Monitoring and controlling character traits in
  language models, 2025.
\newblock URL \url{https://arxiv.org/abs/2507.21509}.

\bibitem[Cho et~al.(2025)Cho, Luo, Kato, Kaenbyou, and
  Inoue]{cho2025mechanisticfinetuning}
Hakaze Cho, Peng Luo, Mariko Kato, Rin Kaenbyou, and Naoya Inoue.
\newblock Mechanistic fine-tuning for in-context learning.
\newblock In Yonatan Belinkov, Aaron Mueller, Najoung Kim, Hosein Mohebbi,
  Hanjie Chen, Dana Arad, and Gabriele Sarti (eds.), \emph{Proceedings of the
  8th BlackboxNLP Workshop: Analyzing and Interpreting Neural Networks for
  NLP}, pp.\  330--357, Suzhou, China, November 2025. Association for
  Computational Linguistics.
\newblock ISBN 979-8-89176-346-3.
\newblock \doi{10.18653/v1/2025.blackboxnlp-1.21}.
\newblock URL \url{https://aclanthology.org/2025.blackboxnlp-1.21/}.

\bibitem[Cobbe et~al.(2021)Cobbe, Kosaraju, Bavarian, Chen, Jun, Kaiser,
  Plappert, Tworek, Hilton, Nakano, Hesse, and
  Schulman]{cobbe2021trainingverifierssolvemath}
Karl Cobbe, Vineet Kosaraju, Mohammad Bavarian, Mark Chen, Heewoo Jun, Lukasz
  Kaiser, Matthias Plappert, Jerry Tworek, Jacob Hilton, Reiichiro Nakano,
  Christopher Hesse, and John Schulman.
\newblock Training verifiers to solve math word problems, 2021.
\newblock URL \url{https://arxiv.org/abs/2110.14168}.

\bibitem[Cohen et~al.(2018)Cohen, Geiger, Köhler, and
  Welling]{cohen2018spherical}
Taco~S. Cohen, Mario Geiger, Jonas Köhler, and Max Welling.
\newblock Spherical {CNN}s.
\newblock In \emph{International Conference on Learning Representations}, 2018.
\newblock URL \url{https://openreview.net/forum?id=Hkbd5xZRb}.

\bibitem[Dunefsky et~al.(2026)Dunefsky, Gurnee, and
  Ameisen]{dunefsky2026shared}
Jacob Dunefsky, Wes Gurnee, and Emmanuel Ameisen.
\newblock A shared subcircuit lets llms count down across tasks, 2026.
\newblock URL \url{https://arxiv.org/abs/2607.12279}.

\bibitem[Elazar et~al.(2021)Elazar, Ravfogel, Jacovi, and
  Goldberg]{elazar2020amnesicprobing}
Yanai Elazar, Shauli Ravfogel, Alon Jacovi, and Yoav Goldberg.
\newblock Amnesic probing: Behavioral explanation with amnesic counterfactuals.
\newblock \emph{Transactions of the Association for Computational Linguistics},
  9:\penalty0 160--175, 03 2021.
\newblock ISSN 2307-387X.
\newblock \doi{10.1162/tacl_a_00359}.
\newblock URL \url{https://doi.org/10.1162/tacl_a_00359}.

\bibitem[Engels et~al.(2025)Engels, Michaud, Liao, Gurnee, and
  Tegmark]{engels2025language}
Joshua Engels, Eric~J Michaud, Isaac Liao, Wes Gurnee, and Max Tegmark.
\newblock Not all language model features are one-dimensionally linear.
\newblock In \emph{The Thirteenth International Conference on Learning
  Representations}, 2025.
\newblock URL \url{https://openreview.net/forum?id=d63a4AM4hb}.

\bibitem[Feucht et~al.(2026)Feucht, Haklay, Bhalla, Wurgaft, Rager, Merullo,
  McGrath, Sarfati, Lewis, Lubana, Fel, and Geiger]{feucht2026arithmetic}
Sheridan Feucht, Tal Haklay, Usha Bhalla, Daniel Wurgaft, Can Rager, Jack
  Merullo, Thomas McGrath, Rapha{\"e}l Sarfati, Owen Lewis, Ekdeep~Singh
  Lubana, Thomas Fel, and Atticus Geiger.
\newblock Arithmetic in the wild: Llama uses standard addition to reason about
  cyclic concepts.
\newblock In \emph{Third Conference on Language Modeling}, 2026.
\newblock URL \url{https://openreview.net/forum?id=92qkFSqoqk}.

\bibitem[Finlayson et~al.(2021)Finlayson, Mueller, Gehrmann, Shieber, Linzen,
  and Belinkov]{finlayson2021causal}
Matthew Finlayson, Aaron Mueller, Sebastian Gehrmann, Stuart Shieber, Tal
  Linzen, and Yonatan Belinkov.
\newblock Causal analysis of syntactic agreement mechanisms in neural language
  models.
\newblock In \emph{Proceedings of the 59th Annual Meeting of the Association
  for Computational Linguistics and the 11th International Joint Conference on
  Natural Language Processing (Volume 1: Long Papers)}, pp.\  1828--1843.
  Association for Computational Linguistics, 2021.
\newblock \doi{10.18653/v1/2021.acl-long.144}.
\newblock URL \url{https://aclanthology.org/2021.acl-long.144}.

\bibitem[Finzi et~al.(2021)Finzi, Benton, and Wilson]{finzi2021residual}
Marc~Anton Finzi, Gregory Benton, and Andrew~Gordon Wilson.
\newblock Residual pathway priors for soft equivariance constraints.
\newblock In A.~Beygelzimer, Y.~Dauphin, P.~Liang, and J.~Wortman Vaughan
  (eds.), \emph{Advances in Neural Information Processing Systems}, 2021.
\newblock URL \url{https://openreview.net/forum?id=k505ekjMzww}.

\bibitem[Fu et~al.(2026)Fu, Zhou, Belkin, Sharan, and Jia]{fu2026convergent}
Deqing Fu, Tianyi Zhou, Mikhail Belkin, Vatsal Sharan, and Robin Jia.
\newblock Convergent evolution: How different language models learn similar
  number representations.
\newblock In \emph{Third Conference on Language Modeling}, 2026.
\newblock URL \url{https://openreview.net/forum?id=RIoMmOkZ8E}.

\bibitem[Furuta et~al.(2024)Furuta, Minegishi, Iwasawa, and
  Matsuo]{furuta2024towards}
Hiroki Furuta, Gouki Minegishi, Yusuke Iwasawa, and Yutaka Matsuo.
\newblock Towards empirical interpretation of internal circuits and properties
  in grokked transformers on modular polynomials.
\newblock \emph{Transactions on Machine Learning Research}, 2024.
\newblock ISSN 2835-8856.
\newblock URL \url{https://openreview.net/forum?id=MzSf70uXJO}.

\bibitem[Geiger et~al.(2021)Geiger, Lu, Icard, and Potts]{geiger2021causal}
Atticus Geiger, Hanson Lu, Thomas Icard, and Christopher Potts.
\newblock Causal abstractions of neural networks.
\newblock In \emph{Advances in Neural Information Processing Systems},
  volume~34, pp.\  9574--9586, 2021.
\newblock URL \url{https://arxiv.org/abs/2106.02997}.

\bibitem[Geiger et~al.(2022)Geiger, Wu, Lu, Rozner, Kreiss, Icard, Goodman, and
  Potts]{geiger2021iit}
Atticus Geiger, Zhengxuan Wu, Hanson Lu, Josh Rozner, Elisa Kreiss, Thomas
  Icard, Noah Goodman, and Christopher Potts.
\newblock Inducing causal structure for interpretable neural networks.
\newblock In Kamalika Chaudhuri, Stefanie Jegelka, Le~Song, Csaba Szepesvari,
  Gang Niu, and Sivan Sabato (eds.), \emph{Proceedings of the 39th
  International Conference on Machine Learning}, volume 162 of
  \emph{Proceedings of Machine Learning Research}, pp.\  7324--7338. PMLR,
  17--23 Jul 2022.
\newblock URL \url{https://proceedings.mlr.press/v162/geiger22a.html}.

\bibitem[Geiger et~al.(2024)Geiger, Wu, Potts, Icard, and
  Goodman]{geiger2023das}
Atticus Geiger, Zhengxuan Wu, Christopher Potts, Thomas Icard, and Noah
  Goodman.
\newblock Finding alignments between interpretable causal variables and
  distributed neural representations.
\newblock In Francesco Locatello and Vanessa Didelez (eds.), \emph{Proceedings
  of the Third Conference on Causal Learning and Reasoning}, volume 236 of
  \emph{Proceedings of Machine Learning Research}, pp.\  160--187. PMLR, 01--03
  Apr 2024.
\newblock URL \url{https://proceedings.mlr.press/v236/geiger24a.html}.

\bibitem[Geiger et~al.(2025)Geiger, Ibeling, Zur, Chaudhary, Chauhan, Huang,
  Arora, Wu, Goodman, Potts, and Icard]{geiger2026causal}
Atticus Geiger, Duligur Ibeling, Amir Zur, Maheep Chaudhary, Sonakshi Chauhan,
  Jing Huang, Aryaman Arora, Zhengxuan Wu, Noah Goodman, Christopher Potts, and
  Thomas Icard.
\newblock Causal abstraction: A theoretical foundation for mechanistic
  interpretability.
\newblock \emph{Journal of Machine Learning Research}, 26\penalty0
  (83):\penalty0 1--64, 2025.
\newblock URL \url{https://jmlr.org/papers/v26/23-0058.html}.

\bibitem[Grant et~al.(2026)Grant, Han, Tartaglini, and
  Potts]{grant2026addressing}
Satchel Grant, Simon~Jerome Han, Alexa Tartaglini, and Christopher Potts.
\newblock Addressing divergent representations from causal interventions on
  neural networks.
\newblock In C.~Vondrick, B.~Hariharan, C.~Raffel, L.~Pinto, D.~Yang, and
  A.~Faust (eds.), \emph{International Conference on Learning Representations},
  volume 2026, pp.\  11330--11356, 2026.
\newblock URL
  \url{https://proceedings.iclr.cc/paper_files/paper/2026/file/133e588e1429f9f1e25b215da145580e-Paper-Conference.pdf}.

\bibitem[Gurnee et~al.(2025)Gurnee, Ameisen, Kauvar, Tarng, Pearce, Olah, and
  Batson]{gurnee2025when}
Wes Gurnee, Emmanuel Ameisen, Isaac Kauvar, Julius Tarng, Adam Pearce, Chris
  Olah, and Joshua Batson.
\newblock When models manipulate manifolds: The geometry of a counting task.
\newblock \emph{Transformer Circuits Thread}, 2025.
\newblock URL
  \url{https://transformer-circuits.pub/2025/linebreaks/index.html}.

\bibitem[Hase et~al.(2023)Hase, Bansal, Kim, and
  Ghandeharioun]{hase2023localization}
Peter Hase, Mohit Bansal, Been Kim, and Asma Ghandeharioun.
\newblock Does localization inform editing? surprising differences in
  causality-based localization vs. knowledge editing in language models.
\newblock In \emph{Advances in Neural Information Processing Systems},
  volume~36, pp.\  17643--17668. Curran Associates, Inc., 2023.
\newblock \doi{10.52202/075280-0774}.
\newblock URL
  \url{https://papers.nips.cc/paper_files/paper/2023/hash/3927bbdcf0e8d1fa8aa23c26f358a281-Abstract-Conference.html}.

\bibitem[Hendrycks et~al.(2021)Hendrycks, Burns, Basart, Zou, Mazeika, Song,
  and Steinhardt]{hendrycks2021measuring}
Dan Hendrycks, Collin Burns, Steven Basart, Andy Zou, Mantas Mazeika, Dawn
  Song, and Jacob Steinhardt.
\newblock Measuring massive multitask language understanding.
\newblock In \emph{International Conference on Learning Representations}, 2021.
\newblock URL \url{https://openreview.net/forum?id=d7KBjmI3GmQ}.

\bibitem[Hu et~al.(2022)Hu, Shen, Wallis, Allen-Zhu, Li, Wang, Wang, and
  Chen]{hu2022lora}
Edward~J Hu, Yelong Shen, Phillip Wallis, Zeyuan Allen-Zhu, Yuanzhi Li, Shean
  Wang, Lu~Wang, and Weizhu Chen.
\newblock Lo{RA}: Low-rank adaptation of large language models.
\newblock In \emph{International Conference on Learning Representations}, 2022.
\newblock URL \url{https://openreview.net/forum?id=nZeVKeeFYf9}.

\bibitem[Hu et~al.(2026)Hu, Yin, Jordan, Steinhardt, and
  Chen]{hu2026understanding}
Xinyan Hu, Kayo Yin, Michael~I. Jordan, Jacob Steinhardt, and Lijie Chen.
\newblock Understanding in-context learning of addition via activation
  subspaces.
\newblock In \emph{Third Conference on Language Modeling}, 2026.
\newblock URL \url{https://openreview.net/forum?id=7QTqrbW27Z}.

\bibitem[Huang et~al.(2023)Huang, Wu, Mahowald, and Potts]{huang2023inducing}
Jing Huang, Zhengxuan Wu, Kyle Mahowald, and Christopher Potts.
\newblock Inducing character-level structure in subword-based language models
  with type-level interchange intervention training.
\newblock In Anna Rogers, Jordan Boyd-Graber, and Naoaki Okazaki (eds.),
  \emph{Findings of the {A}ssociation for {C}omputational {L}inguistics: {ACL}
  2023}, pp.\  12163--12180, Toronto, Canada, July 2023. Association for
  Computational Linguistics.
\newblock \doi{10.18653/v1/2023.findings-acl.770}.
\newblock URL \url{https://aclanthology.org/2023.findings-acl.770/}.

\bibitem[Jain et~al.(2024)Jain, Kirk, Lubana, Dick, Tanaka, Rockt{\"a}schel,
  Grefenstette, and Krueger]{jain2024mechanistically}
Samyak Jain, Robert Kirk, Ekdeep~Singh Lubana, Robert~P. Dick, Hidenori Tanaka,
  Tim Rockt{\"a}schel, Edward Grefenstette, and David Krueger.
\newblock Mechanistically analyzing the effects of fine-tuning on procedurally
  defined tasks.
\newblock In \emph{The Twelfth International Conference on Learning
  Representations}, 2024.
\newblock URL \url{https://openreview.net/forum?id=A0HKeKl4Nl}.

\bibitem[Kadl{\v{c}}{\'i}k et~al.(2025)Kadl{\v{c}}{\'i}k, {\v{S}}tef{\'a}nik,
  Mickus, Kucha{\v{r}}, and Spiegel]{kadlcik2025pretrained}
Marek Kadl{\v{c}}{\'i}k, Michal {\v{S}}tef{\'a}nik, Timothee Mickus, Josef
  Kucha{\v{r}}, and Michal Spiegel.
\newblock Pre-trained language models learn remarkably accurate representations
  of numbers.
\newblock In Christos Christodoulopoulos, Tanmoy Chakraborty, Carolyn Rose, and
  Violet Peng (eds.), \emph{Proceedings of the 2025 Conference on Empirical
  Methods in Natural Language Processing}, pp.\  26705--26714, Suzhou, China,
  November 2025. Association for Computational Linguistics.
\newblock ISBN 979-8-89176-332-6.
\newblock \doi{10.18653/v1/2025.emnlp-main.1356}.
\newblock URL \url{https://aclanthology.org/2025.emnlp-main.1356/}.

\bibitem[Kantamneni \& Tegmark(2025)Kantamneni and
  Tegmark]{kantamneni2025trigonometry}
Subhash Kantamneni and Max Tegmark.
\newblock Language models use trigonometry to do addition, 2025.
\newblock URL \url{https://arxiv.org/abs/2502.00873}.

\bibitem[Karkada et~al.(2026)Karkada, Korchinski, Nava, Wyart, and
  Bahri]{karkada2026symmetry}
Dhruva Karkada, Daniel~James Korchinski, Andres Nava, Matthieu Wyart, and
  Yasaman Bahri.
\newblock Symmetry in language statistics shapes the geometry of model
  representations.
\newblock In \emph{Forty-third International Conference on Machine Learning},
  2026.
\newblock URL \url{https://openreview.net/forum?id=XQeMPEkfdd}.

\bibitem[Koh et~al.(2020)Koh, Nguyen, Tang, Mussmann, Pierson, Kim, and
  Liang]{koh2020concept}
Pang~Wei Koh, Thao Nguyen, Yew~Siang Tang, Stephen Mussmann, Emma Pierson, Been
  Kim, and Percy Liang.
\newblock Concept bottleneck models.
\newblock In \emph{Proceedings of the 37th International Conference on Machine
  Learning}, volume 119 of \emph{Proceedings of Machine Learning Research},
  pp.\  5338--5348. PMLR, 2020.
\newblock URL \url{https://proceedings.mlr.press/v119/koh20a.html}.

\bibitem[Korchinski et~al.(2025)Korchinski, Karkada, Bahri, and
  Wyart]{korchinski2025emergence}
Daniel~James Korchinski, Dhruva Karkada, Yasaman Bahri, and Matthieu Wyart.
\newblock On the emergence of linear analogies in word embeddings.
\newblock In \emph{The Thirty-ninth Annual Conference on Neural Information
  Processing Systems}, 2025.
\newblock URL \url{https://openreview.net/forum?id=VV1HfpDCF1}.

\bibitem[Levy \& Geva(2025)Levy and Geva]{levy2025language}
Amit~Arnold Levy and Mor Geva.
\newblock Language models encode numbers using digit representations in base
  10.
\newblock In Luis Chiruzzo, Alan Ritter, and Lu~Wang (eds.), \emph{Proceedings
  of the 2025 Conference of the Nations of the Americas Chapter of the
  Association for Computational Linguistics: Human Language Technologies
  (Volume 2: Short Papers)}, pp.\  385--395, Albuquerque, New Mexico, April
  2025. Association for Computational Linguistics.
\newblock ISBN 979-8-89176-190-2.
\newblock \doi{10.18653/v1/2025.naacl-short.33}.
\newblock URL \url{https://aclanthology.org/2025.naacl-short.33/}.

\bibitem[Llama~Team(2024)]{grattafiori2024llama3}
AI~@~Meta Llama~Team.
\newblock The llama 3 herd of models, 2024.
\newblock URL \url{https://arxiv.org/abs/2407.21783}.

\bibitem[Loshchilov \& Hutter(2019)Loshchilov and
  Hutter]{loshchilov2018decoupled}
Ilya Loshchilov and Frank Hutter.
\newblock Decoupled weight decay regularization.
\newblock In \emph{International Conference on Learning Representations}, 2019.
\newblock URL \url{https://openreview.net/forum?id=Bkg6RiCqY7}.

\bibitem[Makelov et~al.(2023)Makelov, Lange, and Nanda]{makelov2023illusions}
Aleksandar Makelov, Georg Lange, and Neel Nanda.
\newblock Is this the subspace you are looking for? an interpretability
  illusion for subspace activation patching, 2023.
\newblock URL \url{https://arxiv.org/abs/2311.17030}.

\bibitem[Marchetti et~al.(2026)Marchetti, Kunin, Myers, Acosta, and
  Miolane]{marchetti2026sequential}
Giovanni~Luca Marchetti, Daniel Kunin, Adele Myers, Francisco Acosta, and Nina
  Miolane.
\newblock Sequential group composition: A window into the mechanics of deep
  learning.
\newblock In \emph{Forty-third International Conference on Machine Learning},
  2026.
\newblock URL \url{https://openreview.net/forum?id=oFD9pc53pF}.

\bibitem[Meng et~al.(2022)Meng, Bau, Andonian, and Belinkov]{meng2022locating}
Kevin Meng, David Bau, Alex Andonian, and Yonatan Belinkov.
\newblock Locating and editing factual associations in {GPT}.
\newblock \emph{Advances in Neural Information Processing Systems}, 36, 2022.
\newblock arXiv:2202.05262.

\bibitem[Merity et~al.(2017)Merity, Xiong, Bradbury, and
  Socher]{merity2017pointer}
Stephen Merity, Caiming Xiong, James Bradbury, and Richard Socher.
\newblock Pointer sentinel mixture models.
\newblock In \emph{International Conference on Learning Representations}, 2017.
\newblock URL \url{https://openreview.net/forum?id=Byj72udxe}.

\bibitem[Modell et~al.(2025)Modell, Rubin-Delanchy, and
  Whiteley]{modell2025origins}
Alexander Modell, Patrick Rubin-Delanchy, and Nick Whiteley.
\newblock The origins of representation manifolds in large language models,
  2025.
\newblock URL \url{https://arxiv.org/abs/2505.18235}.

\bibitem[Morwani et~al.(2024)Morwani, Edelman, Oncescu, Zhao, and
  Kakade]{morwani2024feature}
Depen Morwani, Benjamin~L. Edelman, Costin-Andrei Oncescu, Rosie Zhao, and
  Sham~M. Kakade.
\newblock Feature emergence via margin maximization: case studies in algebraic
  tasks.
\newblock In \emph{The Twelfth International Conference on Learning
  Representations}, 2024.
\newblock URL \url{https://openreview.net/forum?id=i9wDX850jR}.

\bibitem[Mueller et~al.(2026)Mueller, Brinkmann, Li, Marks, Pal, Prakash,
  Rager, Sankaranarayanan, Sen~Sharma, Sun, Todd, Bau, and
  Belinkov]{mueller2026quest}
Aaron Mueller, Jannik Brinkmann, Millicent Li, Samuel Marks, Koyena Pal, Nikhil
  Prakash, Can Rager, Aruna Sankaranarayanan, Arnab Sen~Sharma, Jiuding Sun,
  Eric Todd, David Bau, and Yonatan Belinkov.
\newblock The quest for the right mediator: Surveying mechanistic
  interpretability for {NLP} through the lens of causal mediation analysis.
\newblock \emph{Computational Linguistics}, 52\penalty0 (1):\penalty0 331--378,
  2026.
\newblock \doi{10.1162/coli.a.572}.
\newblock URL \url{https://aclanthology.org/2026.cl-1.10/}.

\bibitem[Nanda et~al.(2023)Nanda, Chan, Lieberum, Smith, and
  Steinhardt]{nanda2023progress}
Neel Nanda, Lawrence Chan, Tom Lieberum, Jess Smith, and Jacob Steinhardt.
\newblock Progress measures for grokking via mechanistic interpretability.
\newblock In \emph{The Eleventh International Conference on Learning
  Representations}. OpenReview.net, 2023.
\newblock URL \url{https://openreview.net/forum?id=9XFSbDPmdW}.

\bibitem[Nishi et~al.(2025)Nishi, Ramesh, Okawa, Khona, Tanaka, and
  Lubana]{nishi2025representation}
Kento Nishi, Rahul Ramesh, Maya Okawa, Mikail Khona, Hidenori Tanaka, and
  Ekdeep~Singh Lubana.
\newblock Representation shattering in transformers: A synthetic study with
  knowledge editing.
\newblock In \emph{Forty-second International Conference on Machine Learning},
  2025.
\newblock URL \url{https://openreview.net/forum?id=BKOeyZal0x}.

\bibitem[Park et~al.(2025)Park, Choe, Jiang, and Veitch]{park2025hierarchical}
Kiho Park, Yo~Joong Choe, Yibo Jiang, and Victor Veitch.
\newblock The geometry of categorical and hierarchical concepts in large
  language models.
\newblock In \emph{The Thirteenth International Conference on Learning
  Representations}. OpenReview.net, 2025.
\newblock URL \url{https://openreview.net/forum?id=bVTM2QKYuA}.

\bibitem[Pearl(2001)]{pearl2001directindirecteffects}
Judea Pearl.
\newblock Direct and indirect effects, 2001.
\newblock URL \url{https://arxiv.org/abs/1301.2300}.

\bibitem[Penedo et~al.(2024)Penedo, Kydl{\'\i}{\v{c}}ek, allal, Lozhkov,
  Mitchell, Raffel, Werra, and Wolf]{penedo2024fineweb}
Guilherme Penedo, Hynek Kydl{\'\i}{\v{c}}ek, Loubna~Ben allal, Anton Lozhkov,
  Margaret Mitchell, Colin Raffel, Leandro~Von Werra, and Thomas Wolf.
\newblock The fineweb datasets: Decanting the web for the finest text data at
  scale.
\newblock In \emph{The Thirty-eight Conference on Neural Information Processing
  Systems Datasets and Benchmarks Track}, 2024.
\newblock URL \url{https://openreview.net/forum?id=n6SCkn2QaG}.

\bibitem[Prakash et~al.(2024)Prakash, Shaham, Haklay, Belinkov, and
  Bau]{prakash2024finetuning}
Nikhil Prakash, Tamar~Rott Shaham, Tal Haklay, Yonatan Belinkov, and David Bau.
\newblock Fine-tuning enhances existing mechanisms: A case study on entity
  tracking.
\newblock In \emph{The Twelfth International Conference on Learning
  Representations}, 2024.
\newblock URL \url{https://openreview.net/forum?id=8sKcAWOf2D}.

\bibitem[Prakash et~al.(2025)Prakash, Ren, Moritz, and Assogba]{prakash2025cca}
Nikhil Prakash, Donghao Ren, Dominik Moritz, and Yannick Assogba.
\newblock Constructive circuit amplification: Improving math reasoning in llms
  via targeted sub-network updates, 2025.
\newblock URL \url{https://arxiv.org/abs/2512.16914}.

\bibitem[Prakash et~al.(2026)Prakash, Shapira, Sen~Sharma, Riedl, Belinkov,
  Shaham, Bau, and Geiger]{prakash2026lookback}
Nikhil Prakash, Natalie Shapira, Arnab Sen~Sharma, Christoph Riedl, Yonatan
  Belinkov, Tamar Shaham, David Bau, and Atticus Geiger.
\newblock Language models use lookbacks to track beliefs.
\newblock In C.~Vondrick, B.~Hariharan, C.~Raffel, L.~Pinto, D.~Yang, and
  A.~Faust (eds.), \emph{International Conference on Learning Representations},
  volume 2026, pp.\  7657--7694, 2026.
\newblock URL
  \url{https://proceedings.iclr.cc/paper_files/paper/2026/file/0d1941f73833eb107939f62d323d6cc0-Paper-Conference.pdf}.

\bibitem[Prieto et~al.(2026)Prieto, Stevinson, Barsbey, Birdal, and
  Mediano]{prieto2026from}
Lucas Prieto, Edward Stevinson, Melih Barsbey, Tolga Birdal, and Pedro A.~M.
  Mediano.
\newblock From data statistics to feature geometry: How correlations shape
  superposition.
\newblock In \emph{The Fourteenth International Conference on Learning
  Representations}, 2026.
\newblock URL \url{https://openreview.net/forum?id=7akSRQS5Xh}.

\bibitem[Raissi et~al.(2019)Raissi, Perdikaris, and
  Karniadakis]{raissi2019physics}
M.~Raissi, P.~Perdikaris, and G.E. Karniadakis.
\newblock Physics-informed neural networks: A deep learning framework for
  solving forward and inverse problems involving nonlinear partial differential
  equations.
\newblock \emph{Journal of Computational Physics}, 378:\penalty0 686--707,
  2019.
\newblock ISSN 0021-9991.
\newblock \doi{https://doi.org/10.1016/j.jcp.2018.10.045}.
\newblock URL
  \url{https://www.sciencedirect.com/science/article/pii/S0021999118307125}.

\bibitem[Ravichander et~al.(2021)Ravichander, Belinkov, and
  Hovy]{ravichander2021probing}
Abhilasha Ravichander, Yonatan Belinkov, and Eduard Hovy.
\newblock Probing the probing paradigm: Does probing accuracy entail task
  relevance?
\newblock In Paola Merlo, Jorg Tiedemann, and Reut Tsarfaty (eds.),
  \emph{Proceedings of the 16th Conference of the {E}uropean Chapter of the
  {A}ssociation for {C}omputational {L}inguistics: Main Volume}, pp.\
  3363--3377, Online, April 2021. Association for Computational Linguistics.
\newblock \doi{10.18653/v1/2021.eacl-main.295}.
\newblock URL \url{https://aclanthology.org/2021.eacl-main.295/}.

\bibitem[Rumelhart et~al.(1986)Rumelhart, Hinton, and
  Williams]{rumelhart1986learning}
David~E Rumelhart, Geoffrey~E Hinton, and Ronald~J Williams.
\newblock Learning representations by back-propagating errors.
\newblock \emph{nature}, 1986.

\bibitem[Sarfati et~al.(2026)Sarfati, Bigelow, Wurgaft, Merullo, Geiger, Lewis,
  McGrath, and Lubana]{sarfati2026shape}
Raphaël Sarfati, Eric Bigelow, Daniel Wurgaft, Jack Merullo, Atticus Geiger,
  Owen Lewis, Tom McGrath, and Ekdeep~Singh Lubana.
\newblock The shape of beliefs: Geometry, dynamics, and interventions along
  representation manifolds of language models' posteriors, 2026.
\newblock URL \url{https://arxiv.org/abs/2602.02315}.

\bibitem[Sch{\"o}nemann(1966)]{schonemann1966generalized}
Peter~H. Sch{\"o}nemann.
\newblock A generalized solution of the orthogonal {P}rocrustes problem.
\newblock \emph{Psychometrika}, 31\penalty0 (1):\penalty0 1--10, 1966.

\bibitem[Sinii \& Balagansky(2026)Sinii and Balagansky]{sinii2026chasing}
Viacheslav Sinii and Nikita Balagansky.
\newblock Chasing the counting manifold in open llms, 2026.
\newblock URL \url{https://huggingface.co/spaces/t-tech/manifolds}.

\bibitem[Stolfo et~al.(2023)Stolfo, Belinkov, and
  Sachan]{stolfo2023mechanistic}
Alessandro Stolfo, Yonatan Belinkov, and Mrinmaya Sachan.
\newblock A mechanistic interpretation of arithmetic reasoning in language
  models using causal mediation analysis.
\newblock In \emph{Proceedings of the 2023 Conference on Empirical Methods in
  Natural Language Processing}, pp.\  7035--7052. Association for Computational
  Linguistics, 2023.
\newblock \doi{10.18653/v1/2023.emnlp-main.435}.
\newblock URL \url{https://aclanthology.org/2023.emnlp-main.435/}.

\bibitem[Todd et~al.(2024)Todd, Li, Sharma, Mueller, Wallace, and
  Bau]{todd2024function}
Eric Todd, Millicent Li, Arnab~Sen Sharma, Aaron Mueller, Byron~C Wallace, and
  David Bau.
\newblock Function vectors in large language models.
\newblock In \emph{The Twelfth International Conference on Learning
  Representations}, 2024.
\newblock URL \url{https://openreview.net/forum?id=AwyxtyMwaG}.

\bibitem[Todd et~al.(2026)Todd, Brinkmann, Gandikota, and
  Bau]{todd2026incontext}
Eric Todd, Jannik Brinkmann, Rohit Gandikota, and David Bau.
\newblock In-context algebra.
\newblock In \emph{The Fourteenth International Conference on Learning
  Representations}, 2026.
\newblock URL \url{https://openreview.net/forum?id=J2peqXPQbB}.

\bibitem[Vig et~al.(2020)Vig, Gehrmann, Belinkov, Qian, Nevo, Singer, and
  Shieber]{vig2020investigating}
Jesse Vig, Sebastian Gehrmann, Yonatan Belinkov, Sharon Qian, Daniel Nevo,
  Yaron Singer, and Stuart Shieber.
\newblock Investigating gender bias in language models using causal mediation
  analysis.
\newblock In \emph{Advances in Neural Information Processing Systems},
  volume~33, pp.\  12388--12401. Curran Associates, Inc., 2020.
\newblock URL
  \url{https://proceedings.neurips.cc/paper_files/paper/2020/file/92650b2e92217715fe312e6fa7b90d82-Paper.pdf}.

\bibitem[Wang et~al.(2021)Wang, Walters, and Yu]{wang2021incorporating}
Rui Wang, Robin Walters, and Rose Yu.
\newblock Incorporating symmetry into deep dynamics models for improved
  generalization.
\newblock In \emph{International Conference on Learning Representations}, 2021.
\newblock URL \url{https://openreview.net/forum?id=wta_8Hx2KD}.

\bibitem[Wu et~al.(2022)Wu, Geiger, Rozner, Kreiss, Lu, Icard, Potts, and
  Goodman]{wu2021distill}
Zhengxuan Wu, Atticus Geiger, Josh Rozner, Elisa Kreiss, Hanson Lu, Thomas
  Icard, Christopher Potts, and Noah~D. Goodman.
\newblock Causal distillation for language models.
\newblock In \emph{Proceedings of the 2022 Conference of the North American
  Chapter of the Association for Computational Linguistics: Human Language
  Technologies}, pp.\  4288--4295, Seattle, United States, July 2022.
  Association for Computational Linguistics.
\newblock \doi{10.18653/v1/2022.naacl-main.318}.
\newblock URL \url{https://aclanthology.org/2022.naacl-main.318}.

\bibitem[Wu et~al.(2023{\natexlab{a}})Wu, D'Oosterlinck, Geiger, Zur, and
  Potts]{wu2023causalproxy}
Zhengxuan Wu, Karel D'Oosterlinck, Atticus Geiger, Amir Zur, and Christopher
  Potts.
\newblock Causal proxy models for concept-based model explanations.
\newblock In Andreas Krause, Emma Brunskill, Kyunghyun Cho, Barbara Engelhardt,
  Sivan Sabato, and Jonathan Scarlett (eds.), \emph{Proceedings of the 40th
  International Conference on Machine Learning}, volume 202 of
  \emph{Proceedings of Machine Learning Research}, pp.\  37313--37334. PMLR,
  23--29 Jul 2023{\natexlab{a}}.
\newblock URL \url{https://proceedings.mlr.press/v202/wu23b.html}.

\bibitem[Wu et~al.(2023{\natexlab{b}})Wu, Geiger, Icard, Potts, and
  Goodman]{wu2023interpretability}
Zhengxuan Wu, Atticus Geiger, Thomas Icard, Christopher Potts, and Noah
  Goodman.
\newblock Interpretability at scale: Identifying causal mechanisms in alpaca.
\newblock In \emph{Thirty-seventh Conference on Neural Information Processing
  Systems}, 2023{\natexlab{b}}.
\newblock URL \url{https://openreview.net/forum?id=nRfClnMhVX}.

\bibitem[Wu et~al.(2024{\natexlab{a}})Wu, Arora, Wang, Geiger, Jurafsky,
  Manning, and Potts]{wu2024reft}
Zhengxuan Wu, Aryaman Arora, Zheng Wang, Atticus Geiger, Dan Jurafsky,
  Christopher~D Manning, and Christopher Potts.
\newblock Re{FT}: Representation finetuning for language models.
\newblock In \emph{The Thirty-eighth Annual Conference on Neural Information
  Processing Systems}, 2024{\natexlab{a}}.
\newblock URL \url{https://openreview.net/forum?id=fykjplMc0V}.

\bibitem[Wu et~al.(2024{\natexlab{b}})Wu, Geiger, Huang, Arora, Icard, Potts,
  and Goodman]{wu2024replyillusions}
Zhengxuan Wu, Atticus Geiger, Jing Huang, Aryaman Arora, Thomas Icard,
  Christopher Potts, and Noah~D. Goodman.
\newblock A reply to makelov et al. (2023)'s "interpretability illusion"
  arguments, 2024{\natexlab{b}}.
\newblock URL \url{https://arxiv.org/abs/2401.12631}.

\bibitem[Wurgaft et~al.(2026)Wurgaft, Rager, Kowal, Shyam, Feucht, Bhalla,
  Haklay, Bigelow, Sarfati, McGrath, Lewis, Merullo, Goodman, Fel, Geiger, and
  Lubana]{wurgaft2026manifold}
Daniel Wurgaft, Can Rager, Matthew Kowal, Vasudev Shyam, Sheridan Feucht, Usha
  Bhalla, Tal Haklay, Eric Bigelow, Raphael Sarfati, Thomas McGrath, Owen
  Lewis, Jack Merullo, Noah Goodman, Thomas Fel, Atticus Geiger, and
  Ekdeep~Singh Lubana.
\newblock Manifold steering reveals the shared geometry of neural network
  representation and behavior, 2026.
\newblock URL \url{https://arxiv.org/abs/2605.05115}.

\bibitem[Yin et~al.(2024)Yin, Ye, and
  Durrett]{yin2024lofitlocalizedfinetuningllm}
Fangcong Yin, Xi~Ye, and Greg Durrett.
\newblock Lofit: Localized fine-tuning on {LLM} representations.
\newblock In \emph{The Thirty-eighth Annual Conference on Neural Information
  Processing Systems}, 2024.
\newblock URL \url{https://openreview.net/forum?id=dfiXFbECSZ}.

\bibitem[Zhong et~al.(2023)Zhong, Liu, Tegmark, and Andreas]{zhong2023pizza}
Ziqian Zhong, Ziming Liu, Max Tegmark, and Jacob Andreas.
\newblock The clock and the pizza: Two stories in mechanistic explanation of
  neural networks.
\newblock In \emph{Advances in Neural Information Processing Systems},
  volume~36, pp.\  27223--27250. Curran Associates, Inc., 2023.
\newblock \doi{10.52202/075280-1186}.
\newblock URL
  \url{https://papers.nips.cc/paper_files/paper/2023/hash/56cbfbf49937a0873d451343ddc8c57d-Abstract-Conference.html}.

\bibitem[Zhou et~al.(2024)Zhou, Fu, Sharan, and Jia]{zhou2024fourier}
Tianyi Zhou, Deqing Fu, Vatsal Sharan, and Robin Jia.
\newblock Pre-trained large language models use fourier features to compute
  addition, 2024.
\newblock URL \url{https://arxiv.org/abs/2406.03445}.

\bibitem[Zhou et~al.(2026)Zhou, Fu, Soltanolkotabi, Jia, and
  Sharan]{zhou2025fone}
Tianyi Zhou, Deqing Fu, Mahdi Soltanolkotabi, Robin Jia, and Vatsal Sharan.
\newblock {FoNE}: Precise single-token number embeddings via {Fourier}
  features.
\newblock In \emph{International Conference on Learning Representations},
  volume 2026, pp.\  28485--28516, 2026.
\newblock URL
  \url{https://proceedings.iclr.cc/paper_files/paper/2026/hash/30253ebdaa75eed08c4b10c75aa8f781-Abstract-Conference.html}.

\bibitem[Zhu et~al.(2025)Zhu, Dai, and Sui]{zhu2025language}
Fangwei Zhu, Damai Dai, and Zhifang Sui.
\newblock Language models encode the value of numbers linearly.
\newblock In Owen Rambow, Leo Wanner, Marianna Apidianaki, Hend Al-Khalifa,
  Barbara~Di Eugenio, and Steven Schockaert (eds.), \emph{Proceedings of the
  31st International Conference on Computational Linguistics}, pp.\  693--709,
  Abu Dhabi, UAE, January 2025. Association for Computational Linguistics.
\newblock URL \url{https://aclanthology.org/2025.coling-main.47/}.

\bibitem[Zou et~al.(2023)Zou, Phan, Chen, Campbell, Guo, Ren, Pan, Yin,
  Mazeika, Dombrowski, Goel, Li, Byun, Wang, Mallen, Basart, Koyejo, Song,
  Fredrikson, Kolter, and Hendrycks]{zou2023representation}
Andy Zou, Long Phan, Sarah Chen, James Campbell, Phillip Guo, Richard Ren,
  Alexander Pan, Xuwang Yin, Mantas Mazeika, Ann-Kathrin Dombrowski, Shashwat
  Goel, Nathaniel Li, Michael~J. Byun, Zifan Wang, Alex Mallen, Steven Basart,
  Sanmi Koyejo, Dawn Song, Matt Fredrikson, J.~Zico Kolter, and Dan Hendrycks.
\newblock Representation engineering: A top-down approach to {AI} transparency.
\newblock \emph{arXiv preprint arXiv:2310.01405}, 2023.
\newblock URL \url{https://arxiv.org/abs/2310.01405}.

\bibitem[Zur et~al.(2024)Zur, Kreiss, D{'}Oosterlinck, Potts, and
  Geiger]{zur2024updating}
Amir Zur, Elisa Kreiss, Karel D{'}Oosterlinck, Christopher Potts, and Atticus
  Geiger.
\newblock Updating {CLIP} to prefer descriptions over captions.
\newblock In Yaser Al-Onaizan, Mohit Bansal, and Yun-Nung Chen (eds.),
  \emph{Proceedings of the 2024 Conference on Empirical Methods in Natural
  Language Processing}, pp.\  20178--20187, Miami, Florida, USA, November 2024.
  Association for Computational Linguistics.
\newblock \doi{10.18653/v1/2024.emnlp-main.1125}.
\newblock URL \url{https://aclanthology.org/2024.emnlp-main.1125}.

\end{thebibliography}
\bibliographystyle{iclr2027_conference}

\clearpage
\appendix

\section{Months Task: Experimental Details}
\subsection{Training Task LoRAs}
\label{appdx:task_lora_training}
In this section, we provide more details about how we train LoRAs with end-to-end supervision, as well as several ablations that measure how hyperparameters like location, rank, training data range, etc. affect the generalization performance of task LoRAs.
All task LoRAs are trained with a standard cross-entropy loss at the final token position between the model's prediction and the correct month.

We train task LoRAs at the MLP of a single layer (i.e., the gate, up, and down projections), the LoRA update: $\Delta = s\,(xA^\top)B^\top$ with $s = 2$, is applied at every token position.
We use the AdamW optimizer~\citep{loshchilov2018decoupled} with a learning rate of $2\times10^{-3}$, weight decay of $0.01$, gradient-norm clipping at $1.0$, and a constant learning rate schedule, and we train for $200$ steps over the training set: $X_{\texttt{sum}\leq 24}$.
The base model is Llama-3.1-8B in \texttt{bf16} and the weights are kept frozen, while the LoRA parameters are kept in \texttt{fp32}.
The task LoRA results reported in the main paper use a rank-8 LoRA trained at the layer-$3$ MLP which reaches $71.9\%$ accuracy on the months task.

We score a prediction as correct if the highest-scoring token among the twelve month tokens is the correct month.
We use a restricted argmax to the twelve month tokens to simplify evaluation, but also find that the model's top predicted token is a month token for $99.3\%$ of prompts.

\begin{figure}[h]
    \centering
    \includegraphics[width=0.75\linewidth]{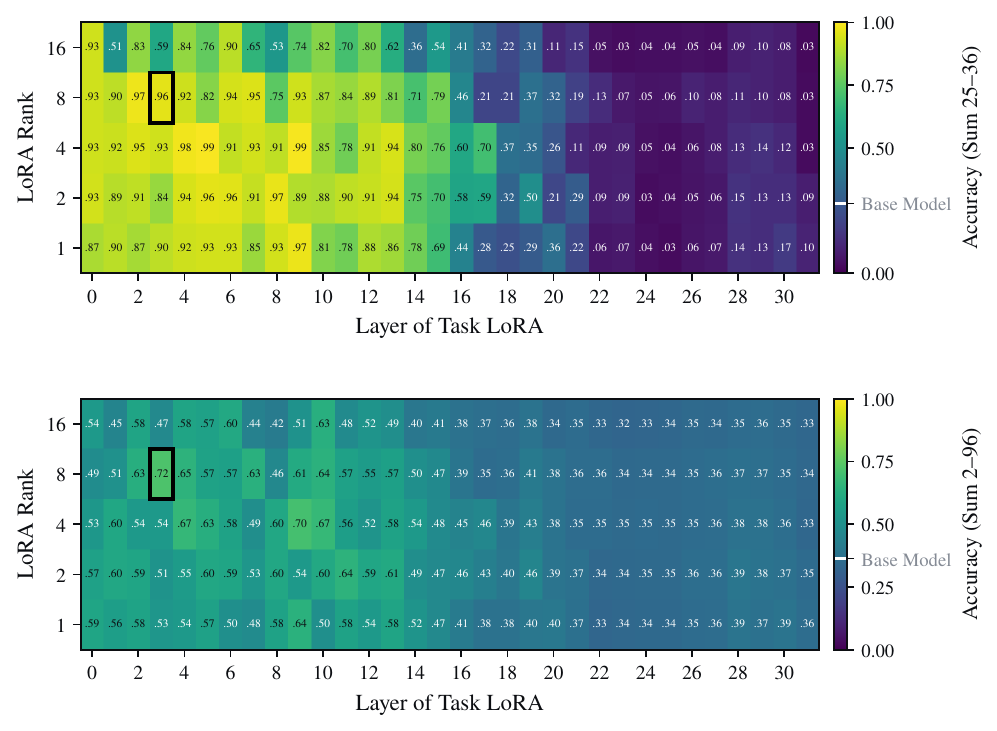}
    \caption{Comparing performance of task LoRAs trained at different layers ($0-31$) with different ranks (\{1,2,4,8,16\}). 
    We measure performance on the first held-out cycle: $X_{\texttt{sum}\in[25,36]}$ (top), and the full dataset $X_{\texttt{sum}\leq96}$ (bottom). 
    The base model scores $27.8\%$ on the held-out cycle range. We find that training a task LoRA at early layer MLPs ($l \leq 14$) for any rank $\leq 8$ is able to fix almost all of the errors the base model makes. After about layer $14$, training a task LoRA no longer meaningfully improves model performance compared to the base model's accuracy. The task LoRA configuration reported in the main paper (layer 3, rank 8) is shown with a black box.}
    \label{fig:months_task_lora_layer_rank}
\end{figure}
\textbf{Training layer and rank.} In Figure~\ref{fig:months_task_lora_layer_rank}, we investigate the effect of training LoRAs of different ranks ($\{1,2,4,8,16\}$) and at different layers (i.e., the MLP of layer $l\in\{0,...31\}$), averaged over three seeds. The task LoRA configuration used in the main paper (layer 3, rank 8) is shown with a black box.
For each layer/rank combination, we evaluate the LoRAs on the first held-out cycle: $X_{\texttt{sum}\in[25,36]}$, and compare to the Llama-3.1-8B's performance of $27.8\%$ on this same range. 
We find that training a low-rank task LoRA at a range of early layer MLPs ($l \leq 14$) is sufficient to solve the majority of the hold-out prompts ($95{-}100\%$). However, task LoRAs trained at later layers do not meaningfully improve task performance beyond the base model's accuracy.
We found this early/late distinction surprising, as prior work \citep{feucht2026arithmetic} hypothesized that some of the model's errors were due to incorrectly computing the $\bmod$, which would be most likely to happen in later layers after the \texttt{sum} computation (which was previously localized to layer $18$).

\begin{table}
\centering\small
\caption{Comparing single layer vs. multi-layer LoRA training. Each LoRA reported here is rank-1 and trained on $X_{\texttt{sum}\leq24}$ with end-to-end supervision. We report the full accuracy as well as three evaluation bands. The first band is in-range: $\texttt{sum}\in[2,24]$, the other two are held-out ranges: $\texttt{sum}\in[25,48]$, and $\texttt{sum}\in[49,96]$. Scores are mean $\pm$ standard deviation over 3 seeds. Multi-layer LoRAs use a learning rate of $5\times10^{-4}$.}
\label{tab:one_vs_all_layers}
\begin{tabular}{llcrrrrr}
\toprule
& & & & \multicolumn{4}{c}{Accuracy (\%)} \\
\cmidrule(l){5-8}
Objective & Location & Layers & \# Params. & $\texttt{sum}\in[2,96]$ & $[2,24]$ & $[25,48]$ & $[49,96]$ \\
\midrule
--- & Base model & --- & 0 & 36.3 & 79.0 & 22.9 & 27.4 \\
\midrule
Task CE & MLP & 4 & 55K & 54.3 $\pm$ 4.2 & 100.0 $\pm$ 0.0 & 68.2 $\pm$ 7.8 & 30.7 $\pm$ 4.2 \\
Task CE & MLP & 0--16 & 940K & 55.2 $\pm$ 1.6 & 100.0 $\pm$ 0.0 & 70.8 $\pm$ 3.1 & 31.0 $\pm$ 4.4 \\
Task CE & MLP & 0--31 & 1.77M & 56.5 $\pm$ 1.4 & 100.0 $\pm$ 0.0 & 60.9 $\pm$ 2.3 & 38.5 $\pm$ 2.9 \\
Task CE & MLP + Attn & 0--31 & 2.62M & 51.3 $\pm$ 3.2 & 100.0 $\pm$ 0.0 & 55.7 $\pm$ 3.1 & 31.3 $\pm$ 5.2 \\
\bottomrule
\end{tabular}
\end{table}

\textbf{Task LoRA on a single layer vs. every layer.} In Table~\ref{tab:one_vs_all_layers}, we measure the effect that training LoRAs on multiple layers has, compared to training a LoRA on just a single layer.
We report either a single LoRA trained at the layer $4$ MLP, or many LoRAs trained at layers $0-16$, $0-31$, and $0-31$ including attention layers. 
We find that when training LoRAs with task supervision, adding more parameters/layers does not consistently improve performance compared to training a single LoRA. The only noticeable improvement is that training on layer $0-31$ MLPs improves accuracy on the $\texttt{sum}\in[49,96]$ range by ${\sim}8\%$, but this also hurts performance on the $\texttt{sum}\in[25,48]$ range by the same amount.

\begin{figure}[h]
    \centering
    \includegraphics[width=0.75\linewidth]{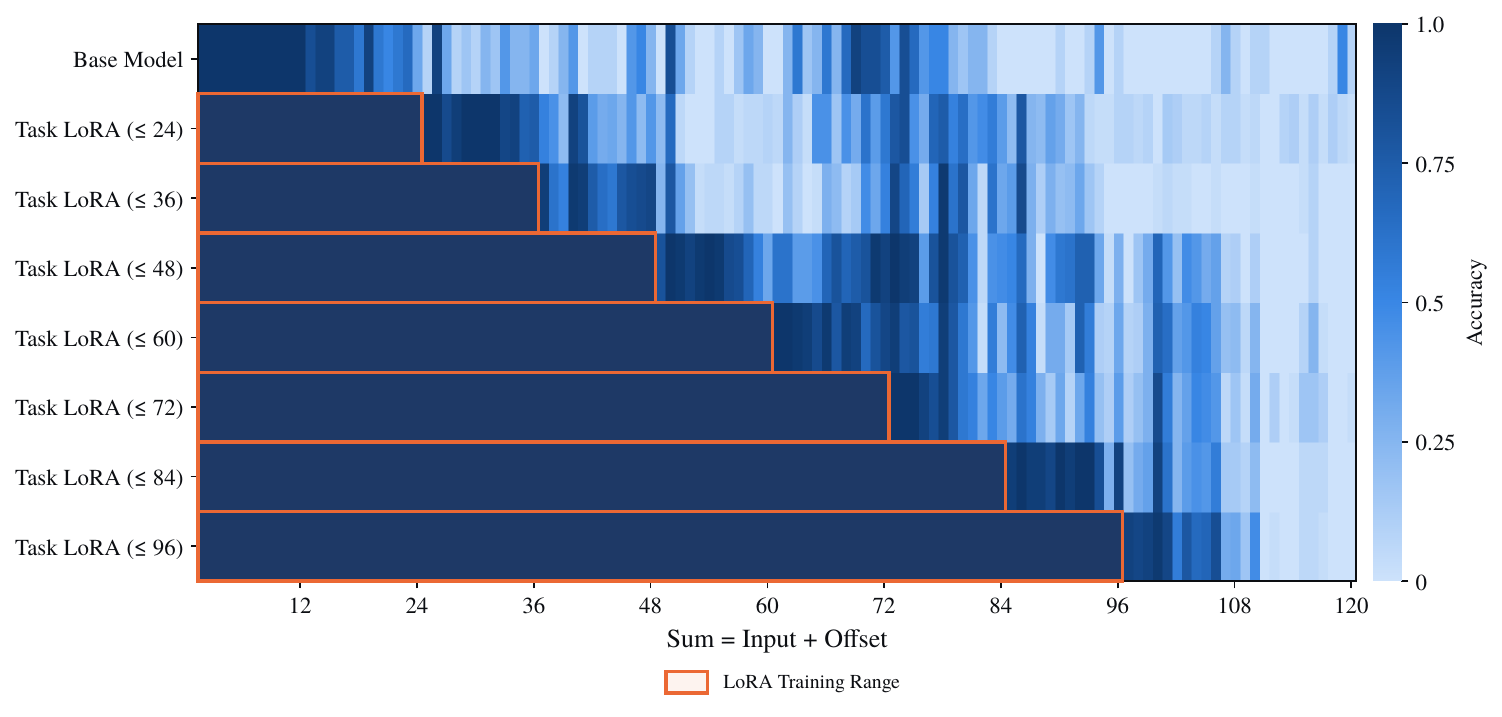}
    \caption{Performance of Llama-3.1-8B on the months task, grouped by \texttt{sum}. The model achieves $100\%$ on the task when the \texttt{sum} $\leq 12$, but performance drops for larger values. Training a LoRA with end-to-end supervision (i.e., ``Task LoRA'') on a subset of data in cycle-length increments ($X_{\texttt{sum}\leq k}$) can correct errors within its training range (shown in orange), but the generalization is shallow, often only up to one held-out cycle beyond the training range.}
    \label{fig:months_task_lora_ladder}
\end{figure}

\textbf{Training data range.} In Figure~\ref{fig:months_task_lora_ladder}, we show the results of training LoRAs on increasing amounts of data in cycle-length increments (i.e., $X_{\texttt{sum}\leq24}$, $X_{\texttt{sum}\leq36}$, ..., $X_{\texttt{sum}\leq96}$). We find that fine-tuning with end-to-end supervision ``fixes'' almost all errors \textit{within} the LoRA's training distribution, but that the generalization to unseen data is shallow, often only extending for up to one held-out cycle beyond the training range. Because we see the same generalization pattern for each training range we test, we use a single fixed training range ($X_{\texttt{sum}\leq24}$) for other experiments to ensure matched comparison.

\subsection{Fourier Probes}
\label{appdx:fourier_probes}
\textbf{Training Fourier Probes.} We follow \citet{feucht2026arithmetic} and train Fourier probes using a simple addition task where prompts have the following template $x=$``\textit{a+b=}''. The full probe training dataset $X_{a+b}$ is constructed by varying $a$ and $b$ over each possible value of $a,b \in \{1,...,199\}$.
For a given layer $l$ in the model $f_\theta$, and for each period $T \in \{2, ..., 100\}$, we train two affine probes to match the target sine and cosine harmonics. Each probe is trained by minimizing their corresponding MSE objective over $X_{a+b}$:
\begin{equation}
\label{eq:fourier_probes}
    \text{MSE}\Bigg(\langle\mathbf{w}_{\sin}^{(l,T)},h_{x}\rangle + b_{\sin}^{(l,T)}, \sin\!\left(\tfrac{2\pi c(x)}{T}\right)\Bigg),  \;
    \text{MSE}\Bigg(\langle\mathbf{w}_{\cos}^{(l,T)},h_{x}\rangle + b_{\cos}^{(l,T)},\cos\!\left(\tfrac{2\pi c(x)}{T}\right)\Bigg).
\end{equation}
where $\mathbf{w}_{\text{cos}}^{(l, T)}, \mathbf{w}_{\text{sin}}^{(l, T)} \in \mathbb{R}^{d}$ and $b_{\text{cos}}^{(l, T)}, b_{\text{sin}}^{(l, T)} \in \mathbb{R}$ and $c$ is the variable being probed. We train distinct probes for each value of $c$, which in this case, we have $c\in\{a, b, a+b\}$. 

When selecting probes for the months task, we use the Fourier probe trained at the corresponding variable slot.
To probe for representational structure of the \texttt{input} variable, we use probes trained with $\text{sin}\left(\frac{2\pi a}{T}\right)$ or $\text{cos}\left(\frac{2\pi a}{T}\right)$ as the target. 
To probe for representational structure of the \texttt{offset} variable, we use probes trained with $\text{sin}\left(\frac{2\pi b}{T}\right)$ or $\text{cos}\left(\frac{2\pi b}{T}\right)$ as the target. 
Similarly, we use probes trained with $\text{sin}\left(\frac{2\pi (a+b)}{T}\right)$ or $\text{cos}\left(\frac{2\pi (a+b)}{T}\right)$ as the target to probe for structure in the \texttt{sum} variable.

\begin{figure}
    \centering
    \includegraphics[width=\linewidth]{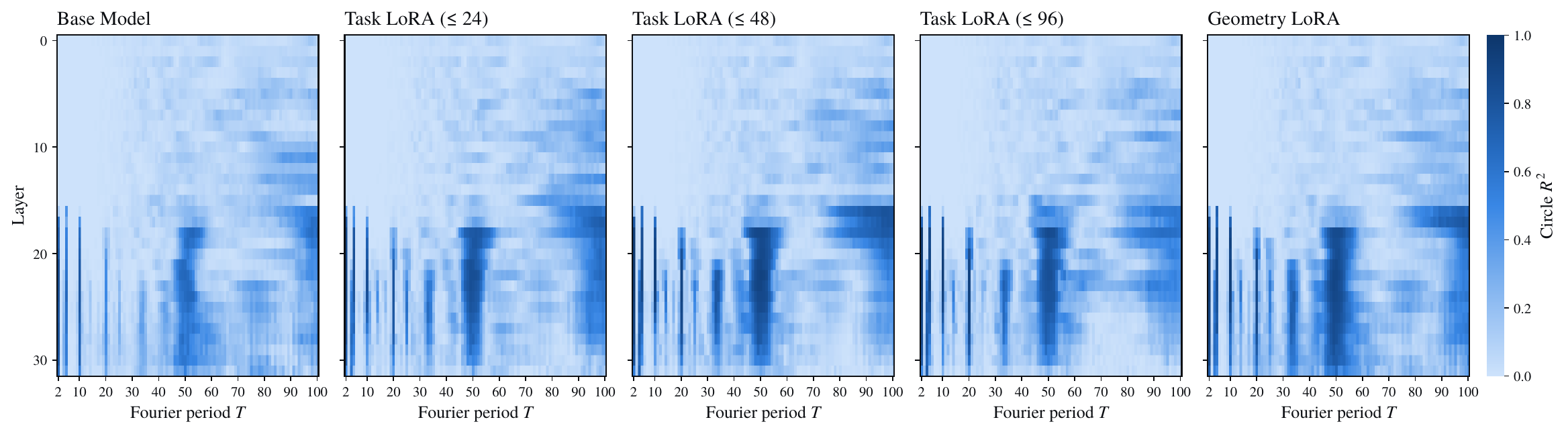}
    \caption{LoRA fine-tuning sharpens existing base-10 Fourier representations. We report the $R^2$ score of Fourier planes for all layers and across all periods $T\in\{2, ..., 100\}$ for Llama-3.1-8B and several LoRAs. Each subplot shows the $R^2$ value at a layer (row) and period $T$ (column) for the base model or a fine-tuned variant. Roughly the same periods have high $R^2$ values in each in case, with fine-tuned variants having stronger peaks around the previously identified base-10 periods: $\{2, 5, 10, 20, 50\}$.}
    \label{fig:months-period-heatmap}
\end{figure}

\begin{figure}
    \centering
    \includegraphics[width=\linewidth]{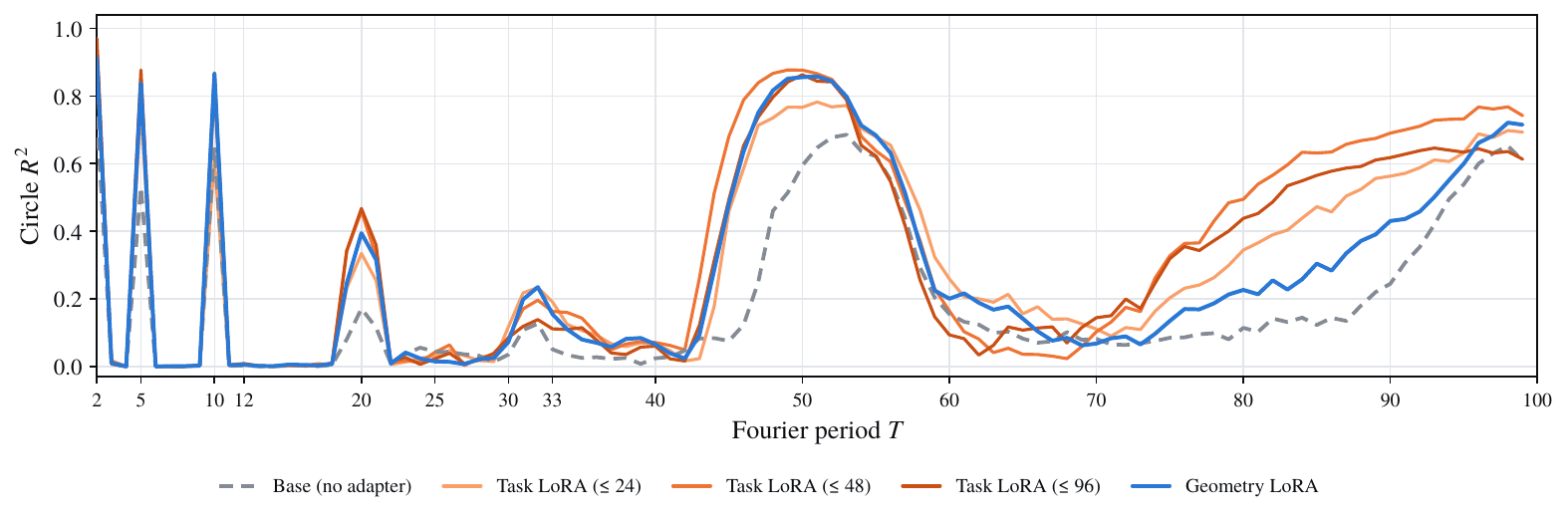}
    \caption{LoRA fine-tuning sharpens existing Fourier representations at Layer 18. We compute the $R^2$ value between Fourier planes and ideal circular representations at Layer 18 in Llama-3.1-8B. Task LoRAs (orange) and a geometry LoRA (blue) increase the $R^2$ of base-10 periods ($T\in\{2, 5, 10, 20,50\}$) compared to the base model (shown in gray), leaving most other high-frequency periods largely unaffected.}
    \label{fig:months-period-sharpening-L18}
\end{figure}

\begin{figure}
    \centering
    \includegraphics[width=\linewidth]{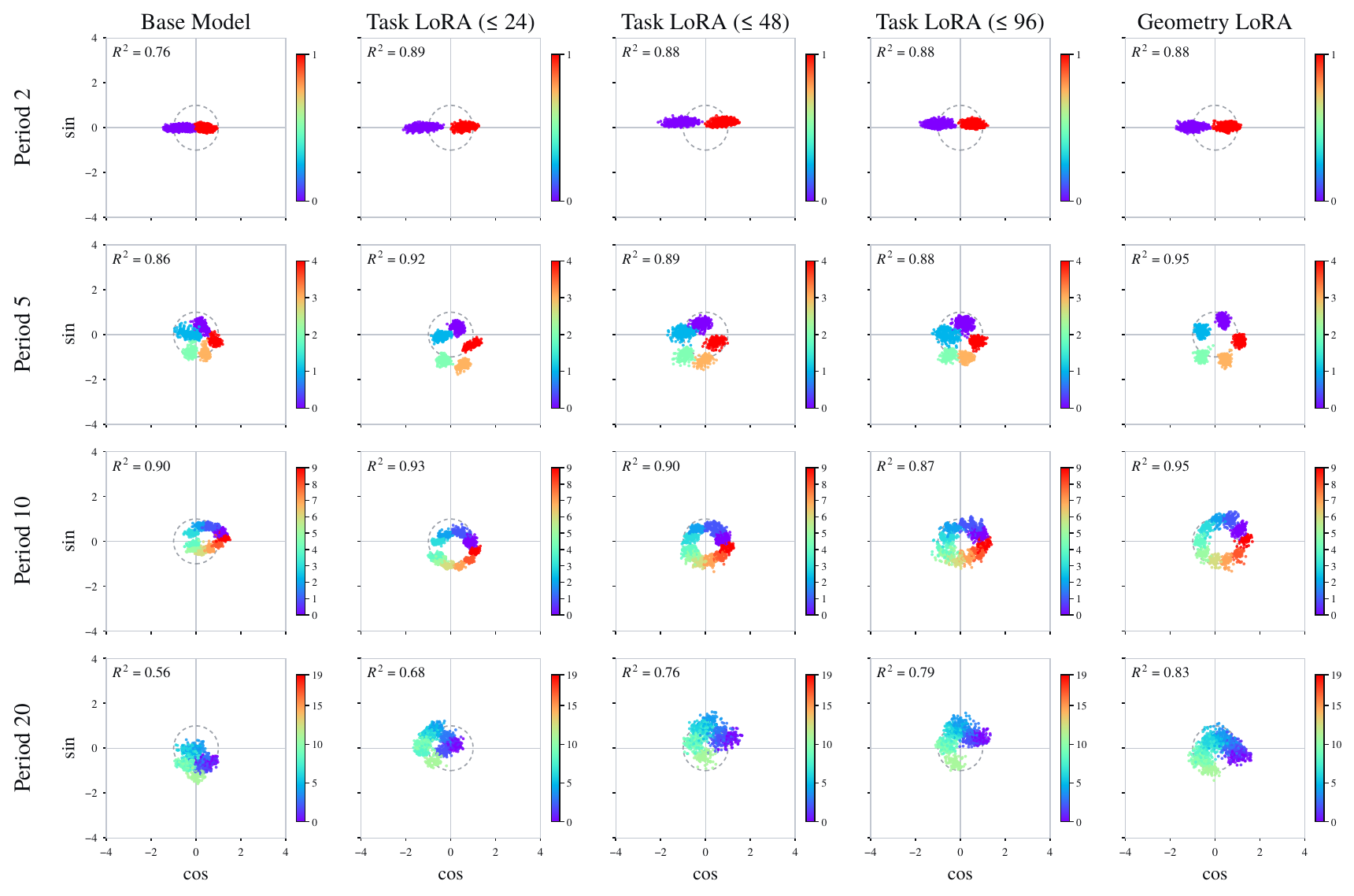}
    \caption{Projection of the months data onto Fourier planes of the \texttt{input} variable for periods $T=\{2, 5, 10, 20\}$ at layer $18$.
    Each point corresponds to a prompt $x \in X_{\texttt{sum}\leq96}$, and is colored by $\texttt{input}(x) \bmod T$. 
    We compare projections from the base model (left column), and several LoRAs: Task LoRA ($\texttt{sum}\leq24$), Task LoRA ($\texttt{sum}\leq48$), Task LoRA ($\texttt{sum}\leq96$), and a Geometry LoRA ($\texttt{sum}\leq24$). Since \texttt{input} only has $|\mathcal{V}_c|=12$ possible values, projecting it onto the period $T=20$ plane forms an arc instead of a circle.}
    \label{fig:months-L18-input-projection-appdx}
\end{figure}

\begin{figure}
    \centering
    \includegraphics[width=\linewidth]{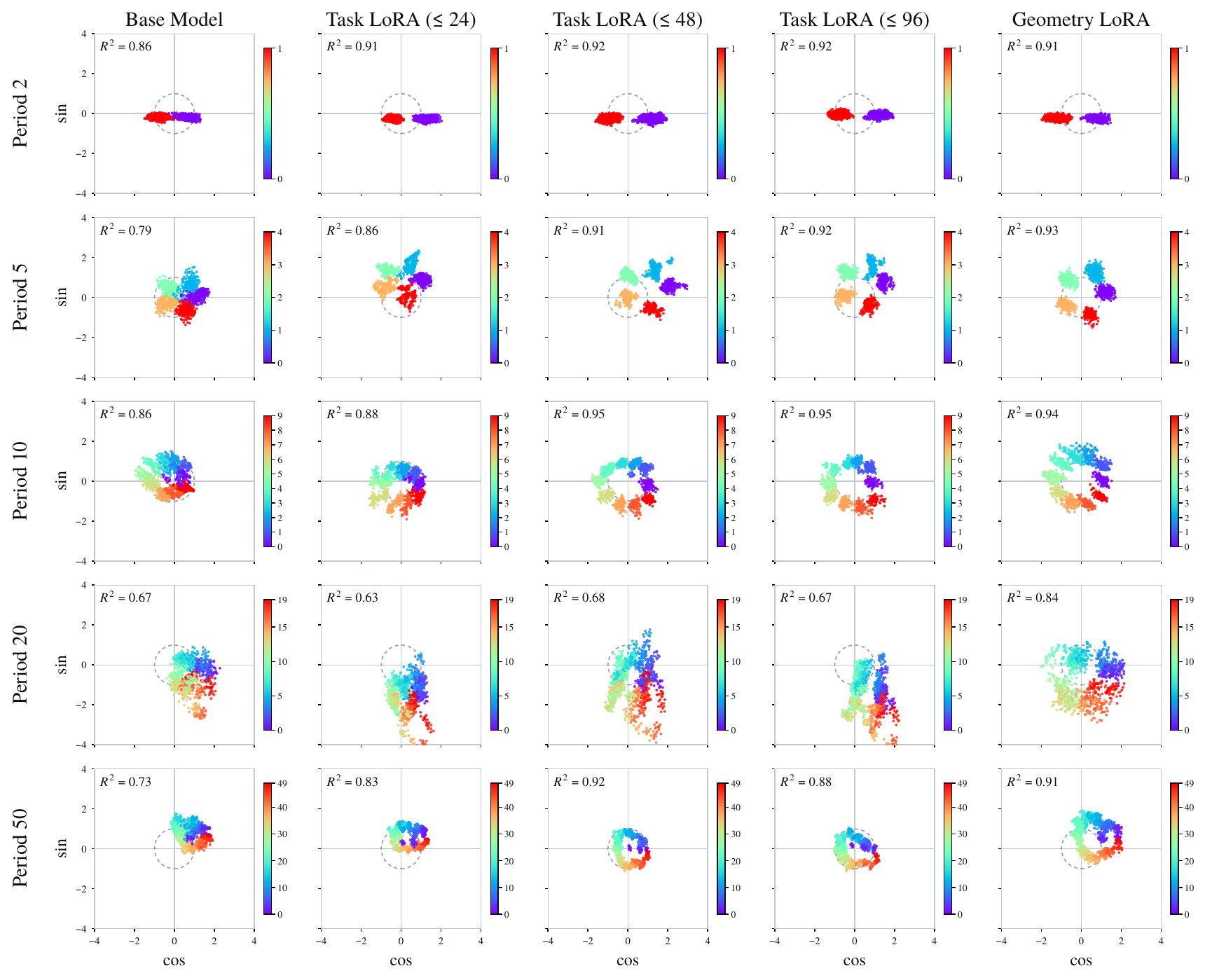}
    \caption{Projection of the months data onto Fourier planes of the \texttt{offset} variable for periods $T=\{2, 5, 10, 20, 50\}$ at layer $18$.
    Each point corresponds to a prompt $x \in X_{\texttt{sum}\leq96}$, and is colored by $\texttt{offset}(x) \bmod T$. 
    We compare projections from the base model (left column), and several LoRAs: Task LoRA ($\texttt{sum}\leq24$), Task LoRA ($\texttt{sum}\leq48$), Task LoRA ($\texttt{sum}\leq96$), and a Geometry LoRA ($\texttt{sum}\leq24$).}
    \label{fig:months-L18-offset-projection-appdx}
\end{figure}

\begin{figure}
    \centering
    \includegraphics[width=\linewidth]{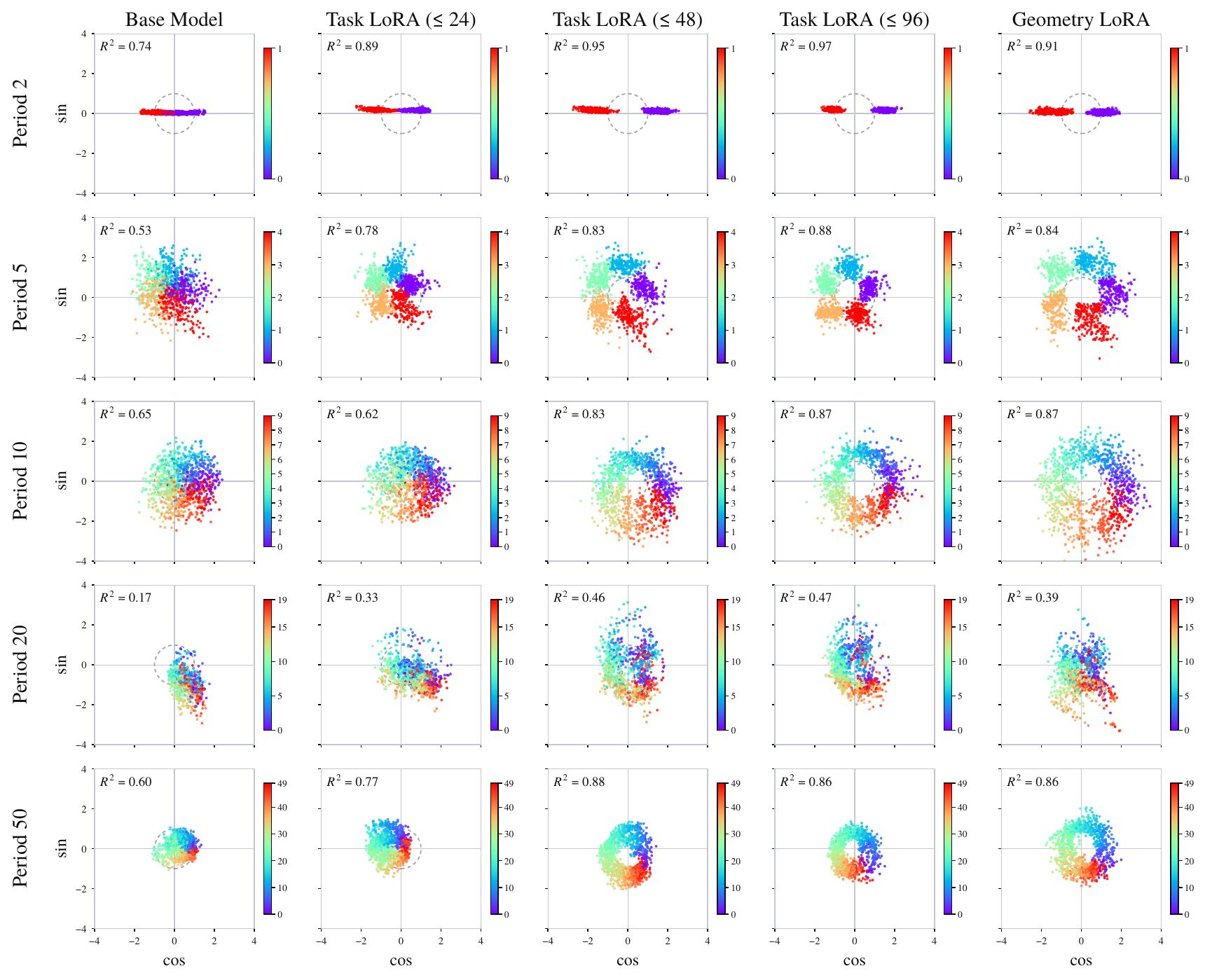}
    \caption{Projection of the months data onto Fourier planes of the \texttt{sum} variable for periods $T=\{2, 5, 10, 20, 50\}$ at layer $18$.
    Each point corresponds to a prompt $x \in X_{\texttt{sum}\leq96}$, and is colored by $\texttt{sum}(x) \bmod T$. 
    We compare projections from the base model (left column), and several LoRAs: Task LoRA ($\texttt{sum}\leq24$), Task LoRA ($\texttt{sum}\leq48$), Task LoRA ($\texttt{sum}\leq96$), and a Geometry LoRA ($\texttt{sum}\leq24$).}
    \label{fig:months-L18-sum-projection-appdx}
\end{figure}

\begin{figure}
    \centering
    \includegraphics[width=\linewidth]{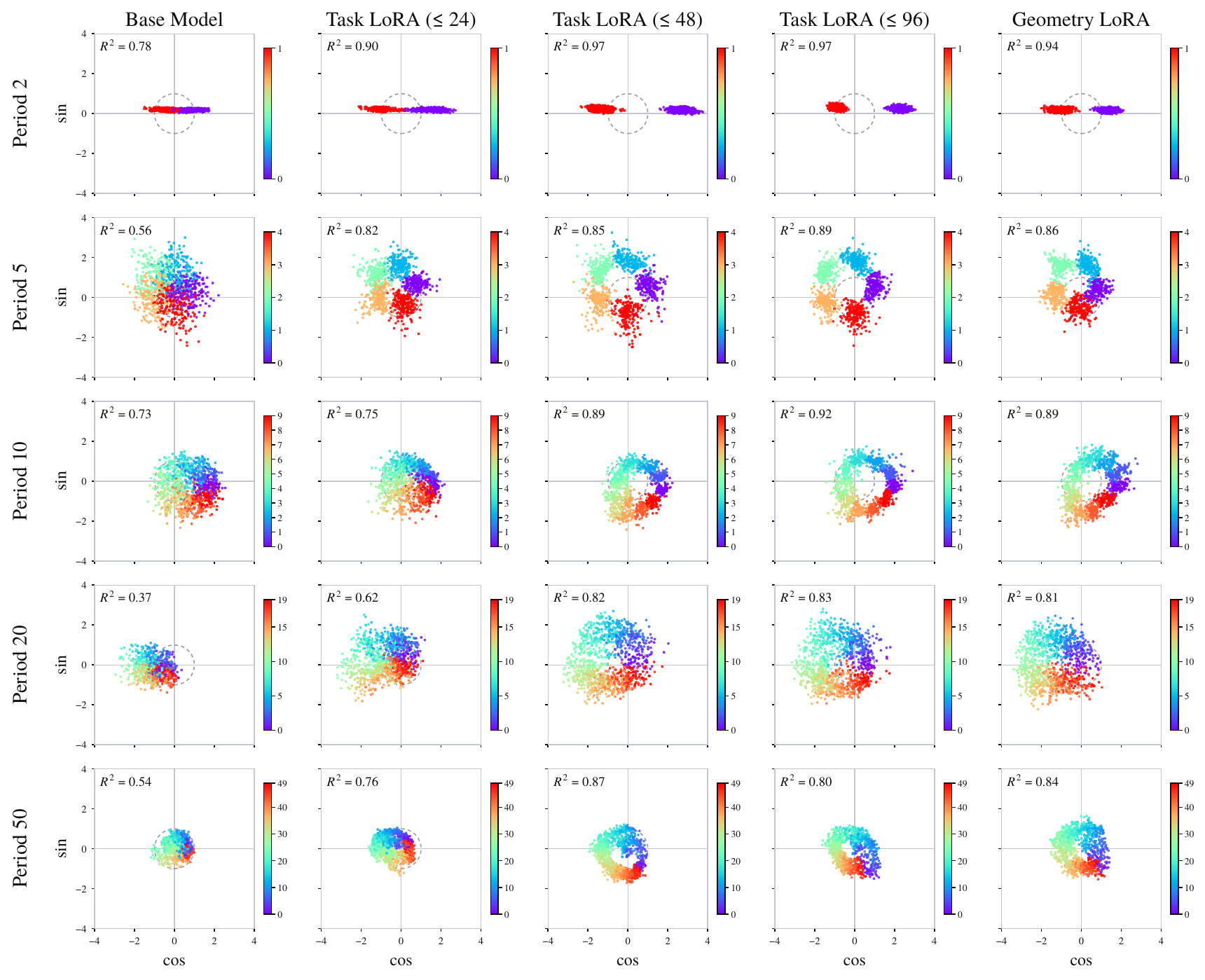}
    \caption{Projection of the months data onto Fourier planes of the \texttt{sum} variable for periods $T=\{2, 5, 10, 20, 50\}$ at layer $20$.
    Each point corresponds to a prompt $x \in X_{\texttt{sum}\leq96}$, and is colored by $\texttt{sum}(x) \bmod T$. 
    We compare projections from the base model (left column), and several LoRAs: Task LoRA ($\texttt{sum}\leq24$), Task LoRA ($\texttt{sum}\leq48$), Task LoRA ($\texttt{sum}\leq96$), and a Geometry LoRA ($\texttt{sum}\leq24$).}
    \label{fig:months-L20-sum-projection-appdx}
\end{figure}

\begin{figure}
    \centering
    \includegraphics[width=\linewidth]{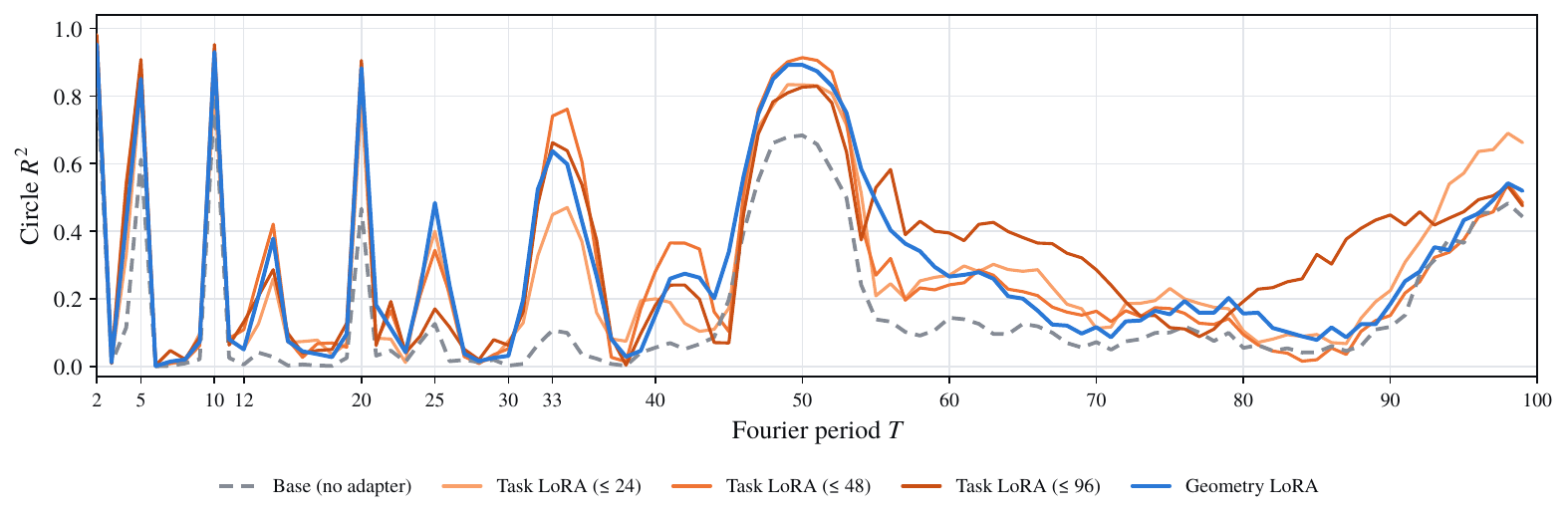}
    \caption{LoRA fine-tuning sharpens existing base-10 Fourier representations at Layer 22. We compute the $R^2$ value between Fourier planes for the \texttt{sum} and ideal circular representations at Layer 22 in Llama-3.1-8B. Task LoRAs (orange) and our Geometry LoRA (blue) increase the $R^2$ of base-10 periods ($T\in\{2, 5, 10, 20,50\}$) compared to the base model (shown in gray), leaving most other periods largely unaffected. Interestingly, for this layer, additional periods also appear to sharpen which are not (strongly) present in the base model including $T\in \{14, 25, 33\}$.}
    \label{fig:months-period-sharpening-L22}
\end{figure}

\begin{figure}
    \centering
    \includegraphics[width=\linewidth]{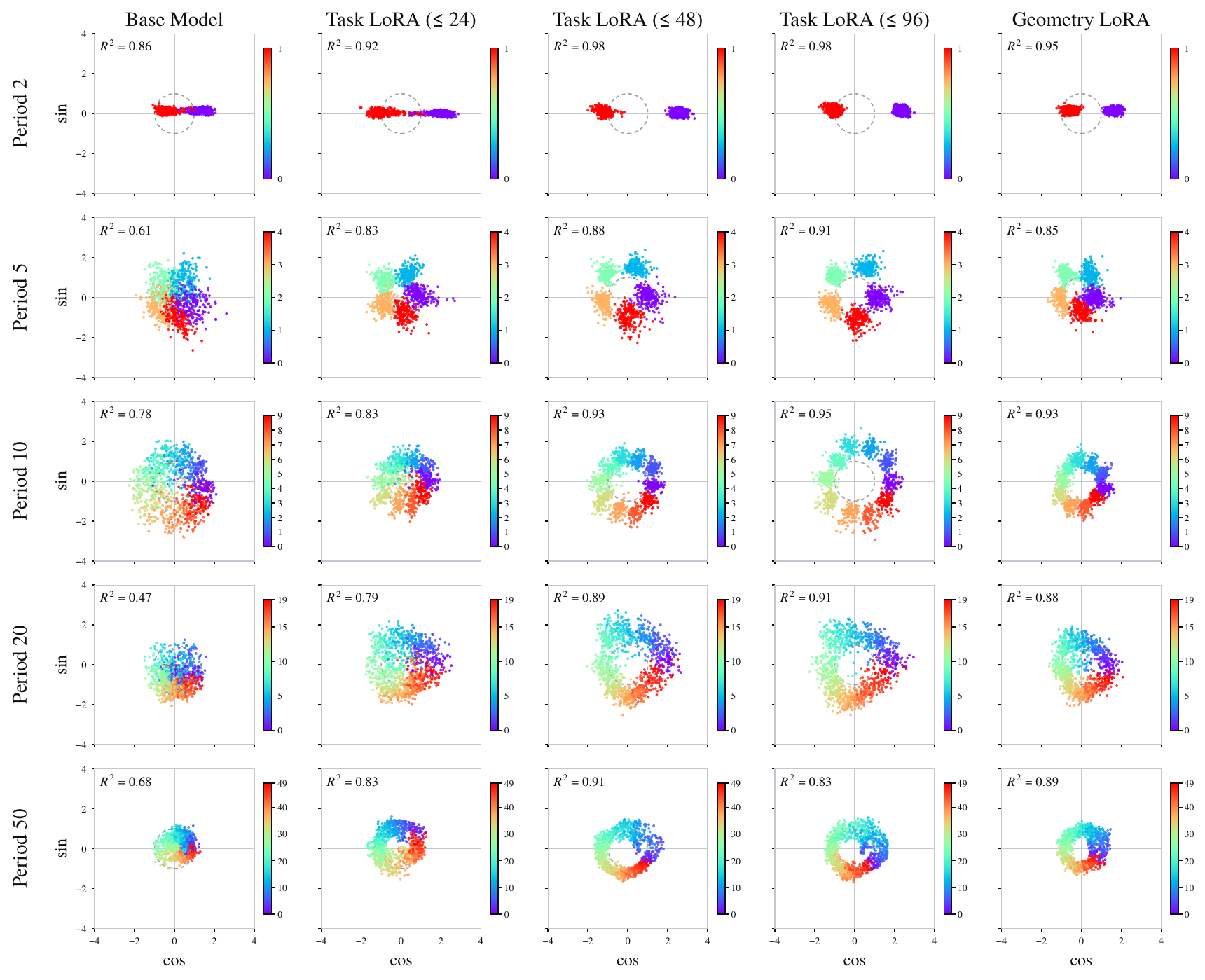}
    \caption{Projection of the months data onto Fourier planes of the \texttt{sum} variable for periods $T=\{2, 5, 10, 20, 50\}$ at layer $22$.
    Each point corresponds to a prompt $x \in X_{\texttt{sum}\leq96}$, and is colored by $\texttt{sum}(x) \bmod T$. 
    We compare projections from the base model (left column), and several LoRAs: Task LoRA ($\texttt{sum}\leq24$), Task LoRA ($\texttt{sum}\leq48$), Task LoRA ($\texttt{sum}\leq96$), and a Geometry LoRA ($\texttt{sum}\leq24$).}
    \label{fig:months-L22-sum-projection-appdx}
\end{figure}

\textbf{Evaluating Fourier Probes.} Figure~\ref{fig:months-period-heatmap} shows the $R^2$ scores for Fourier probes for every layer $l\in\{0,...,31\}$, and period $T\in\{2,...,100\}$ for both Llama-3.1-8B and several fine-tuned variants. 
We compute $R^2$ as the variance of the projected activations explained by the ideal ring after Procrustes alignment---measuring whether points lie on a circle at their correct phases.
Across each setting, the same periods have high $R^2$ scores, but the LoRA fine-tuned models have stronger peaks (higher $R^2$ values) at the base-10 periods: $T \in \{2, 5, 10, 20, 50\}$ than the base model.
This aligns with prior work showing that fine-tuning strengthens existing computations already present in pretrained models \citep{prakash2024finetuning, jain2024mechanistically}. 
Figure~\ref{fig:months-period-sharpening-L18} shows the $R^2$ scores of Fourier planes for layer $18$ (i.e., row $18$ of the heatmap in Figure~\ref{fig:months-period-heatmap}), while Figures~\ref{fig:months-L18-input-projection-appdx}, \ref{fig:months-L18-offset-projection-appdx}, and \ref{fig:months-L18-sum-projection-appdx} show a projection of the months data ($X_{\texttt{sum}\leq96}$) onto the \texttt{input}, \texttt{offset}, and \texttt{sum} Fourier planes at layer $18$ for base-10 periods. 
We include similar plots of the \texttt{sum} for layer $20$: Figures~\ref{fig:months-L20-sum-projection-appdx}, and layer $22$: Figure~\ref{fig:months-L22-sum-projection-appdx}.
Figure~\ref{fig:months-period-sharpening-L22} shows the $R^2$ scores of Fourier planes for layer $22$ (i.e., row $22$ of the heatmap in Figure~\ref{fig:months-period-heatmap}).

\newpage
\subsection{Geometry LoRA Training Details}
\label{appdx:geometry_lora_training}
In this section, we provide additional details about how we train LoRAs with geometric supervision.
Training a geometry LoRA requires making two choices: (1) where to supervise the geometry (i.e., under some readout $\phi$), and (2) what the geometry should look like (i.e., an idealized target geometry $\mathbf{G}$).
For the months task, all geometry LoRAs use a set of Fourier planes whose coordinates lie in a $k$-dimensional subspace $\mathbf{W}$, and select a target structure that matches ideal Fourier plane geometry.

The geometry LoRA results reported in the main paper use a rank-$1$ LoRA trained at the MLP of layer $4$ (i.e., the gate, up, and down projections), and supervised at the final-token residual stream at layer $18$.
We use $\alpha_r = 1.45$ for the initial Procrustes alignment of the target centroids $\boldsymbol{\mu}_v$, a spread ratio target of $\tau = 1.25$, and $\lambda = 0.3$ to balance the two loss terms. 
The frame of each plane (the base model's mean centroid $\boldsymbol{\bar\mu}_c(\theta_0)$, radius $r_0$, and orientation $\mathbf{Q}_0$) is fit once using the base model's projected hidden states over $X_{\texttt{sum}\leq24}$ and held fixed.
We use the same optimizer settings as the task LoRA (\S~\ref{appdx:task_lora_training}) and train for $200$ gradient steps over the same training set used by the task LoRA.
Averaged over three seeds, this geometry LoRA reaches $73.3\pm4.1\%$ accuracy on the months task.

\begin{figure}
    \centering
    \includegraphics[width=0.75\linewidth]{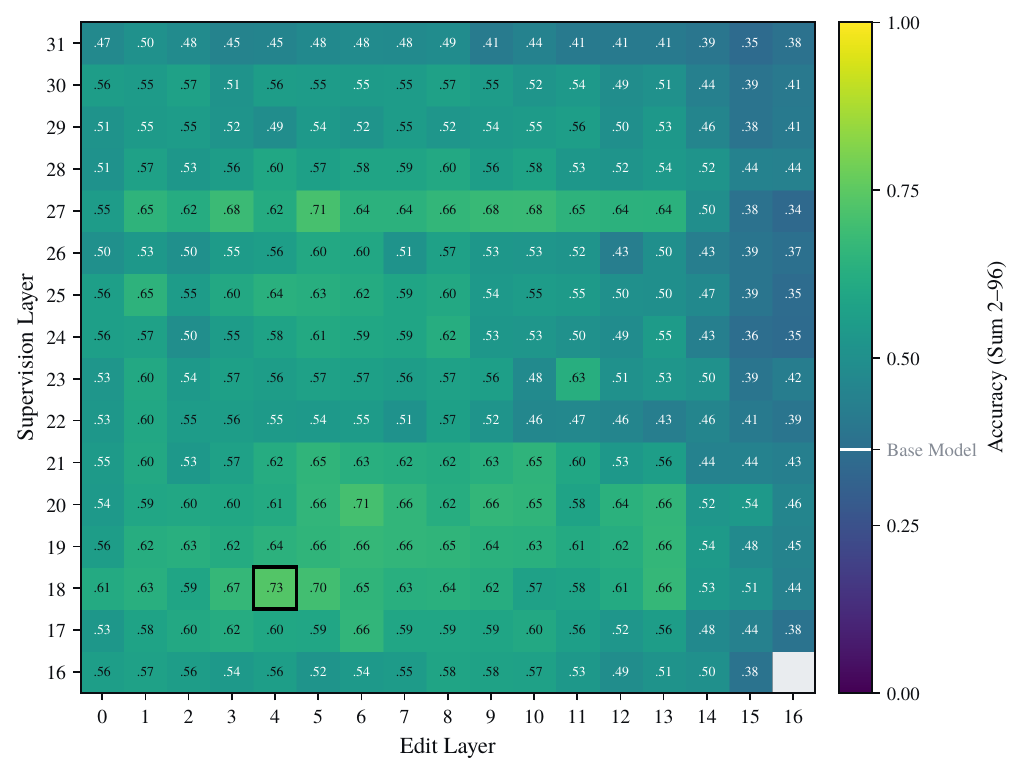}
    \caption{Comparing performance of geometry LoRAs trained and supervised at different layer combinations. The edit layer, where the LoRA is applied, ranges from $0{-}16$ and supervision (where the loss is computed) ranges from layer $16{-}31$. Each cell represents the accuracy of a rank-$1$ geometry LoRA trained on $X_{\texttt{sum}\leq24}$ and scored over the full months dataset: $X_{\texttt{sum}\leq 96}$. 
    The geometry LoRA configuration used in the main paper (edit layer $4$, supervision layer $18$, rank-$1$) is shown with a black box. 
    Supervising before layer $16$ or editing at layer $17$ or later gives little to no improvement over the base model.
    The best edit layers are around layers $3{-}7$, while the best supervision layers are around layers $18{-}20$, with layer $27$ also having strong effects.}
    \label{fig:months_geometry_lora_edit_supervision}
\end{figure}

\textbf{Sweeping over edit layer and supervision layer.} In Figure~\ref{fig:months_geometry_lora_edit_supervision}, we show the results of geometry LoRAs trained and supervised at different layer combinations. The edit layer, where the LoRA update is applied ranges from $0{-}16$, and the supervision layers range from $16{-}31$. The geometry LoRA configuration used in the main paper (edit layer $4$, supervision layer $18$, rank-$1$) is shown with a black box. We find the best edit layers are in the range of $3{-}7$ while the best supervision layers are around $18{-}20$, aligning with the \citet{feucht2026arithmetic}'s localization of the sum computation.

\begin{figure}
    \centering
    \includegraphics[width=0.8\linewidth]{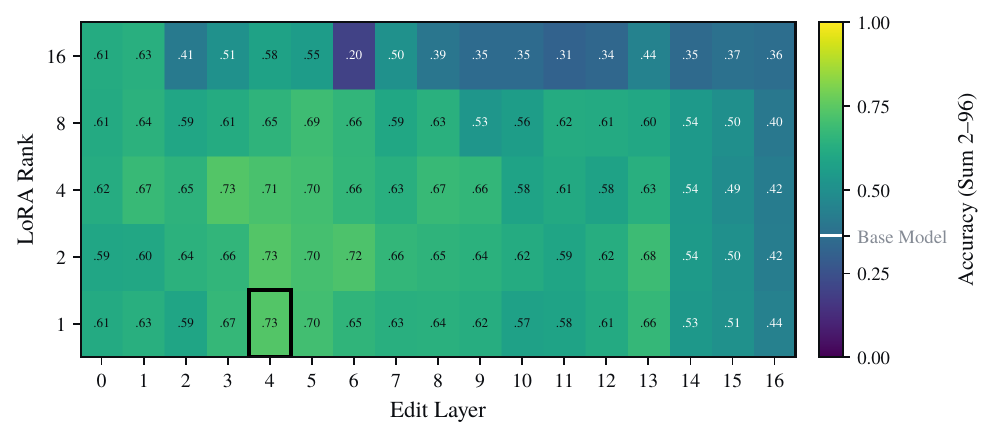}
    \caption{Comparing LoRA rank vs. edit layer. Months task performance of geometry LoRAs trained at different edit layers ($0{-}16$) with different ranks (\{1,2,4,8,16\}) for a fixed supervision layer ($l=18$). 
    We report performance on the full dataset $X_{\texttt{sum}\leq96}$. 
    We find that training a geometry LoRA on the MLP within layers $3{-}6$ has the strongest effect, while rank has little effect (ranks $1{-}4$ achieve similar performance for a given edit layer). 
    The geometry LoRA configuration used in the main paper (edit layer $4$, supervision layer $18$, rank-$1$) is shown with a black box.}
    \label{fig:months_geometry_lora_rank_layer}
\end{figure}
\textbf{LoRA Rank.} In Figure~\ref{fig:months_geometry_lora_rank_layer}, we sweep the LoRA rank over $\{1, 2, 4, 8, 16\}$ at every edit layer $0$--$16$ with geometric supervision fixed at layer $18$. We find that the best cell at each rank lies at edit layers $3$--$6$, with ranks $1$--$4$ achieving similar scores (e.g., at the layer-$4$ MLP they score $73.3\%$, $73.1\%$, and $71.1\%$ overall). Because of this, we use rank-$1$ geometry LoRAs throughout.

\textbf{Geometry LoRA on a single layer vs. multilayer.} In Table~\ref{tab:one_vs_all_layers_geometry}, we compare the effect of training a LoRA on just a single layer to training LoRAs on multiple layers.
For the geometry LoRAs, we fix the supervision layer to layer 18, and report either a single LoRA trained at the layer $4$ MLP, or trained at all layers (MLP, and MLP + Attention) up to the supervision layer (i.e., $0{-}17$). 
Training geometry LoRAs with multiple edit layers follows the opposite trend compared to task LoRAs trained at multiple layers. Training on all layers can improve performance on near-range sums ($\texttt{sum}\in[25{-}48]$) compared to the single layer LoRA, but performance drops on the far-range sums ($\texttt{sum}\in[49{-}96]$) by about the same amount.
\begin{table}[h]
\centering\small
\caption{Comparing single vs. multi-layer LoRA training. Each LoRA is rank-1 and trained on $X_{\texttt{sum}\leq24}$ with only geometric supervision at layer $18$. We report the full accuracy as well as three evaluation bands. The first band is in-range: $\texttt{sum}\in[2,24]$, the other two are held-out ranges: $\texttt{sum}\in[25,48]$, and $\texttt{sum}\in[49,96]$. Scores are mean $\pm$ standard deviation over 3 seeds. Multi-layer LoRAs use a learning rate of $5\times10^{-4}$.}
\label{tab:one_vs_all_layers_geometry}
\begin{tabular}{llcrrrrr}
\toprule
& & & & \multicolumn{4}{c}{Accuracy (\%)} \\
\cmidrule(l){5-8}
Objective & Location & Layers & \# Params. & $\texttt{sum}\in[2,96]$ & $[2,24]$ & $[25,48]$ & $[49,96]$ \\
\midrule
--- & Base model & --- & 0 & 36.3 & 79.0 & 22.9 & 27.4 \\
\midrule
Geometry & MLP & 4 & 55K & 73.3 $\pm$ 4.1 & 96.8 $\pm$ 0.7 & 66.9 $\pm$ 7.1 & 67.9 $\pm$ 3.8 \\
Geometry & MLP & 0--17 & 995K & 67.5 $\pm$ 0.5 & 98.9 $\pm$ 0.3 & 77.1 $\pm$ 0.3 & 51.2 $\pm$ 1.2 \\
Geometry & MLP + Attn & 0--17 & 1.47M & 66.4 $\pm$ 0.3 & 98.6 $\pm$ 0.0 & 78.0 $\pm$ 0.9 & 48.8 $\pm$ 0.6 \\
\bottomrule
\end{tabular}
\end{table}

\textbf{Hyperparameter selection and result reporting.} As seen in the previous sections, both the task LoRA and geometry LoRA have several hyperparameters to select such as the edit layer, supervision layer, rank, etc. 
For simplicity, we report the results of the best task LoRA and best geometry LoRA in the main paper, as the main focus of our paper is to show that geometry can be used as a learning signal.
We find that when using the first held-out cycle (i.e., $\texttt{sum}\in[25,36]$) as a validation split for hyperparameters, the task LoRA and geometry LoRA performance is very similar---the task LoRA selected in this way achieves a total accuracy of $62.8\pm6.5\%$, with $44.7\pm9.6\%$ on the held-out range of $\texttt{sum} \in[37,96]$, while the geometry LoRA selected in this way achieves $62.8\pm1.8\%$ total accuracy, with $45.9\pm2.4\%$ on the held-out range of \texttt{sum}s.
When comparing matched configurations (i.e., same edit layer and rank from layers $0{-}16$ $\times$ ranks $1{-}16$) between the task LoRA and geometry LoRA, we find that the geometry LoRA is actually slightly better than the task LoRA in most of the matched configuration pairs (see Table~\ref{tab:matched_task_geometry}).

\begin{table}[h]
\centering
    \caption{Task LoRA and geometry LoRA accuracy on the months task, compared across matched configurations (same edit layer and rank). There are 85 pairs, spanning layers 0--16 and ranks $\{1,2,4,8,16\}$. The geometry LoRA is better in the majority of matched configurations, especially on the held-out sums.}
    \label{tab:matched_task_geometry}
    \begin{tabular}{lcc}
    \toprule
    & $\texttt{sum} \in[2,96]$ & $\texttt{sum} \in[37,96]$ \\
    \midrule
    Base Model & 36.3\% & 25.6\%\\ 
    Task LoRA (mean over pairs)     & 54.6\% & 35.8\% \\
    Geometry LoRA (mean over pairs) & 57.5\% & 47.3\% \\
    \midrule
    Geometry LoRA better            & 61 / 85  & 73 / 85 \\
    \bottomrule
    \end{tabular}
\end{table}

\subsubsection{Ablations of the Geometric Objective}
\label{appdx:geom_ablations}
\textbf{Per-prompt vs.\ per-value supervision.} The geometry loss in Equation~\ref{eq:loss_mean} aligns empirical per-\texttt{value} centroids $\boldsymbol{\hat\mu}_v(\theta)$ to their targets $\boldsymbol{\mu}_v$.
A natural alternative is to supervise each prompt directly, aligning each projected activation $\phi(\mathbf{h}_x(\theta))$ to its target centroid. This can be achieved by replacing each centroid $\boldsymbol{\hat\mu}_v(\theta)$ in Equation~\ref{eq:loss_mean} with individual activations $\phi(\mathbf{h}_x(\theta))$, and averaging over all prompts in the training set $X_{\texttt{sum}\leq24}$:
\begin{equation}
\label{eq:loss_perprompt}
    \mathcal{L}_{\text{prompt}} = \frac{1}{|X|}\sum_{v\in\mathcal{V}_c}\sum_{x\in X_v}\frac{1}{r_0^{2}}\bigl\lVert \phi(\mathbf{h}_x(\theta)) - \boldsymbol{\mu}_v \bigr\rVert_2^{2}
\end{equation}
where $r_0^2$ is the same as in Section~\ref{sec:notation}.

We compare per-prompt and per-value supervision for training geometry LoRAs and find that both objectives can achieve good performance on the months task. The best per-prompt geometry LoRA achieves 69.6\% accuracy, while the best per-value geometry LoRA achieves 73.4\% accuracy (see Table~\ref{tab:perprompt_ablation}).
We find that per-prompt supervision is more dependent on the spread term ($\mathcal{L}_\text{spread}$; Equation~\ref{eq:loss_spread}) than per-value supervision. Ablating the spread term (setting $\lambda=0$ in Equation~\ref{eq:full_loss}) drops the performance of per-prompt supervision by ${\sim}24.6\%$, from $69.6\%$ to $45.0\%$.
while only dropping performance of per-value supervision by ${\sim}2.2\%$, from $73.4\%$ to $71.2\%$ (see Table~\ref{tab:perprompt_ablation}, $\lambda{=}0$).

\begin{figure}
    \centering
    \includegraphics[width=0.8\linewidth]{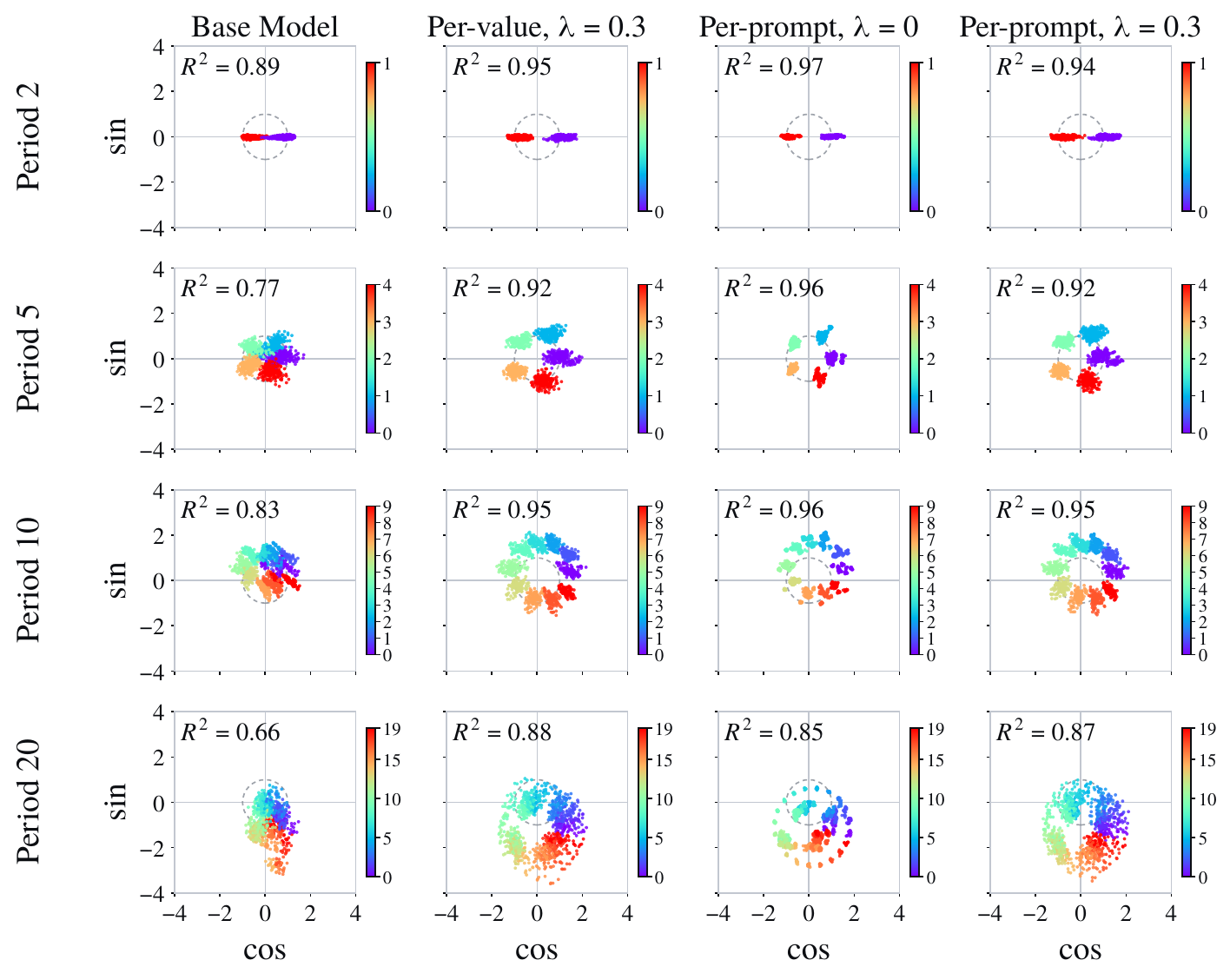}
    \caption{Layer-18 projections of all data ($X_\texttt{sum}\leq96$) onto the \texttt{offset} planes for the base model, a per-value geometry LoRA, and geometry LoRAs trained with the per-prompt loss without ($\lambda = 0$) and with ($\lambda = 0.3$) the spread term, colored by $\texttt{offset} \bmod T$. Per-prompt supervision without the spread term collapses training data to each residue (third column), but introducing the spread term restores the within-class spread (fourth column).}
    \label{fig:perprompt_projections}
\end{figure}

One reason for this dependence on the spread term might be because the per-prompt loss comes with an implicit regularization on the spread that drives each value toward a single point.
We can write $\phi(\mathbf{h}_x(\theta)) - \boldsymbol{\mu}_v = \bigl(\phi(\mathbf{h}_x(\theta)) - \boldsymbol{\hat\mu}_v(\theta)\bigr) + \bigl(\boldsymbol{\hat\mu}_v(\theta) - \boldsymbol{\mu}_v\bigr)$, and because deviations from a centroid sum to zero, the cross term vanishes and the per-prompt loss can be written as the sum of a centroid-matching term and within-value spread term:
\begin{equation}
\label{eq:loss_perprompt_decomp}
    \mathcal{L}_{\text{prompt}}
    = \sum_{v\in\mathcal{V}_c}\frac{|X_v|}{|X|}\,\frac{1}{r_0^{2}}\bigl\lVert \boldsymbol{\hat\mu}_v(\theta) - \boldsymbol{\mu}_v \bigr\rVert_2^{2}
    \;+\;
    \frac{\sigma_c(\theta)^{2}}{r_0^{2}},
\end{equation}
where $\sigma_c(\theta)$ is the within-value spread of Equation~\ref{eq:rms_spread}.
The first term is $\mathcal{L}_{\text{mean}}$ with each value weighted by its number of prompts, and the second is minimized at $\sigma_c(\theta) = 0$, causing the collapse to a single point.
The spread loss term $\mathcal{L}_{\text{spread}}$ in Equation~\ref{eq:loss_spread} instead anchors $\sigma_c(\theta)$ at a ratio of the base model spread ($\tau\cdot\sigma_c(\theta_0)$), which can effectively ``rescue'' the spread-collapsing term in $\mathcal{L}_\text{prompt}$. We see this in Figure~\ref{fig:perprompt_projections}, which projects activations at layer $18$ onto the \texttt{offset} Fourier planes and shows that points collapse to target centroids when $\lambda=0$, but recover for non-zero values of $\lambda$.

\begin{table}[h]
\centering\small
\caption{Per-prompt vs.\ per-value supervision, with and without the spread term, $\mathcal{L}_\text{spread}$ (Eq.~\ref{eq:loss_spread}). All rows use the standard geometry LoRA settings (rank $1$, edit-layer $4$, supervised at layer $18$) and differ only in the loss. No spread term rows are when $\lambda{=}0$. Accuracy is the mean $\pm$ standard deviation over 3 seeds. We report accuracy on the months task, as well as in-range ($\texttt{sum}\in[2,24]$) and held-out ranges ($\texttt{sum}\in[25,48], \texttt{sum}\in[49,96]$). Per-value with $\lambda = 0.3$ is the geometry LoRA reported in the main text (marked with $^*$).}
\label{tab:perprompt_ablation}
\begin{tabular}{llc|ccc}
\toprule
Supervision & $\lambda$ & $\texttt{sum} \in [2,96]$ & $\texttt{sum} \in [2,24]$ & $\texttt{sum} \in [25,48]$ & $\texttt{sum} \in [49,96]$ \\
\midrule
Per-value & 0   & $71.2 \pm 2.8\%$ & $92.5 \pm 2.1\%$ & $63.2 \pm 1.6\%$ & $67.4 \pm 5.3\%$ \\
Per-value$^*$ & 0.3$^*$ & $73.3 \pm 4.1\%$ & $96.8 \pm 0.7\%$ & $66.9 \pm 7.1\%$ & $67.9 \pm 3.8\%$ \\
Per-value & 1   & $73.4 \pm 1.0\%$ & $96.7 \pm 1.3\%$ & $66.7 \pm 1.2\%$ & $68.3 \pm 2.8\%$ \\
Per-value & 3   & $69.8 \pm 2.0\%$ & $93.8 \pm 3.3\%$ & $62.5 \pm 2.1\%$ & $64.8 \pm 1.5\%$ \\
\midrule
Per-prompt & 0   & $45.0 \pm 3.8\%$ & $65.4 \pm 7.7\%$ & $29.3 \pm 3.9\%$ & $45.4 \pm 2.9\%$ \\
Per-prompt & 0.3 & $68.7 \pm 1.9\%$ & $94.3 \pm 2.7\%$ & $62.6 \pm 0.2\%$ & $62.4 \pm 4.6\%$ \\
Per-prompt & 1   & $69.3 \pm 1.8\%$ & $94.0 \pm 0.5\%$ & $60.6 \pm 1.1\%$ & $64.6 \pm 3.9\%$ \\
Per-prompt & 3   & $69.6 \pm 2.7\%$ & $94.6 \pm 2.7\%$ & $62.4 \pm 3.8\%$ & $64.1 \pm 2.3\%$ \\
\bottomrule
\end{tabular}
\end{table}

\begin{figure}
    \centering
    \includegraphics[width=0.95\linewidth]{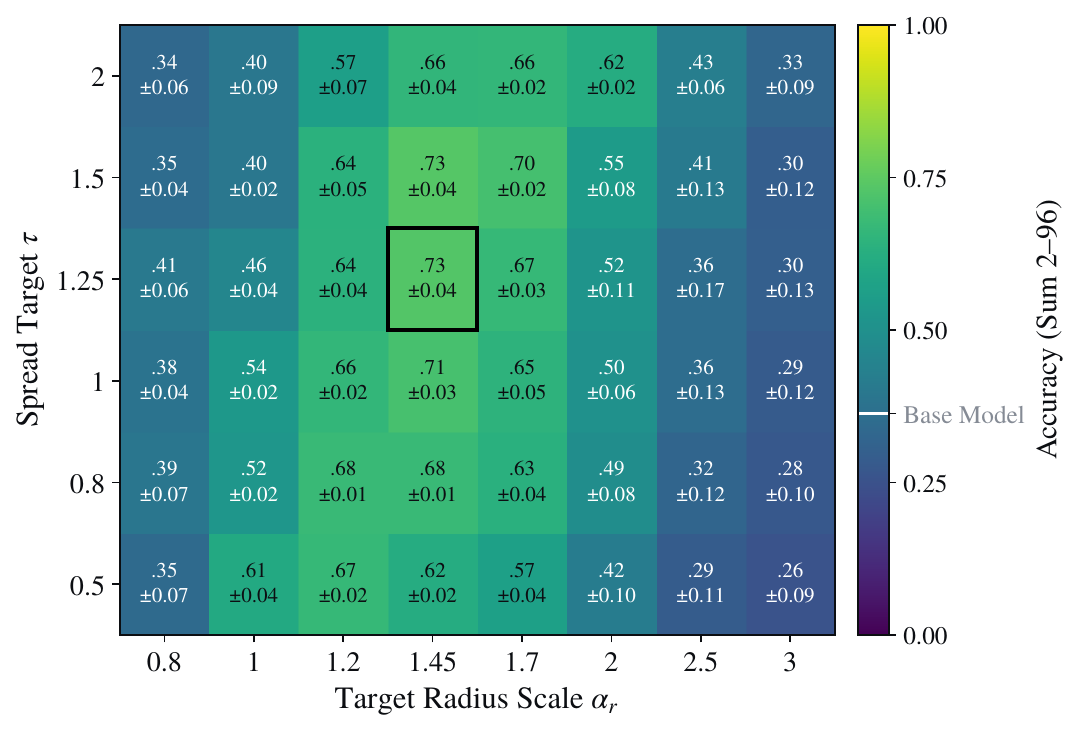}
    \caption{Sensitivity of the geometry LoRA to the radius $\alpha_r$ and the spread target ratio $\tau$. Each cell shows the months-task-performance of a rank-$1$ geometry LoRA trained at layer $4$, supervised at layer $18$, with $\lambda = 0.3$. Accuracy is reported as a mean over $3$ seeds ($\pm$ standard deviation). The black box denotes the geometry LoRA used in the main paper ($\alpha_r = 1.45$, $\tau = 1.25$). We find that performance depends mainly on $\alpha_r$, while $\tau$ has a smaller, but still relevant effect.}
    \label{fig:months_alpha_tau}
\end{figure}

\textbf{Sensitivity to $\alpha_r$ and $\tau$.}
The geometry loss (Equation~\ref{eq:full_loss}) has two hyperparameters: 
(1) the radius $\alpha_r$, which determines how far the target centroids are from the center relative to the base model $f_{\theta_0}$, and (2) the spread ratio $\tau$, which determines the within-value spread relative to the base model. We sweep $\alpha_r \in \{0.8, 1.0, 1.2, 1.45, 1.7, 2.0, 2.5, 3.0\}$ and $\tau \in \{0.5, 0.8, 1.0, 1.25, 1.5, 2.0\}$ using the geometry LoRA training settings described in \S~\ref{appdx:geometry_lora_training}, and report results in Figure~\ref{fig:months_alpha_tau}.

\subsubsection{Fourier Planes as a Supervision Site}
\label{appdx:fourier_supervision}
In this section, we evaluate two design choices: (1) whether supervising geometry on a subspace spanned by trained Fourier probes is a meaningful choice and (2) the effect of supervising an increasing number of Fourier planes.

\textbf{Are Fourier planes a meaningful supervision site?}
We measure whether our choice of the Fourier planes that make up the supervision site $\mathbf{W}$ is meaningful by training a geometry LoRA in the same way on a random subspace of the same dimension.
To supervise our geometry LoRA we use eight Fourier planes ($\{c\in\texttt{input-month}, \texttt{offset}\} \times T \in \{2,5,10,20\}$), thus for our random control we sample eight random orthonormal Gaussian planes and place ideal targets on a ring the same way as we do for the Fourier probe planes (see Equations~\ref{eq:ideal_ring} and ~\ref{eq:ring_target}). We use the same training settings as the standard geometry LoRA, only having substituted our Fourier plane construction of $\mathbf{W}$ for a random projection to a subspace of the same dimension.

We find that a random-subspace geometry LoRA achieves an overall accuracy of $28.4\%$ on the months task, compared to $73.3\%$ when using the Fourier planes, and $36.3\%$ for the base model.  
Similarly, we find that training a geometry LoRA within the Fourier planes, but randomly permuting the value-to-centroid assignment also hurts performance, dropping from $36.3\%$ (base model performance) to $10.2\%$ (fine-tuned).

\begin{figure}
    \centering
    \includegraphics[width=\linewidth]{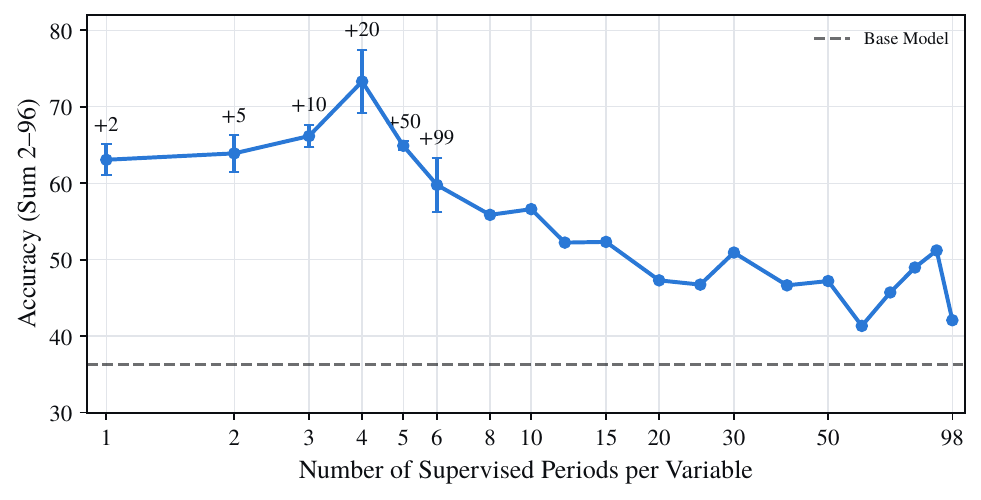}
    \caption{Supervising on many Fourier planes hurts performance. We measure the effect on performance of increasing the number of supervised Fourier planes for geometry LoRAs. Periods are chosen first from $\{2,5,10,20,50\}$ in order, and then by their probe's $R^2$ value. The x-axis is on a log scale, as we test supervising Fourier planes up to 99 periods. We find that supervising more than six periods hurts performance compared to using just a single period ($T=2$), which surprisingly achieves $63.0\%$ accuracy. The best configuration uses four periods: $\mathcal{T}=\{2,5,10,20\}$, which achieves $73.3\%$. The base model's performance is plotted as a dashed line ($36.3\%$)}
    \label{fig:fourier_planes_ladder}
\end{figure}

\textbf{Supervising too many Fourier planes hurts performance.} We measure the effect that training geometry LoRAs with an increasing number of supervised periods (e.g., $\mathcal{T}{=}\{2\}, \mathcal{T}{=}\{2,5\}, \mathcal{T}{=}\{2,5,10\}, ... ,\mathcal{T}{=}\{2, ..., 100\}$) has on performance (see Figure~\ref{fig:fourier_planes_ladder}). 
We fix the supervision layer at layer $l^*=18$, and add periods to the supervision set $\mathcal{T}$ by first including the base-10 Fourier features (2, 5, 10, 20, 50) in order, and then selecting others based on their $R^2$ value (see Figure~\ref{fig:months-period-sharpening-L18}).
We find that using an increasing number of periods for geometric supervision hurts performance, and that four periods: $\mathcal{T}{=}\{2,5,10,20\}$ is the best supervision set. Surprisingly, supervising only a single Fourier plane ($\mathcal{T}{=}\{2\}$) already achieves around $63.0\%$ accuracy, while using planes of four periods gets $73.3\%$ accuracy.

\subsubsection{Alternate Forms of Months Variable Supervision}
\label{appdx:alternate_months_supervision}
In this section, we test how much of the geometry LoRA's improvement comes from supervising the right variables, and how much comes from the idealized Fourier geometry. We compare the Fourier idealization described in the main paper to three alternate training objectives: (1) direct regression of the variable values (\texttt{input} and \texttt{offset}) through linear probes (e.g., training a LoRA so that a linear probe reads out $11$ for an \texttt{offset} of \textit{eleven}), (2) placing each variable's values evenly spaced along a line under the same linear probes, and (3) using the base model's centroids in the Fourier planes as targets instead of idealized circular coordinates.

\textbf{Setup.} Each baseline uses the same LoRA settings (rank-$1$, edit layer $4$, supervision layer $18$) and training data ($X_{\texttt{sum}\leq24}$), as the geometry LoRA (\S\ref{appdx:geometry_lora_training}), and changes only the readout $\phi$, or the target $\mathbf{G}$.
For (1) and (2), we fit one linear probe per variable, $z_c(\mathbf{h}) = \mathbf{w}_c^\top\mathbf{h} + b_c$, by ridge regression of the variable's value on the layer-$18$ hidden states of the training prompts.
We fit these probes on the months prompts because we found linear probes fit on the addition prompts, do not read out these variables on the months prompts (unlike the Fourier probes, which do transfer reasonably well; \S\ref{appdx:fourier_probes}) and training against them does not improve over the base model ($19.0\%$ for direct regression and $29.0\%$ for the evenly spaced line). We describe the baselines below:
\begin{itemize}
    \item[(1)] \textbf{Direct regression.} We train with the per-prompt loss (Equation~\ref{eq:loss_perprompt}) on each probe's readout $z_c$, with the literal variable value as the target ($\mu_v = v$) and no spread term.
    \item[(2)] \textbf{Evenly spaced line.} We train with the per-value geometry loss (Equation~\ref{eq:full_loss}) on $z_c$, where the targets are evenly spaced, centered at the base model's mean centroid, and scaled to radius $\alpha_r r_0$ (i.e., Equation~\ref{eq:ring_target}, but the ring $\mathbf{u}_v$ is replaced by evenly spaced points on a line).
    \item[(3)] \textbf{Base model centroids.} We use the same eight Fourier planes and loss as the geometry LoRA, but replace the ideal ring with the base model's own centroids scaled about their mean, $\boldsymbol{\mu}_v = \boldsymbol{\bar\mu}_c(\theta_0) + \alpha_r\bigl(\boldsymbol{\hat\mu}_v(\theta_0) - \boldsymbol{\bar\mu}_c(\theta_0)\bigr)$. This target has the same center and radius ($\alpha_r r_0$) as the ring, so it differs from our method only in the shape of the target.
\end{itemize}

\textbf{Results.} For each method, we report the best configuration averaged over three seeds, and for (1) and (2), we also vary which variables are supervised. The results are contained in Table~\ref{tab:alternate_months_supervision}. Each of the baselines supervising the \texttt{input} and \texttt{offset} improves over the base model, achieving $45.9\%$, $60.5\%$, and $61.9\%$ respectively, compared to the base model's $36.3\%$ (see Table~\ref{tab:alternate_months_supervision}). For the linear baselines, supervising the offset seemed to be about as effective as both variables together.
However, none of them outperform the idealized circular Fourier geometry which achieves $73.3\%$. 
This shows that supervising the right variables can be helpful, but that the ideal structure also contributes additional gains not captured by these baselines. 

\begin{table}[h]
\centering
\caption{Months task accuracy ($\texttt{sum}\in[2,96]$) for alternate forms of variable supervision. We report the best configuration of each method, as the mean $\pm$ standard deviation over 3 seeds. The base-model centroids and the idealized ring are both in the same Fourier planes as the geometry LoRA.}
\label{tab:alternate_months_supervision}
\begin{tabular}{lcccc}
\toprule
& \multicolumn{4}{c}{Supervised Variables} \\
\cmidrule(lr){2-5}
Method & None & \texttt{input} & \texttt{offset} & \texttt{input} + \texttt{offset} \\
\midrule
Base Model             & $36.3\%$ & --               & --                & -- \\
\midrule
Direct Regression      & --  & $40.4 \pm 1.1\%$ & $52.2 \pm 10.7\%$ & $45.9 \pm 5.5\%$ \\
Evenly Spaced Line     & --  & $45.7 \pm 2.1\%$ & $59.7 \pm 2.5\%$  & $60.5 \pm 3.9\%$ \\
Base-Model Centroids   & --  & $56.4\pm10.2\%$  & $63.2\pm0.9\%$        & $61.9 \pm 1.9\%$ \\
\midrule
Geometry LoRA (Idealized Circles)  & --  & $63.9\pm2.6\%$ & $67.6\pm7.1\%$ & $\mathbf{73.3 \pm 4.1\%}$ \\
\bottomrule
\end{tabular}
\end{table}

\subsection{Activation Patching Details}
\label{appdx:subspace_difference}
In this section, we provide additional details about the months task activation patching experiments mentioned in Section~\ref{sec:fourier_patching_not_sufficient}.
We find that patching a fine-tuned model's Fourier-plane coordinates into the base model does not transfer the accuracy gains that fine-tuning achieves ($36.3\%{\to}37.3\%$), but that patching a low-dimensional difference in activations can recover most of it ($36.3\%{\to}73.0\%$). Here, we study the difference in hidden states between the base model and the fine-tuned model to understand what other changes are encoded in the fine-tuned model's supervision layer activations that enable this recovery.

\textbf{Patching Setup.} For every prompt $x \in X_{\texttt{sum}\leq 96}$ we take the final-token hidden state at layer $18$ from the base model, $\mathbf{h}_x(\theta_0)$, and from a fine-tuned model, $\mathbf{h}_x(\theta)$, and construct a matrix of differences $\mathbf{D}$ with rows: $\mathbf{h}_x(\theta) - \mathbf{h}_x(\theta_0)$. We compute the SVD decomposition of $\mathbf{D}$ and take the top $k$ right-singular vectors, $\mathbf{V}_k$, which are the $k$ directions along which the two models' activations differ the most, and denote $\mathbf{V}_k\mathbf{V}_k^\top$ as the subspace projection.
We can then perform subspace-level interventions by patching activations from the fine-tuned model to replace the base model's activations within the targeted subspace. We write this intervention as: 
\begin{equation}
\label{eq:subspace_patching}
  \mathbf{h}_x(\theta_0) \leftarrow \mathbf{V}_k\mathbf{V}_k^\top\mathbf{h}_x(\theta) + (\mathbf{I} - \mathbf{V}_k\mathbf{V}_k^\top)\mathbf{h}_x(\theta_0).
\end{equation}
We test the effect of patching activations taken from the task LoRA and the geometry LoRA for varying values of $k$ and report the results in Table~\ref{tab:diff_rank}. 
We find that patching in activations from the fine-tuned model into the base model along the top-$k$ difference directions is able to recover a large portion of the performance. Patching along just $k{=}16$ directions recovers ${\sim}91{-}94\%$ of the fine-tuned models performance. Patching within a random subspace has no effect from either LoRA ($37.2\%$ vs $36.3\%$).

\begin{table}[h]
\centering
\small
\caption{Replacing the base model activations with the fine-tuned model's at layer $18$ coordinates in the top-$k$ difference directions. Each column shows the performance when patching within the top-$k$ directions, while the fine-tuned column shows the fine-tuned model's performance itself, without patching interventions. For the random row, we select a random rank-matched subspace to patch into for comparison. Patching $k{=}16$ difference dimensions recovers ${\sim}91{-}94\%$ of the fine-tuned model performance. For reference, the base model achieves $36.3\%$ accuracy on this task.}
\label{tab:diff_rank}
\begin{tabular}{l ccccccc @{\hskip 1.5em} |c}
\toprule
 & \multicolumn{7}{c}{Patched top-$k$ directions} & \\
\cmidrule(lr){2-8}
Source & $k{=}1$ & $k{=}2$ & $k{=}4$ & $k{=}8$ & $k{=}16$ & $k{=}32$ & $k{=}64$ & Fine-tuned \\
\midrule
Task LoRA & 54.1\% & 52.1\% & 57.3\% & 61.3\% & 65.5\% & 68.4\% & 71.3\% & 71.9\% \\
Geometry LoRA & 49.2\% & 49.8\% & 53.7\% & 59.8\% & 68.9\% & 71.7\% & 73.0\% & 73.3\% \\
\midrule
Random, any source & 36.4\% & 36.4\% & 36.4\% & 36.5\% & 36.5\% & 36.7\% & 37.2\% & -- \\
\bottomrule
\end{tabular}
\end{table}

\subsubsection{What does the activation difference encode?}
Table~\ref{tab:diff_rank} shows that we can patch a small subspace within the hidden state of the fine-tuned model into the base model and recover most of the fine-tuning performance. 
In this section, we try to understand what these directions encode, and focus on the $k{=}16$ case. To do so, we test whether the activations we patch within this subspace correspond to one of the months task variables $c \in \{\texttt{input}, \texttt{offset}, \texttt{sum}\}$. 
We group prompts by the value of a variable and measure the variance of $\mathbf{D}$ that the group mean can explain. We find the \texttt{input} can explain $13$--$15\%$ of the variance, while the \texttt{sum} explains $44\%$ of the variance and the \texttt{offset} explains $64\%$ for the geometry LoRA ($47\%$ and $73\%$ for the task LoRA). 

\textbf{Synthetic patching functions.} To test whether the subspace can be described as causally carrying one of these variables, we check whether we can replace the fine-tuned model's subspace coordinates $\mathbf{c}_x(\theta) = \mathbf{V}_k\mathbf{V}_k^\top\mathbf{h}_x(\theta)$, with a fitted function of the prompt corresponding to a variable, or a variation of one.
For example, we compute the per-\texttt{sum} mean ($\mathbf{c}_x(\theta) \leftarrow \mathbf{\bar c}_{x}(\theta) = \frac{1}{|X_{\texttt{sum}=v}|}\sum_{x\in X_{\texttt{sum}=v}}{\mathbf{c}_x(\theta)}$), which is the average subspace coordinates over all prompts that share a \texttt{sum} value $v$ and patch it into the base model for all prompts with that same value $\texttt{sum}(x){=}v$, and measure performance doing this over all prompts. 
We similarly compute the per-\texttt{offset} and per-\texttt{input} means, as well as a swap function where we replace the coordinates $\mathbf{c}_x$, with those of another prompt that shares the same variable value (e.g., patching from ``\textit{What month is four months after June?}'' into ``\textit{What month is four months after April?}'' because they share the same \texttt{offset}).

We construct and evaluate these functions by patching the fitted values into the base model, reporting the results over $X_{\texttt{sum}\leq96}$ in Table~\ref{tab:diff_synthetic}.
We find that patching the per-\texttt{sum} mean within the difference direction subspace recovers $82.5{-}87.4\%$ of the fine-tuned model's performance for $k=16$, suggesting that what the difference encodes is primarily a representation of the \texttt{sum}. Replacing the hidden state within the subspace with coordinates derived from the other variables either have no effect or make the base model performance worse. Thus, we find that supervising the \texttt{input} and \texttt{offset} Fourier planes to have cleaner geometry (\S\ref{sec:months_geometric_alignment}) also appears to cause the model to construct a cleaner representation of the sum that resides outside of the supervised Fourier plane frame, showing beneficial off-target effects of fine-tuning that are not captured by the representations we target for supervision.

\begin{table}
\centering
\small
\caption{Patching constructed vectors into the top-$16$ difference directions at layer $18$. We report the overall accuracy over three LoRA seeds. For reference, the base model achieves $36.3\%$ accuracy on the months task, the task LoRA achieves $71.9\%$, and the geometry LoRA achieves $73.3\%$. We find that patching the per-\texttt{sum}-derived activations is better at recovering most of the fine-tuning gains compared to \texttt{offset}-, or \texttt{input}-derived fitted activations.}
\label{tab:diff_synthetic}
\begin{tabular}{lcc}
\toprule
Written into the subspace & Task LoRA & Geometry LoRA \\
\midrule
Fine-tuned model, no patching & 71.9\% & 73.3\% \\
\midrule
Fine-tuned model's own component & 65.5\% & 68.9\% \\
\midrule
\multicolumn{3}{l}{\textit{Functions of the sum}} \\
Per-\texttt{sum} mean & 62.9\% & 60.5\% \\
Another prompt with the same sum & 59.8\% & 56.4\% \\
\midrule
\multicolumn{3}{l}{\textit{Functions of the operands}} \\
Per-\texttt{offset} mean & 38.5\% & 22.2\% \\
Per-\texttt{input} mean & 31.8\% & 20.7\% \\
Another prompt with the same \texttt{input} & 16.6\% & 13.1\% \\
Another prompt with the same \texttt{offset} & 19.1\% & 13.8\% \\
\bottomrule
\end{tabular}
\end{table}

\subsection{Evaluating Off-Target Effects of Fine-tuning}
\label{appdx:months-specificity}
In this section, we evaluate task and geometry LoRAs on several external tasks to measure their effect on the model's general capabilities.
These include: FineWeb~\citep{penedo2024fineweb}, WikiText-103~\citep{merity2017pointer}, MMLU~\citep{hendrycks2021measuring}, and GSM8K~\citep{cobbe2021trainingverifierssolvemath}.
On FineWeb, we measure bits per byte (bpb) over 600 documents. On WikiText-103, we measure perplexity (ppl). For MMLU we evaluate on the full test set using 5-shot prompting and report the accuracy ($\%$). For GSM8K, we evaluate on the full test set using 8-shot chain-of-thought with greedy decoding and compute the exact match accuracy on the final output number.
We find that the geometry LoRA and task LoRAs have little to no effect on general model performance as measured by these other tasks (see Table~\ref{tab:offtarget_benchmarks}).

\begin{table}
\centering\small
\caption{Off-task effects of fine-tuning on the months task. We evaluate Llama-3.1-8B and LoRA fine-tuned variants on several LM benchmark tasks: FineWeb, WikiText-103, MMLU, and GSM8K to measure the effect of fine-tuning on external tasks.
Rows reporting LoRA results are means over 3 seeds.}
\label{tab:offtarget_benchmarks}
\begin{tabular}{lccccc}
\toprule
 & Months & FineWeb & WikiText & MMLU & GSM8K \\
Model & Acc. ($\%$) $\uparrow$ & bpb $\downarrow$ & ppl $\downarrow$ & ($\%$) $\uparrow$ & ($\%$) $\uparrow$ \\
\midrule
Base Model                & 36.3  & 0.709 & 7.75 & 65.4 & 52.5 \\
Geometry LoRA ($\leq 24$) & 73.3  & 0.709 & 7.75 & 65.5 & 52.0 \\
Task LoRA ($\leq 24$)     & 71.9  & 0.710 & 7.78 & 65.0 & 53.5 \\
\midrule
Oracle Task LoRA ($\leq 96$)     & 100.0 & 0.709 & 7.75 & 65.1 & 51.9 \\
\bottomrule
\end{tabular}
\end{table}

\subsection{Comparing Geometric Supervision to Other Baselines}
\label{appdx:other_baselines}
In this section, we report additional baselines on the months task: few-shot prompting (ICL), standard fine-tuning, and Constructive Circuit Amplification (CCA)~\citep{prakash2025cca}, a targeted fine-tuning method that locates and then trains task-specific circuits.

For each baseline/control, we use $X_{\texttt{sum}\leq24}$ as our training set.
For few-shot prompting, we test using $\{4,8,32\}$ in-context examples.
CCA has two settings, where a learned mask is applied (``w/ mask'') or not (``w/o mask'').

\begin{table}
\centering\small
\setlength{\tabcolsep}{5pt}
\caption{Additional methods evaluated on the months task: few-shot prompting, full fine-tuning, and CCA. Few-shot examples are drawn from $\texttt{sum}\leq24$ and excluded from scoring when present in the prompt. Full fine-tuning uses the task loss with the learning rate swept over $10^{-6}$--$10^{-3}$ (best rate, final step, one seed). LoRA and CCA rows are means over 3 seeds.}
\label{tab:months_baselines}
\begin{tabular}{llr|rrr}
\toprule
Method & \# params & $\texttt{sum} \in [2,96]$ & $[2,24]$ & $[25,48]$ & $[49,96]$ \\
\midrule
Base model & -- & 36.3\% & 79.0\% & 22.9\% & 27.4\% \\
\midrule
Few-shot, 4 examples & -- & 44.4\% & 95.1\% & 44.8\% & 25.7\% \\
Few-shot, 8 examples & -- & 41.2\% & 77.7\% & 35.4\% & 30.7\% \\
Few-shot, 32 examples & -- & 40.6\% & 83.7\% & 33.3\% & 28.5\% \\
\midrule
CCA, w/ mask \citep{prakash2025cca} & 296M & 55.5\% & 100.0\% & 69.1\% & 32.5\% \\
CCA, w/o mask \citep{prakash2025cca} & 8.03B & 62.7\% & 100.0\% & 81.2\% & 39.8\% \\
\midrule
Full fine-tuning, all layers & 8.03B & 63.2\% & 100.0\% & 78.5\% & 42.2\% \\
Full fine-tuning, layers 0--15 & 3.49B & 62.7\% & 100.0\% & 76.7\% & 42.0\% \\
Full fine-tuning, layers 16--31 & 3.49B & 49.8\% & 100.0\% & 57.3\% & 27.8\% \\
\midrule
Task LoRA, rank 1, layer 4 & 55K & 54.3\% & 100.0\% & 68.2\% & 30.7\% \\
Task LoRA, rank 8, layer 3 & 442K & 71.9\% & 100.0\% & 78.0\% & 58.6\% \\
\midrule
Geometry LoRA, rank 1, layer 4 & 55K & 73.3\% & 96.8\% & 66.9\% & 67.9\% \\
\bottomrule
\end{tabular}
\end{table}

\section{Linebreaking Task: Experimental Details}

\subsection{Linebreaking Dataset Examples}
\label{appdx:linebreak_dataset}
In this section, we provide a few examples of wrapped FineWeb~\citep{penedo2024fineweb} documents from our linebreaking dataset. See \S~\ref{sec:linebreak} for dataset construction details.
Figure~\ref{fig:linebreak_example_taxi} shows an example of a document being wrapped both at $m{=}20$ characters per line, and $m{=}40$ characters per line. 
Figure~\ref{fig:linebreak_example_seasonal} shows an example of a document wrapped to $m{=}80$ characters.

\colorlet{nlcolor}{black}
\begin{figure}[h]
\centering
\tcbset{lbbox/.style={colback=blue!2, colframe=blue!50!black,
  fonttitle=\bfseries\small\boldmath, fontupper=\ttfamily\footnotesize,
  boxrule=0.5pt, arc=2pt, left=3mm, right=3mm,
  nobeforeafter, box align=top, equal height group=linebreak}}%
\begin{tcolorbox}[lbbox, width=0.38\linewidth, title={Wrapped to $m{=}20$ characters}]
During the spring{\color{nlcolor}\textbackslash n}\\
and summer of 1946{\color{nlcolor}\textbackslash n}\\
two groups of WWII{\color{nlcolor}\textbackslash n}\\
Veterans held{\color{nlcolor}\textbackslash n}\\
meetings to discuss{\color{nlcolor}\textbackslash n}\\
creating new taxi{\color{nlcolor}\textbackslash n}\\
companies. One group{\color{nlcolor}\textbackslash n}\\
called themselves{\color{nlcolor}\textbackslash n}\\
Tuxedo Cab, the{\color{nlcolor}\textbackslash n}\\
other Radio Cab.\ \dots
\end{tcolorbox}\hfill
\begin{tcolorbox}[lbbox, width=\dimexpr0.60\linewidth-5mm\relax, title={Wrapped to $m{=}40$ characters}]
 During the spring and summer of 1946 two{\color{nlcolor}\textbackslash n}\\
  groups of WWII Veterans held meetings to{\color{nlcolor}\textbackslash n}\\
  discuss creating new taxi companies. One{\color{nlcolor}\textbackslash n}\\
  group called themselves Tuxedo Cab, the{\color{nlcolor}\textbackslash n}\\
  other Radio Cab. Each group appeared{\color{nlcolor}\textbackslash n}\\
  before the Portland City Council to{\color{nlcolor}\textbackslash n}\\
  request taxi company operating permits.{\color{nlcolor}\textbackslash n}\\
  City Council told them to leave, join{\color{nlcolor}\textbackslash n}\\
  forces and come back as one company.{\color{nlcolor}\textbackslash n}\\
  They got together, appeared before\ \dots
\end{tcolorbox}
\caption{An example document from the linebreaking dataset. The same FineWeb passage (a Radio Cab company history) is wrapped to width $m=20$ (left) and $m=40$ (right): a newline ({\color{nlcolor}\textbackslash\texttt{n}}) is inserted at the last word boundary that keeps each line to at most $m$ characters.}
\label{fig:linebreak_example_taxi}
\end{figure}

\begin{figure}
\centering
\begin{tcolorbox}[colback=blue!2, colframe=blue!50!black,
  fonttitle=\bfseries\small\boldmath, fontupper=\ttfamily\footnotesize,
  boxrule=0.5pt, arc=2pt, left=3mm, right=3mm, width=\linewidth,
  title={Wrapped to $m{=}80$ characters}]
September is a confusing time for seasonal purists. I'm not saying I'm one, but{\color{nlcolor}\textbackslash n}\\
I do tend to measure the progress of the seasons by what I eat and at the middle{\color{nlcolor}\textbackslash n}\\
point of this month I am always conflicted. There are still a great many fruits{\color{nlcolor}\textbackslash n}\\
available that I associate with summer (berries, grapes, figs and melons to name{\color{nlcolor}\textbackslash n}\\
a few) but what happens in September is we see an overlap into the fall harvest{\color{nlcolor}\textbackslash n}\\
season. While the growers of summer are still churning out some of the best{\color{nlcolor}\textbackslash n}\\
fruit of the season, apple and pear growers throughout the U.S. are entering{\color{nlcolor}\textbackslash n}\\
into the game and the competition at farmers' markets and grocery stores for\ \dots
\end{tcolorbox}
\caption{A second example from the linebreaking dataset (a FineWeb passage on seasonal eating), wrapped to width $m=80$. As before, a newline ({\color{nlcolor}\textbackslash n}) is inserted at the last word boundary that keeps each line to at most $m$ characters.}
\label{fig:linebreak_example_seasonal}
\end{figure}

\subsection{Training Task LoRAs for Linebreaking}
\label{appdx:linebreak_task_loras}
In this section, we provide additional details about how we train task LoRAs for the linebreaking task. All task LoRAs are trained with a standard next-token cross-entropy loss at every token position.
We train a rank-$1$ LoRA on the attention and MLP projections of layers $0{-}5$, and the LoRA update is applied at every token position.
We use the AdamW optimizer~\citep{loshchilov2018decoupled} with a learning rate of $1\times10^{-4}$, and a constant learning rate schedule, and we train for $400$ steps with a batch size of $4$ over $1{,}600$ FineWeb documents, each wrapped to a random width $m\in[20,150]$.
The base model is Llama-3.1-8B in \texttt{bf16} with weights kept frozen, while the LoRA parameters are kept in \texttt{fp32}.
We use this same LoRA placement---rank, layers, optimizer, and training data---for both the task and geometry LoRAs, so that the comparison between them isolates the loss function: the two differ only in whether the objective is the output-level cross-entropy or geometric supervision (\S\ref{appdx:helix_geometry_lora}).

\subsection{Training Geometry LoRAs for Linebreaking with Helix-Supervision}
\label{appdx:helix_geometry_lora}
In this section, we provide additional details about how we train geometry LoRAs for the linebreaking task.
As with the months task, we must choose (1) where to supervise the geometry (i.e., a readout $\phi$), and (2) what the geometry should look like (i.e., an idealized target geometry $\mathbf{G}$).
For the linebreaking task, the readout we use $\mathbf{W}\in\mathbb{R}^{3\times d}$ is a helix frame: two Fourier probes $\mathbf{w}_{\cos},\mathbf{w}_{\sin}$ trained to match $\cos(2\pi\,\texttt{character-count}(x_t)/T)$ and $\sin(2\pi\,\texttt{character-count}(x_t)/T)$ at period $T$, and the linear \texttt{character-count} probe $\mathbf{w}_{\texttt{count}}$, and we choose the target structure $\mathbf{G}$ to be the ideal helix geometry.
We fit all three probes on the base model using tokens from FineWeb documents wrapped to a single width $m=150$ over counts $20{-}150$, with ridge regression.

The idealized target places each \texttt{character-count} value $v$ on the ideal helix point $\mathbf{u}_v = (\cos(2\pi v/T),\ \sin(2\pi v/T),\ v/T)$, which we align to the base model's per-\texttt{character-count} centroids (computed over counts $1{-}150$) via orthogonal Procrustes once before training and keep fixed throughout (Equation~\ref{eq:ring_target}).
The geometry LoRA results reported in the main paper use a rank-$1$ LoRA trained on the attention and MLP projections of layers $0{-}5$. We use a radius of $\alpha_r=1.2$, a spread ratio target of $\tau=0.5$, and $\lambda=1.0$ to balance the two loss terms.\footnote{We apply the spread term separately to each \texttt{character-count} value and average over values.} We use the same optimizer settings, training data, and number of steps as the task LoRA (\S\ref{appdx:linebreak_task_loras}), and supervise at each token at layer $l^*=5$. 

\textbf{Choosing the period $T$.} For the helix structure, we set $T=155$, the period at which the base model's \texttt{character-count} representation completes a single rotation in the Fourier plane.
We determine this by taking the first two PCs of the per-count means (i.e. the two rotating components in Figure~\ref{fig:linebreak_count_representation}b) and compute the phase of each count as $\theta_v = \operatorname{atan2}(\mathrm{PC}_2, \mathrm{PC}_1)$. We unwrap the phase across counts and fit a line to predict the phase by least squares: $\theta_v = \omega v + b$. The period is then computed as $T = 2\pi/\omega$, which yields $T\approx154.77$ characters.

\subsubsection{Hyperparameter Sweeps for the Linebreaking Geometry LoRA}
\label{appdx:helix_hparam_sweeps} 
Here we provide additional details about how the linebreaking geometry LoRA depends on its main hyperparameters: the LoRA rank, the layer at which we supervise the geometry, and the two terms of the idealized target---the radius $\alpha_r$ and the spread ratio $\tau$. In each sweep, we vary a single axis and hold all other settings fixed at the configuration reported in the main paper (rank-$1$, edit layers $0$--$5$, supervision at layer $5$, $\alpha_r{=}1.2$, $\tau{=}0.5$, $\lambda{=}1.0$), and score each LoRA by its linebreaking error. Values are reported over three seeds.
We note that when comparing task LoRA and geometry LoRA for this task, we use the same training setup and hyperparameters (e.g., edit layer and rank) for a fair comparison. We select the configuration reported in the paper based on a validation set that is similar to, but separate from the held-out documents.

\begin{figure}
    \centering
    \includegraphics[width=\linewidth]{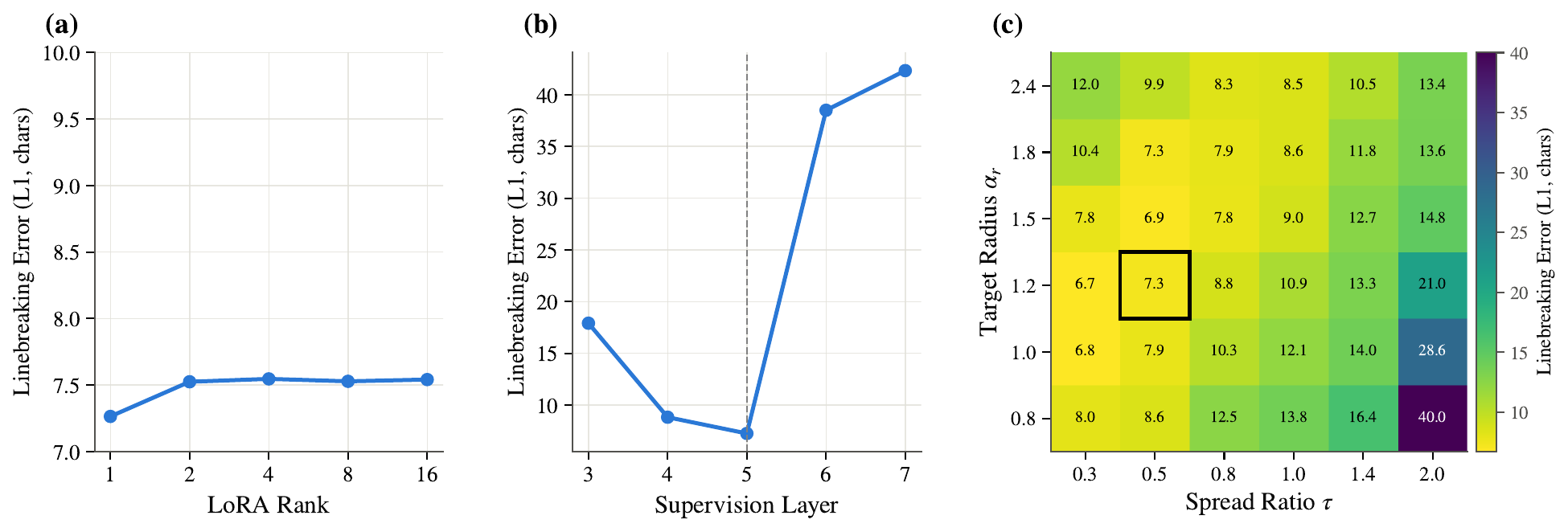}
    \caption{Linebreaking error of the geometry LoRA under three hyperparameter sweeps.
    \textbf{(a)} LoRA rank and \textbf{(b)} supervision layer (where the LoRA edits blocks $0$--$l$ and the geometric loss is applied at block $l$). The dashed line marks the reported supervision layer $l^*{=}5$. \textbf{(c)} Target radius $\alpha_r$ (rows) against spread ratio $\tau$ (columns), with each cell's linebreaking error (L1). The configuration used in the main paper is outlined in black ($\alpha_r{=}1.2$, $\tau{=}0.5$).}
    \label{fig:helix_hparam_sweeps}
\end{figure}

\textbf{LoRA rank.} In Figure~\ref{fig:helix_hparam_sweeps}a, we sweep the LoRA rank over $\{1,2,4,8,16\}$ with the supervision layer fixed at layer $5$. We find that rank has little effect: rank-$1$ achieves the lowest error (7.3 characters) and higher ranks are within ${\sim}0.3$ characters ($7.5$--$7.6$). Because of this, we use rank-$1$ geometry LoRAs throughout.

\textbf{Supervision layer.} In Figure~\ref{fig:helix_hparam_sweeps}b, we sweep the
supervision layer over blocks $3$--$7$, where a LoRA supervised at block $l$ edits blocks $0$--$l$. The linebreaking error is lowest when supervising at layer $5$ ($7.3$ characters), where the \texttt{character-count} linear probe peaks (Figure~\ref{fig:linebreak_count_representation}a). 

\textbf{Sensitivity to $\alpha_r$ and $\tau$.} The geometric loss has two hyperparameters: (1) the radius $\alpha_r$, which determines how far the target centroids lie from the center relative to the base model $f_{\theta_0}$, and (2)  $\tau$, which sets the within-\texttt{character-count} spread ratio. We sweep $\alpha_r\in\{0.8,1.0,1.2,1.5,1.8,2.4\}$ and $\tau\in\{0.3,0.5,0.8,1.0,1.4,2.0\}$ and report results in Figure~\ref{fig:helix_hparam_sweeps}c. 
The error is low for a range of values: an $\alpha_r{\approx}1.0$--$1.5$ and a small spread ratio ($\tau\approx0.3$--$0.5$) achieves a linebreaking error of ${\sim}6.7{-}7.3$ characters. The values used in the main paper ($\alpha_r{=}1.2$, $\tau{=}0.5$) are outlined in black. 

\textbf{Ablating the spread term.} The geometric objective (Equation~\ref{eq:full_loss}) balances the centroid-matching term $\mathcal{L}_{\text{mean}}$ against the within-\texttt{character-count} spread term $\mathcal{L}_{\text{spread}}$ through the weight $\lambda$. For the months task, removing the spread term collapses per-prompt supervision (Appendix~\ref{appdx:geom_ablations}); here we measure its effect on the linebreaking geometry LoRA, which supervises per-\texttt{character-count} centroids. We find that including the spread term helps a little, as setting $\lambda=0$ (no spread term) and retraining with the otherwise-canonical configuration raises the average linebreaking error from $7.3$ to $7.8$ characters. 

\subsubsection{Alternate Forms of Linebreaking Supervision}
\label{appdx:alternate_linebreak_supervision}
Similar to the months task (Appendix~\ref{appdx:alternate_months_supervision}), we test how much of the geometry LoRA's improvement on the linebreaking task comes from supervising the \texttt{character-count} variable, and how much comes from the idealized helix. We compare the helix supervision described in the main paper to several alternate training objectives: (1) direct regression of the \texttt{character-count} through the linear count probe, (2) placing the \texttt{character-count} values evenly spaced along a line under the same probe, (3) supervising only the Fourier coordinates of the helix, without the count coordinate, (4) using the base model's per-\texttt{character-count} centroids in the helix frame as targets instead of the idealized helix, and (5) using the base model's centroids in the top-$3$ PCA subspace of the \texttt{character-count} representation as targets.

\textbf{Setup.} Each baseline uses the same LoRA settings (rank-$1$, edit layers $0{-}5$, supervision layer $l^*=5$) and training data as the geometry LoRA, and changes only the readout $\phi$ or the target $\mathbf{G}$. All readouts (1)-(4) are built from the same three probes as the helix frame, fit on the base model as described in \S\ref{appdx:helix_geometry_lora}. We describe the baselines below:
\begin{itemize}
    \item[(1)] \textbf{Direct regression.} We use only the count probe as the readout, $\phi_\texttt{count}(\mathbf{h}) = \mathbf{w}_\texttt{count}\cdot\mathbf{h} + b_\texttt{count}$, and train with the per-prompt loss (Equation~\ref{eq:loss_perprompt}), applied at every token, with the literal \texttt{character-count} as the target ($\mu_v = v$) and no spread term.
    \item[(2)] \textbf{Evenly spaced line.} We use the same readout $\phi_\texttt{count}$, and train with the per-value geometry loss (Equation~\ref{eq:full_loss}), where the targets are evenly spaced ($u_v = v/T$, the count coordinate of the ideal helix), centered at the base model's mean centroid, and scaled to radius $\alpha_r r_0$. This is equivalent to supervising only the \textit{character-count} coordinate of the helix.
    \item[(3)] \textbf{Fourier coordinates only.} We use only the two Fourier probes as the readout, $\phi_\text{circle}(\mathbf{h}) = (\mathbf{w}_{\cos}\cdot\mathbf{h} + b_{\cos},\ \mathbf{w}_{\sin}\cdot\mathbf{h} + b_{\sin})$, with the ideal circle $\mathbf{u}_v = (\cos(2\pi v/T), \sin(2\pi v/T))$ as the target, aligned to the base model's centroids by orthogonal Procrustes as with the helix.
    \item[(4)] \textbf{Base model centroids.} We use the full helix frame $\phi_\text{helix}$, but replace the ideal helix target with the base model's own per-\texttt{character-count} centroids scaled about their mean, $\boldsymbol{\mu}_v = \boldsymbol{\bar\mu}_c(\theta_0) + \alpha_r\bigl(\boldsymbol{\hat\mu}_v(\theta_0) - \boldsymbol{\bar\mu}_c(\theta_0)\bigr)$. As in the months task, this target has the same center and radius as the helix, so it differs from the geometry LoRA objective only in the shape of the target.
     \item[(5)] \textbf{Base model centroids in the PCA subspace.} We use the top-$3$ principal components of the base model's per-\texttt{character-count} centroids at layer $5$ as the readout, with the base model's own centroids in this subspace, scaled about their mean by $\alpha_r$, as the target.
\end{itemize}
For (2)--(4) and the helix, we use the hyperparameters from the main paper ($\alpha_r = 1.2$, $\tau = 0.5$, $\lambda = 1.0$, learning rate $10^{-4}$).
As a control for the readout, we also train with the ideal helix target in a random $3$-dimensional subspace of the residual stream at layer $5$.

\textbf{Results.} We report linebreaking error for each approach in Table~\ref{tab:alternate_linebreak_supervision}. 
We find that each of these supervision baselines improves over the base model, with linebreaking errors of $7.9$, $8.4$, $8.2$, and $9.0$ characters respectively, compared to the base model's linebreaking error of $13.5$ characters. However, none of them outperform the helix, which obtains an error of $7.3$ characters.
The gains from the helix over these baselines are smaller than in the months task: direct regression of the count comes close to the helix overall.
As in the months task, the base model's own geometry is a weaker target than its idealized version ($9.0$ vs.\ $7.3$), and supervising the ideal helix in a random subspace helps much less ($10.6$), indicating that the choice of readout also matters.
We also supervise within the top-$3$ PCA subspace of the base model's per-\texttt{character-count} centroids, and find that using this helix-like structure as supervision achieves a linebreaking error of $9.6$ characters (and can be lowered to $7.8$ characters by scaling by a large $\alpha_r$), which is close to the ideal helix under the probes, but does not match it. However an idealized helix in the PCA subspace does not help (achieving a linebreaking error of 13.2 characters).
This shows that supervising the count variable without the helix frame can also improve performance a lot, but the helix frame contributes some additional structure that is not captured by the other baselines.

\begin{table}[h]
\centering
\caption{Linebreaking error (characters, lower is better) for alternate forms of \texttt{character-count} supervision, mean $\pm$ standard deviation over 3 seeds. All geometry targets use the configuration from the main paper, except direct regression, whose learning rate is chosen from a small sweep.}
\label{tab:alternate_linebreak_supervision}
\begin{tabular}{llc}
\toprule
Method & Readout & Error $(\downarrow)$ \\
\midrule
Base Model & -- & $13.5$ \\
Task LoRA & -- & $4.6 \pm 0.1$ \\
\midrule
(1) Direct Regression & count probe & $7.9 \pm 0.1$ \\
(2) Evenly Spaced Line & count probe & $8.4 \pm 0.2$ \\
(3) Fourier Coordinates Only & Fourier probes & $8.2 \pm 0.3$ \\
(4) Base-Model Centroids & helix frame & $9.0 \pm 0.4$ \\
\midrule
(5) Base-Model Centroids & PCA (3D) & $9.6 \pm 0.8$ \\
Idealized Helix &  PCA (3D) &  $13.2\pm0.8$ \\
Idealized Helix, Random Subspace & random 3-d & $10.6 \pm 0.7$ \\
\midrule
Geometry LoRA (Idealized Helix) & helix frame & $7.3 \pm 0.1$ \\
\bottomrule
\end{tabular}
\end{table}

\subsection{Character-Count Manifold Experiment Details}
\label{appdx:linebreak_manifold}
In this section, we provide details about experiments characterizing the \texttt{character-count} manifold. 

\begin{figure}
\centering
\includegraphics[width=0.8\linewidth]{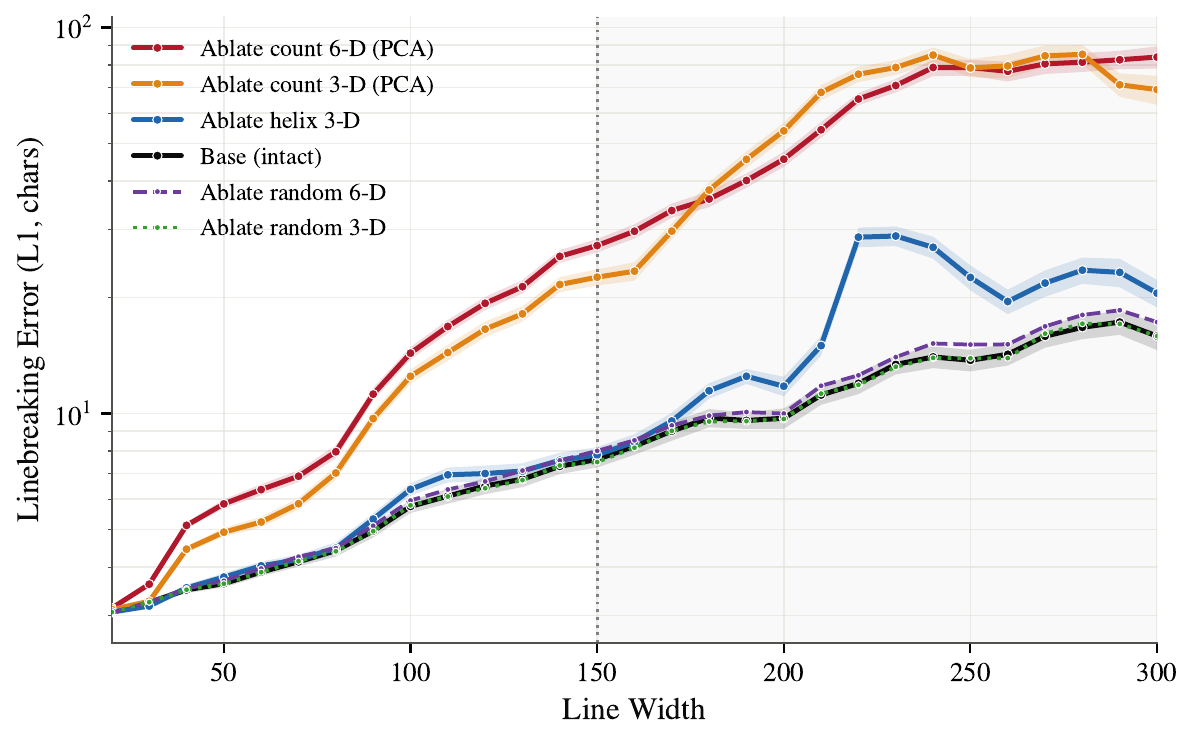}
\caption{Linebreaking error as a function of line width under zero-ablation of the
\texttt{character-count} subspace at layer $5$. Ablating the top-$6$ or top-$3$ PCA directions of the per-count means sharply raises the error, growing with width, while ablating random subspaces has no effect. Ablating the helix frame (the Fourier plane plus the count axis) has little effect in-range, but increases linebreaking error as line width increases.}
\label{fig:ablate_lberror}
\end{figure}

\textbf{PCA zero-ablation experiments.} We test whether the mean \texttt{character-count} subspace is causally used by the model by zero-ablating it at layer $5$ (i.e., projecting activations off the top-$k$ PCA directions of the per-count means) and measuring the effect on linebreaking error. As a control, we apply the same ablation to a random $k$-dimensional subspace. Ablating the count subspace increases linebreaking error by a large margin, and increasingly so at wider widths, while the random control leaves the error at the base model's level (Figure~\ref{fig:ablate_lberror}). Ablating only the top-$3$ directions already inflates the error substantially (e.g., from $7.6$ to $22.5$ characters at width $m=150$). We also tested mean-ablation, which had very similar effects to zero-ablation.

\begin{figure}
\centering
\includegraphics[width=\linewidth]{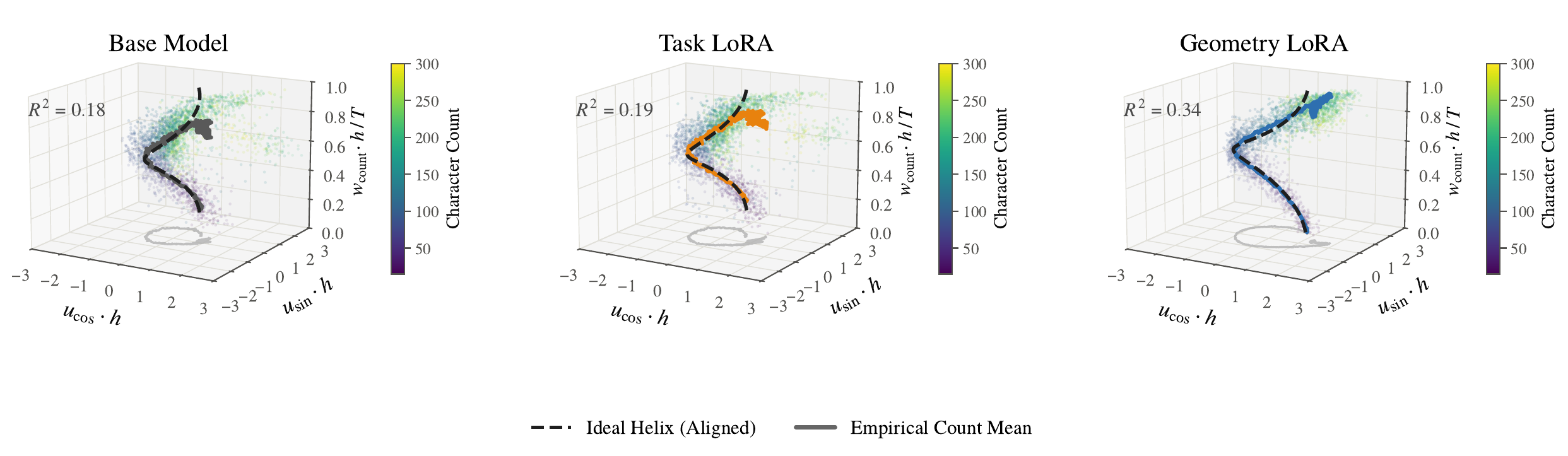}
\caption{The layer-$5$ \texttt{character-count} manifold under the \textbf{(a)} base model, \textbf{(b)} task LoRA, and \textbf{(c)} geometry LoRA. We measure the fraction of variance an idealized helix captures over all tokens ($R^2$). Plotted points are individual token hidden states projected onto a fixed helix (\S~\ref{sec:linebreak_geometric_alignment}) and colored by $\texttt{character-count}\in[20,300]$. The solid curve in each plot is the empirical per-\texttt{character-count} mean and the dashed curve is the idealized helix. In-range ($m \leq 150$), the activations follow the helix, but for the extended held-out range ($m\in[151,300]$), the alignment drops significantly for each model. The base model achieves total $R^2{=}0.18$, the task LoRA achieves $R^2{=}0.19$, and the geometry LoRA achieves $R^2{=}0.34$. The drop in alignment comes as the helix does not rotate a second time within the 3 supervised dimensions.}
\label{fig:helix_fit_to300}
\end{figure}

\textbf{Alignment to an idealized helix.} To measure how closely each model's representation follows an idealized helix, we align a period-$T$ helix ($\mathbf{u}_v {=} (\cos(2\pi v/T),\ \sin(2\pi v/T),\ v/T)$) to the layer-$5$ hidden states of all tokens---aligning each token to the ideal position for its \texttt{character-count}---via orthogonal Procrustes, and report the fraction of variance the aligned helix captures ($R^2$).
Because it is computed over all tokens rather than the per-count means, this metric captures both the shape and the within-count spread. Over the in-range counts ($20{-}150$), the base model already follows the helix well ($R^2=0.69$), the task LoRA sharpens it slightly ($R^2=0.73$), and the geometry LoRA aligns most tightly ($R^2=0.93$; Figure~\ref{fig:helix_fit}). Extended to the full count range ($20{-}300$), the alignment of all three models drops sharply ($0.18$, $0.19$, and $0.34$; Figure~\ref{fig:helix_fit_to300}).

\subsection{Linebreaking Error Distributions}
\label{appdx:error_distributions}

\begin{figure}
    \centering
    \includegraphics[width=0.8\linewidth]{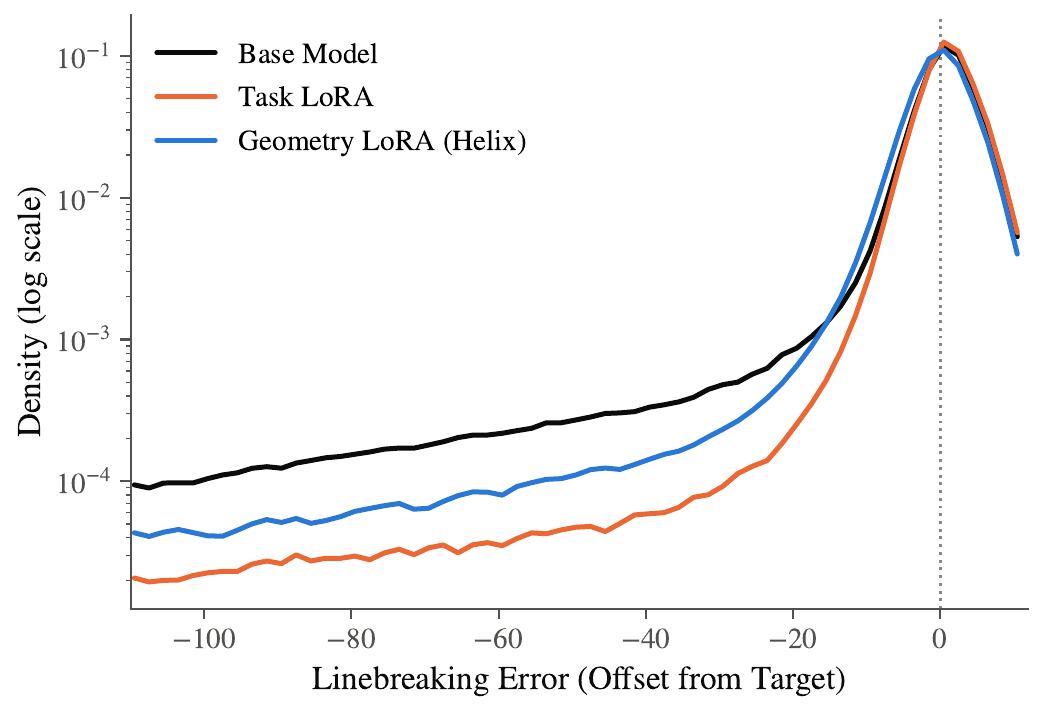}
    \caption{Linebreaking-error distribution on a logarithmic density scale. We plot the same pooled distribution as Figure~\ref{fig:linebreak_behavioral}g, but with a log density axis and a larger error range (up to -120). Near the target, the three models behave similarly. However, the base model has a heavier tail of large, early linebreaking errors of $40-100+$ characters. This is one reason why the mean linebreaking error of the base model is so large (mean error of 13.5 characters vs. 4.6 characters for task LoRA and 7.3 for the geometry LoRA).}
    \label{fig:linebreak_log_distribution}
\end{figure}

In this section, we provide more details on the linebreaking error distribution. We note that because we synthetically insert linebreaks so that each prompt has at most width $m$, it is not possible for linebreaks to be predicted too ``late'', all errors the model makes are predicting a newline too early.
While the per-break error distributions appear similar near the target width (Figure~\ref{fig:linebreak_behavioral}g in \S~\ref{sec:linebreak}), we find the base model has a much heavier tail of large early errors (Figure~\ref{fig:linebreak_log_distribution}), which accounts for its higher mean linebreaking error (mean error of 13.5 characters vs. 4.6 characters for task LoRA and 7.3 for the geometry LoRA). So while the geometry LoRA predicts early linebreaks more often, this happens closer to the target width, whereas the base model predicts early linebreaks much further away from the target width.

\subsection{Helix Activation Patching Details}
\label{appdx:helix_patching}
In this section, we provide additional details regarding the activation patching results mentioned in Section~\ref{sec:linebreak_geometric_alignment}.
Following our analysis in Appendix~\ref{appdx:subspace_difference}, we ask whether the geometry LoRA's improvement can be attributed solely to the \texttt{character-count} representation it supervises, or if there are additional benefits beyond it.
We test this by patching activations from a fine-tuned model into the base model and measuring linebreaking error.
For reference, the base model attains a linebreaking error of $13.5$ characters, the task LoRA's linebreaking error is $4.6$, and the geometry LoRA has a linebreaking error of $7.3$.

\textbf{Patching the supervised helix frame is not sufficient.} We first replace the base model's activations at layer $5$ within the helix frame $\phi_\text{helix}$ (\S\ref{sec:linebreak_geometric_alignment}) with those of a fine-tuned model: $\phi_\text{helix}(\mathbf{h}_x(\theta_0)) \leftarrow \phi_\text{helix}(\mathbf{h}_x(\theta))$, and measure the linebreaking error across token positions over wrapped FineWeb documents (\S\ref{sec:linebreak}).
Patching the geometry LoRA's helix coordinates recovers very little of its fine-tuning improvement over the base model (linebreaking error drops from $13.5 \rightarrow 12.9$ characters), and patching from the task LoRA has the same result, also achieving a linebreaking error of $12.9$ characters. Patching a random three-dimensional subspace has no effect. 
This is similar to the months task patching result (\ref{appdx:subspace_difference}), where geometric supervision yields additional benefits that activation patching within the supervised structure does not capture.

\textbf{Difference directions.} We next patch the directions along which the base and fine-tuned activations differ most.
At a given layer $\ell$, we form a matrix $\mathbf{D}$ of per-token hidden-state differences $\mathbf{h}_x(\theta) - \mathbf{h}_x(\theta_0)$ over a sample of wrapped documents and take its top-$k$ right-singular vectors $\mathbf{V}_k$. We patch the base model within this subspace, using the same subspace patching intervention as in Equation~\ref{eq:subspace_patching}, and measure linebreaking error under the intervention.

We show the results of different sized subspace patching interventions in Table~\ref{tab:patchrank}.  At layer $5$, patching the top $64$-dimensions has no effect. However, patching $k=1024$ dimensions recovers $68\%$ of the geometry LoRA's fine-tuning improvement.
Because the geometry LoRA edits only layers $0$--$5$, replacing the complete layer-$5$ hidden state recovers all of the improvement (a linebreaking error of 7.3 characters).

\begin{table}
    \centering\small
    \caption{Linebreaking error (characters) after patching the top-$k$ activation-difference directions of the geometry LoRA at layer $5$.  Linebreaking error improves with the rank, with about $68\%$ of the fine-tuning improvement recovered at $k=1024$ dimensions. The full LoRA effect can also be achieved by patching the full hidden state, shown in the far right column ($k=4096$).}
    \label{tab:patchrank}
    \begin{tabular}{l ccccccc}
    \toprule
    $k$ & $16$ & $64$ & $128$ & $256$ & $512$ & $1024$ & $4096$ (Full) \\
    \midrule
    Linebreaking error & $16.1$ & $13.6$ & $12.6$ & $11.4$ & $10.4$ & $9.3$ & $7.3$ \\
    \bottomrule
    \end{tabular}
\end{table}

\begin{figure}[h]
    \centering
    \includegraphics[width=0.75\linewidth]{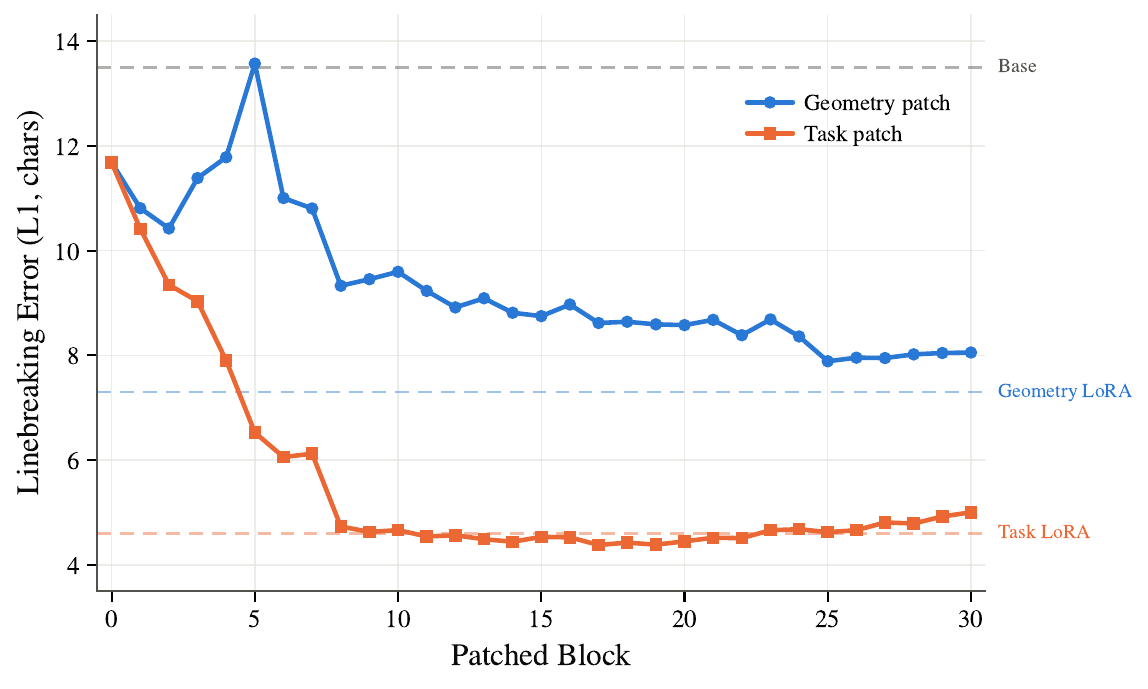}
    \caption{Linebreaking error (L1, characters) of the base model after patching each LoRA's top-$64$ activation-difference directions at a given layer. Dashed lines mark the base model and each LoRA's own error. Patching is less effective at reducing error for both LoRAs before layer 5. When patching from the geometry LoRA, the linebreaking error improves in middle and late layers, approaching the full fine-tuning difference when patching at layer $25$ onward. Patching activations from the task LoRA recovers near the task LoRA performance from layer $8$ and beyond.}
    \label{fig:patch_layer}
\end{figure}

We also find that at layers after the supervision layer, patching just $64$-dimensions can recover more of the geometry LoRA's improvements than at layer $5$. Patching at layers $14$--$20$ achieves a linebreaking error of about $9.3$ characters, recovers about $68\%$ of the geometry LoRA's improvement when patching the difference at late layers (Figure~\ref{fig:patch_layer}).
Patching activations taken from the task LoRA recovers almost all of the fine-tuning performance after layer $8$. For both settings, patching a low-dimensional difference activation from the fine-tuned model is less effective before layer 5. 

\subsection{Evaluating Off-Target Effects of Fine-tuning on Linebreaking}
\label{appdx:linebreaking-specificity}
In this section, we evaluate task and geometry LoRAs trained on the linebreaking task on several external tasks to measure their effect on the model's general capabilities. We use the same tasks and evaluation setup as in Appendix~\ref{appdx:months-specificity}: FineWeb, WikiText-103, MMLU, and GSM8K.
We find that in this case, the geometry LoRA has some minor decreases in performance, while the task LoRA has little to no effect on general model performance as measured by these other tasks (see Table~\ref{tab:offtarget_benchmarks_linebreak}).

\begin{table}[h]
\centering\small
\caption{Off-task effects of fine-tuning on the linebreaking task. We evaluate Llama-3.1-8B and LoRA fine-tuned variants on several LM benchmark tasks: FineWeb, WikiText-103, MMLU, and GSM8K to measure the effect of fine-tuning on external tasks.
Rows reporting LoRA results are means over 3 seeds.}
\label{tab:offtarget_benchmarks_linebreak}
\begin{tabular}{lccccc}
\toprule
 & Linebreaking & FineWeb & WikiText & MMLU & GSM8K \\
Model & Error (chars) $\downarrow$ & bpb $\downarrow$ & ppl $\downarrow$ & ($\%$) $\uparrow$ & ($\%$) $\uparrow$ \\
\midrule
Base Model     & 13.5 & 0.709 & 7.75 & 65.4 & 52.5 \\
Geometry LoRA  & 7.3  & 0.717 & 7.94 & 62.5 & 49.8 \\
Task LoRA      & 4.6  & 0.712 & 7.80 & 64.9 & 52.5 \\
\bottomrule
\end{tabular}
\end{table}

\end{document}